%% file: main.tex
\documentclass[10pt,twocolumn,letterpaper]{article}

\usepackage{cvpr} 
\usepackage[accsupp]{axessibility} 
\usepackage{graphicx}
\usepackage{booktabs}
\usepackage{float}
\usepackage{marvosym}
\input{preamble}

\definecolor{cvprblue}{rgb}{0.21,0.49,0.74}
\usepackage[pagebackref,breaklinks,colorlinks,citecolor=cvprblue]{hyperref}
\usepackage[sort&compress]{natbib}
\def\paperID{9850} 
\def\confName{CVPR}
\def\confYear{2024}

\title{UVEB: A Large-scale Benchmark and Baseline Towards Real-World Underwater Video Enhancement}
\author{Yaofeng Xie$^{1}$~~Lingwei Kong$^{2}$~~Kai Chen$^{2}$~~Ziqiang Zheng$^{3}$~~Xiao Yu$^2$~~Zhibin Yu$^{1,2,\hspace{0.3mm}\dag}$~~Bing Zheng$^{2}$\\
$^1$College of Electronic Engineering,Ocean University of China\\$^2$Key Laboratory of Ocean Observation and Information of Hainan Province,\\Sanya Oceanographic Institution, Ocean University of China\\$^3$Department of Computer Science and Engineering,The Hong Kong University of Science and Technology\\
{\small $\dag$~corresponding author: yuzhibin@ouc.edu.cn};~~~~\small Project website: \url{https://github.com/yzbouc/UVEB}\\
}

\begin{document}
\bibliographystyle{unsrt}
\maketitle
\input{sec/0_abstract}    
\input{sec/1_intro}

\input{sec/2_Related_Work}
\input{sec/3_Dataset}
\input{sec/4_UVE-Net}

\input{sec/5_Experiments}
\input{sec/6_Limitation}
\input{sec/7_Conclusion}

\clearpage
{
    \small
    \bibliographystyle{ieeenat_fullname}
    \bibliography{main}
}


\end{document}

%% file: preamble.tex
\usepackage[dvipsnames]{xcolor}


%% file: sec/0_abstract.tex
\begin{abstract}
Learning-based underwater image enhancement (UIE) methods have made great progress. However, the lack of large-scale and high-quality paired training samples has become the main bottleneck hindering the development of UIE. The inter-frame information in underwater videos can accelerate or optimize the UIE process. Thus, we constructed the first large-scale high-resolution underwater video enhancement benchmark (UVEB) to promote the development of underwater vision. 
It contains 1,308 pairs of video sequences and more than 453,000 high-resolution with 38\% Ultra-High-Definition (UHD) 4K frame pairs. UVEB comes from multiple countries, containing various scenes and video degradation types to adapt to diverse and complex underwater environments. We also propose the first supervised underwater video enhancement method, UVE-Net. UVE-Net converts the current frame information into convolutional kernels and passes them to adjacent frames for efficient inter-frame information exchange. By fully utilizing the redundant degraded information of underwater videos, UVE-Net completes video enhancement better. Experiments show the effective network design and good performance of UVE-Net.  
\end{abstract}

%% file: sec/1_intro.tex
\section{Introduction}

\label{sec:intro}
\hspace{0.16667in}
Underwater images and videos are essential channels representing various information, but they often suffer from color deviation and blurring due to water scattering. Underwater Image Enhancement (UIE) can improve the color deviation and blurring of underwater images, helping them to be better applied in marine observation. However, UIE is also full of challenges due to its higher ill-posedness than video dehazing~\cite{xu2023video}. Current UIE methods often cannot completely eliminate the effect of water scattering and cannot be widely used in various real underwater scenes~\cite{huang2023contrastive}.
Collecting large-scale data to train deep neural networks and utilizing the fitting ability of neural networks~\cite{fu2022uncertainty} can approximately solve these problems. The limited scale of existing datasets restricts the development of UIE~\cite{huang2023contrastive}. These factors motivate us to construct the first large-scale real-world paired underwater video enhancement dataset. 

Before data-driven UIE methods became popular, people improved the quality of underwater images mainly by estimating physical priors or adjusting image pixel values \cite{lu2015contrast,10.5555/180895.180940}. The emergence of Generate adversarial network~\cite{2014Generative} (GAN) inspired people to explore synthetic paired UIE datasets and UIE methods based on GAN \cite{li2017watergan,fabbri2018enhancing}. The first paired underwater image enhancement benchmark UIEB~\cite{li2019underwater} was proposed in 2019. Paired real-world UIE datasets like UIEB~\cite{li2019underwater} significantly boost the research on supervised UIE methods \cite{fu2022uncertainty,liu2022adaptive}.

However, UIE research is still full of challenges. Although underwater images are not hard to collect, obtaining calibrated paired underwater images with sufficient variety is expensive and difficult~\cite{akkaynak2019sea}. It makes the existing paired real underwater datasets relatively small in scale. A small-scale UIE dataset~\cite {huang2023contrastive} may increase the risk of overfitting the learned models. The requirements for large-scale, real-world paired training samples have become the main bottleneck hindering the development of UIE. Underwater tasks use more videos than single images, and the redundant information of adjacent frames in videos can accelerate or optimize the image enhancement process. 

Considering the above factors, we collect high-resolution videos of diverse underwater scenes with video quality scores to build the first large-scale underwater video enhancement benchmark (UVEB). UVEB contains 1,308 underwater video pairs and 453,874 high-resolution frame pairs. To our knowledge, UVEB is also the largest Ultra-High-Definition (UHD) 4K video dataset (containing 173,797 pairs of UHD 4K frames) in the video enhancement/restoration field and the largest video dataset in the underwater vision field.

To enrich the diversity of samples, we collect underwater videos from multiple regions of the world (more than 20 countries), various underwater scenes (\emph{e,g,} coastal waters, distant sea, rivers, lakes, ports, swimming pools, aquariums, etc.), diverse color casts (\emph{e,g,} blue, green, yellow, white, other colors), and insufficient light underwater videos to construct UVEB. We also provide 2616 manually annotated raw video and ground truth (GT) quality scores to characterize and increase the sample reliability. 

Based on the UVEB dataset, we also provide the first supervised underwater video enhancement network, UVE-Net.
Most existing video enhancement/restoration methods achieve better results through aligning \cite{dai2017deformable,chan2022basicvsr++,zhang2021learning} or aggregating \cite{patil2022video,li2023progressive,jin2023multi} adjacent frames information at the feature or pixel level. While two ways often have a large computational burden and inaccurate frame alignment sometimes also introduces bias to image restoration~\cite{li2023simple}. 
UVE-Net heuristically explores more efficiently and directly inter-frame information interaction at the action level (convolution kernel) without frame alignment or aggregation. 

Unlike the image-level UIE methods, UVE-Net can use the inter-frame information and convert the enhancement process of the low-resolution downsampled middle frame into convolutional kernels and transmit them to the frames to be restored, guiding the frames to complete enhancement more efficiently. In this way, UVE-Net greatly improves the enhancement effect of the underwater videos with less additional computational costs.

We summarize the main contributions as follows:
\begin{itemize}
  \item We collect the first large-scale (1308 pairs of video sequences, 453,874 frame pairs) real-world underwater video enhancement dataset, UVEB. UVEB contains high-resolution video with various scenes and diverse video degradation types.
  \item We provide 2616 additional video quality scores for GT and raw videos. Sufficient experiments confirm the superiority and reliability of the UVEB dataset.
  \item We propose the first supervised underwater video enhancement method, UVE-Net. UVE-Net efficiently utilizes the enhancement process of the downsampled middle frames to guide the underwater video sequences achieve better enhancement.
\end{itemize}

%% file: sec/2_Related_Work.tex
\section{Related Work}

\label{sec: Related Work}

\begin{table*}[t]
\caption{Comparison with SOTA real paired underwater datasets and video restoration datasets.} 
\centering

\resizebox{0.95\textwidth}{!}{%
    \begin{tabular}{cccccccc}
      \toprule  
      Datasets& Venue& Sequece& Frame& Resolution& Annotation \\
      \midrule  
      UIEB~\cite{li2019underwater}& \emph{$T\!I\!P\hspace{0.15em}^{\prime}$\hspace{0.25em}19}& None& 0.89k& 299$\times$168\ $\sim$ 2180$\times$1447\ & Underwater image enhancement\\
      LSUI~\cite{peng2023u}& \emph{$T\!I\!P\hspace{0.15em}^{\prime}$\hspace{0.25em}23}& None& 5k& 256$\times$256\ $\sim$ 1280$\times$1024\ & Underwater image enhancement\\
      \midrule  
      UTB180~\cite{alawode2022utb180}& \emph{$A\!C\!C\!V\hspace{0.15em}^{\prime}$\hspace{0.25em}22}&180&58K & 1920$\times$1080& Underwater video object tracking\\
      DRUVA~\cite{varghese2023self}& \emph{$I\!C\!C\!V\hspace{0.15em}^{\prime}$\hspace{0.25em}23}& 20& 6.11K& 1920$\times$1080& Underwater video depth estimation\\
      \midrule  
      HazeWorld~\cite{xu2023video} & \emph{$C\!V\!P\!R\hspace{0.15em}^{\prime}$\hspace{0.25em}23}& 1271& 326K& 960$\times$720\ $\sim$1588$\times$720& Video dehazing\\
      RVSD~\cite{chen2023snow}& \emph{$I\!C\!C\!V\hspace{0.15em}^{\prime}$\hspace{0.25em}23}& 110& 11.423K& 640$\times$480\ $\sim$3840$\times$2160 & Video desnowing\\
      LHP-Rain~\cite{guo2023sky}& \emph{$I\!C\!C\!V\hspace{0.15em}^{\prime}$\hspace{0.25em}23}& 3000& 1000K& 1920$\times$1080& Video deraining\\
      \midrule  
      Ours& & 1308& 453.874K& 960$\times$528\ $\sim$ 3840$\times$2160\ & Underwater video enhancement\\
      \bottomrule 
    \end{tabular}
    }
    \label{tab:comparison}
\end{table*}

{\bf Underwater Image Enhancement Datasets.}
Completely removing water scattering to collect ideal underwater GT images is difficult. RUIE~\cite{RUIEORE} collects a variety of underwater image images to be enhanced and constructs a test set. Some methods like UWCNN~\cite{li2020underwater} and WaterGAN~\cite{li2017watergan} use land images as GT to synthesize underwater images. However, due to the significant differences between the synthetic image domain and the real underwater image domain, the methods trained from synthetic data are difficult to apply to diverse real-world underwater scenes.
Building UIE datasets with real underwater images with manual voted labels is another solution. 
UIEB~\cite{li2019underwater} and LSUI~\cite{peng2023u} chose the best results produced by current UIE methods as GT. Many new UIE methods such as PUIE~\cite{fu2022uncertainty}, LANet~\cite{liu2022adaptive}, and FspiralGAN~\cite{guan2023fast} trained on these data have shown remarkable improvements in enhancing real underwater images. SAUD~\cite{SAUD} construct an UIE quality evaluation dataset with manual voted labels. We use manual voted labels to construct the UVEB dataset in this work.

\noindent{\bf Underwater Video/Video Enhancement Datasets.}
The high collection and annotation cost~\cite{akkaynak2019sea} of underwater video results in existing underwater video datasets~\cite{varghese2023self,alawode2022utb180} being small in scale or with limited scenes. DRUVA~\cite{varghese2023self} captured 20 videos for underwater video depth estimation. UTB180~\cite{alawode2022utb180} offers 180 video sequences for underwater video object tracking. Models trained with insufficient underwater data would be hard to adapt to the diverse and intricate underwater conditions.  

In contrast, in-air video datasets often have a larger scale. For example, the video dehazing dataset HazeWorld~\cite{xu2023video}, contains 1271 video pairs. Moreover, the LHP-Rain~\cite{guo2023sky} dataset in video deraining includes one million FHD frames. 
To obtain a large-scale underwater video dataset with rich scenes and narrow the development gap between underwater video enhancement and other low-level visual tasks, we construct a large-scale underwater video dataset UVEB to promote the development of underwater vision. 

\noindent{\bf Underwater Image Enhancement Methods.}
UIE methods can be roughly divided into learning-free methods and learning-based methods. The former enhances underwater images through prior estimation \cite{5206515,carlevaris2010initial,galdran2015automatic,lu2015contrast,akkaynak2019sea} or image pixel value adjustment~\cite{10.5555/180895.180940,zhang2022underwater}. The latter learns the mapping of high-quality images through feature extraction, such as weakly supervised UIE methods MateUE~\cite{zhang2023metaue}, Semi-UIR~\cite{huang2023contrastive}, and supervised deep learning methods SpiralGAN~\cite{han2020underwater}, LANet~\cite{liu2022adaptive}, and PUIE~\cite{fu2022uncertainty}.
Research in UIE is currently more focused on designing better UIE methods \cite{fu2022uncertainty,liu2022adaptive}. A few methods, such as 
FA$^+$Net~\cite{jiang2023five} and FspiralGAN~\cite{guan2023fast}, aim to develop faster lightweight networks suitable for underwater conditions with limited computational resources.
These methods often sacrifice quality to ensure speed. 
Quality is still a more important issue in the UIE field, and we explore efficient ways to utilize redundant information in underwater video to achieve better UIE.


\noindent{\bf Video Restoration Methods.}
Existing video restoration methods assist in better image restoration for the current frame image by aligning \cite{dai2017deformable,chan2022basicvsr++,zhang2021learning} or aggregating \cite{patil2022video,li2023progressive,zhang2018adversarial,jin2023multi} adjacent frame information at the feature or pixel level. Although the former can effectively use inter-frame information, inaccurate frame alignment sometimes brings bias to image restoration~\cite{li2023simple}, and frame alignment often has a large computational burden. Although the latter can fully utilize inter-frame information through multi-level aggregation of adjacent frames or three-dimensional convolution to fusion spatiotemporal information, it often has a high computational cost and low efficiency in utilizing redundant information of adjacent frames. We enlighteningly carry out efficient interaction of inter-frame information at the action level (convolutional kernel), transforming the enhancement process of the downsampled middle frame into convolutional kernels (action instructions) and pass them to the current frames to be enhanced, helping them complete enhancement more efficiently.

%% file: sec/3_Dataset.tex
\section{ Large-scale real-world paired Benchmark}

\label{sec: Dataset}
\subsection{Benchmark Collection }

\begin{figure*}[htbp]
    \centering
    \rotatebox{90}{\scriptsize{~~~~~~Raw}}
    \begin{minipage}[b]{0.1185\linewidth}
        \includegraphics[width=\linewidth]{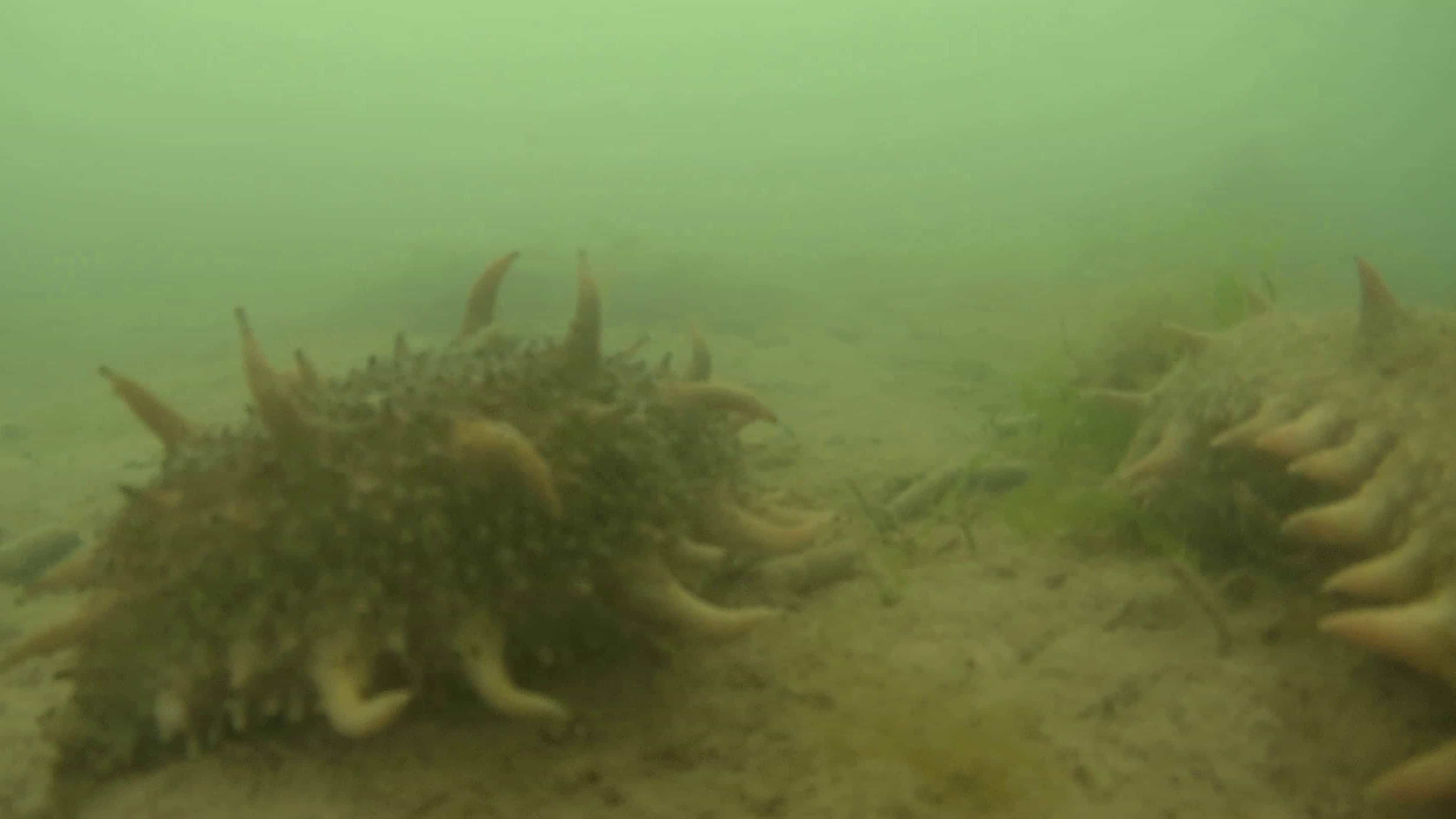}
        \begin{picture}(-20,-20)
            \put(-26,26){\textcolor{white}{17.36}}
        \end{picture}
    \end{minipage}
    \begin{minipage}[b]{0.1185\linewidth}
        \includegraphics[width=\linewidth]{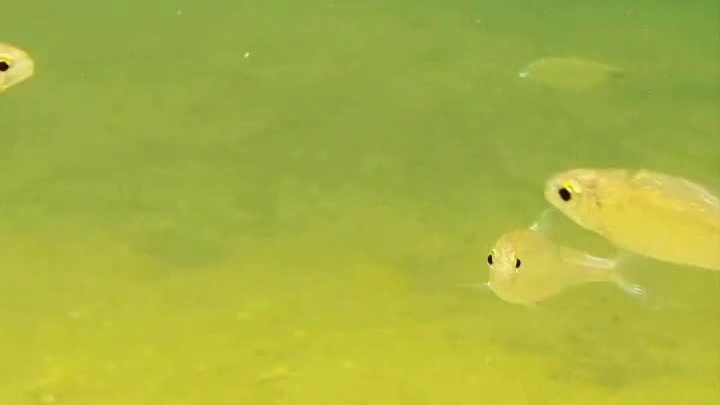}
        \begin{picture}(-10,-10)
            \put(-26,26){\textcolor{white}{35.87}}
        \end{picture}
    \end{minipage}
    \begin{minipage}[b]{0.1185\linewidth}
        \includegraphics[width=\linewidth]{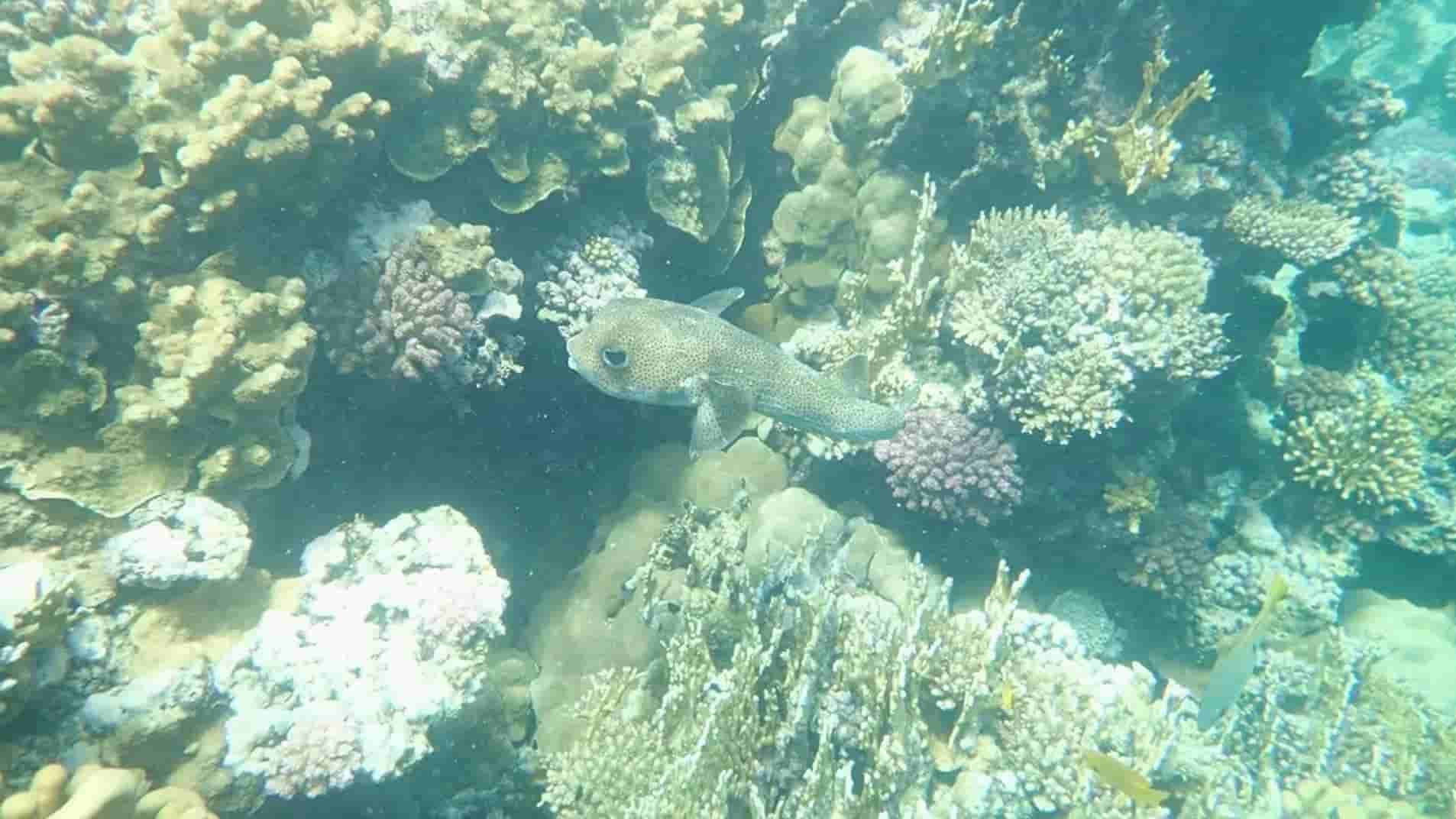}
        \begin{picture}(-10,-10)
            \put(-26,26){\textcolor{white}{58.27}}
        \end{picture}
    \end{minipage} 
    \begin{minipage}[b]{0.1185\linewidth}
        \includegraphics[width=\linewidth]{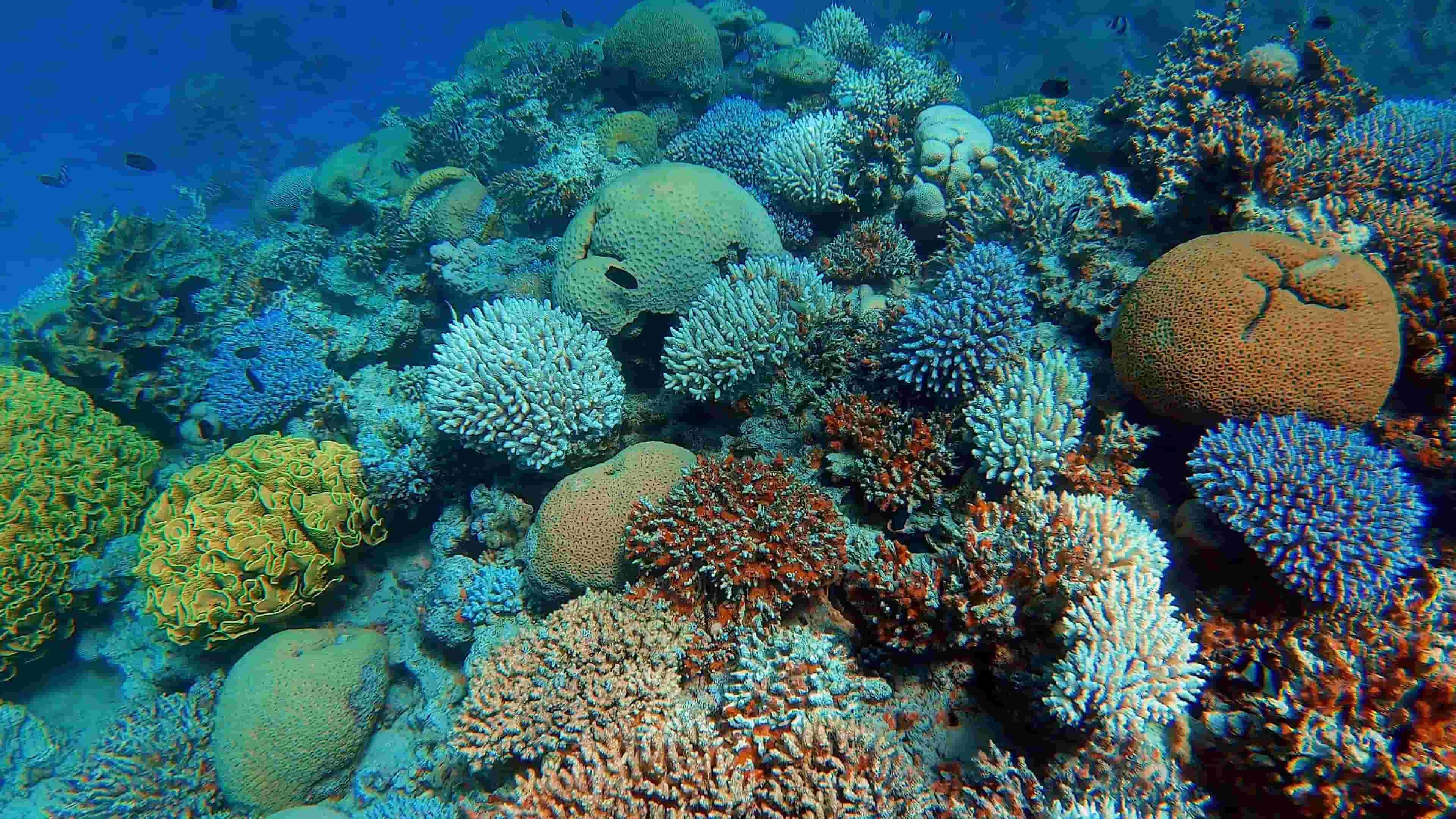}
        \begin{picture}(-10,-10)
            \put(-26,26){\textcolor{white}{58.33}}
        \end{picture}
    \end{minipage}
    \begin{minipage}[b]{0.1185\linewidth}
        \includegraphics[width=\linewidth]{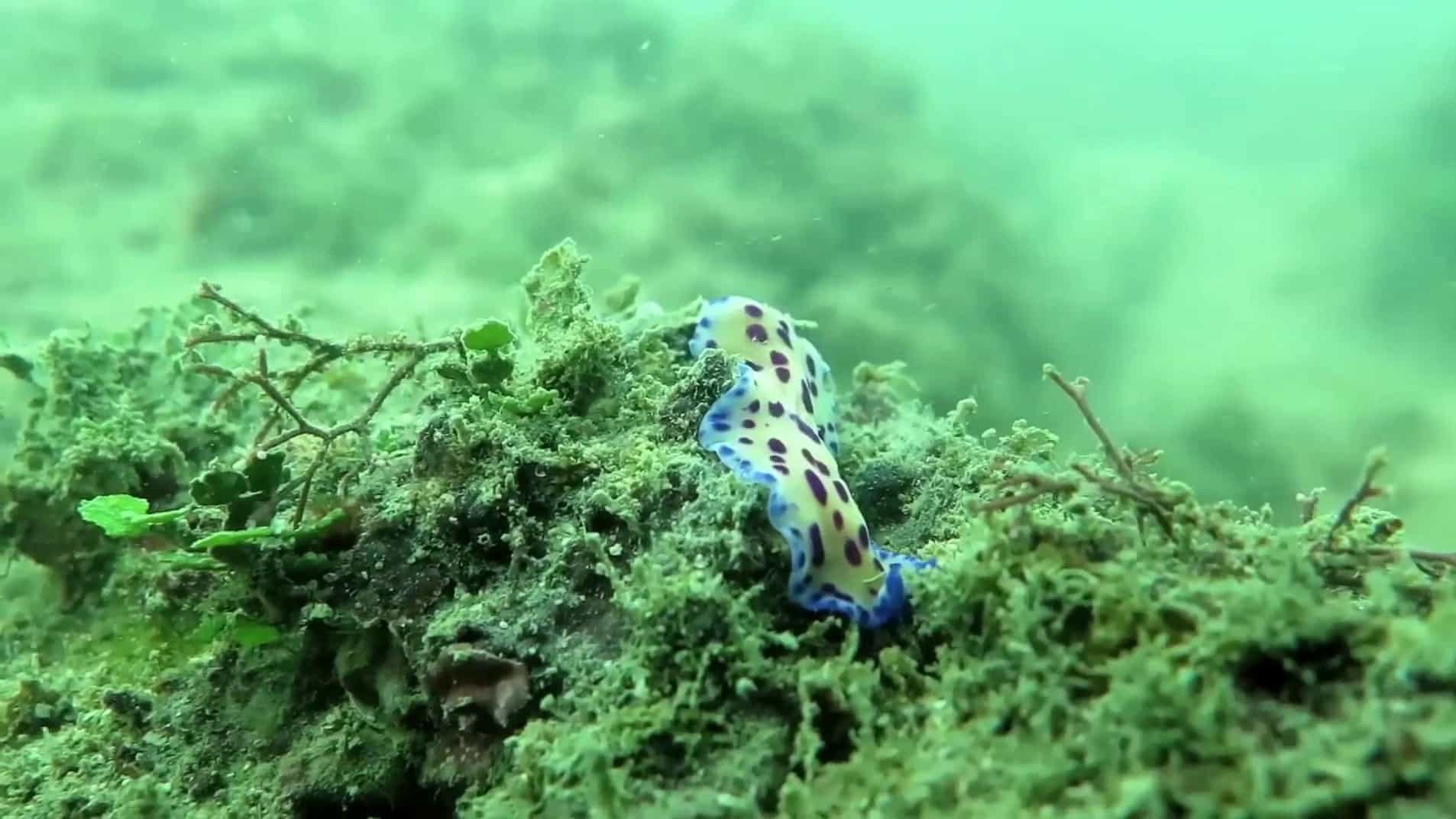}
        \begin{picture}(-10,-10)
            \put(-26,26){\textcolor{white}{51.43}}
        \end{picture}
    \end{minipage}
    \begin{minipage}[b]{0.1185\linewidth}
        \includegraphics[width=\linewidth]{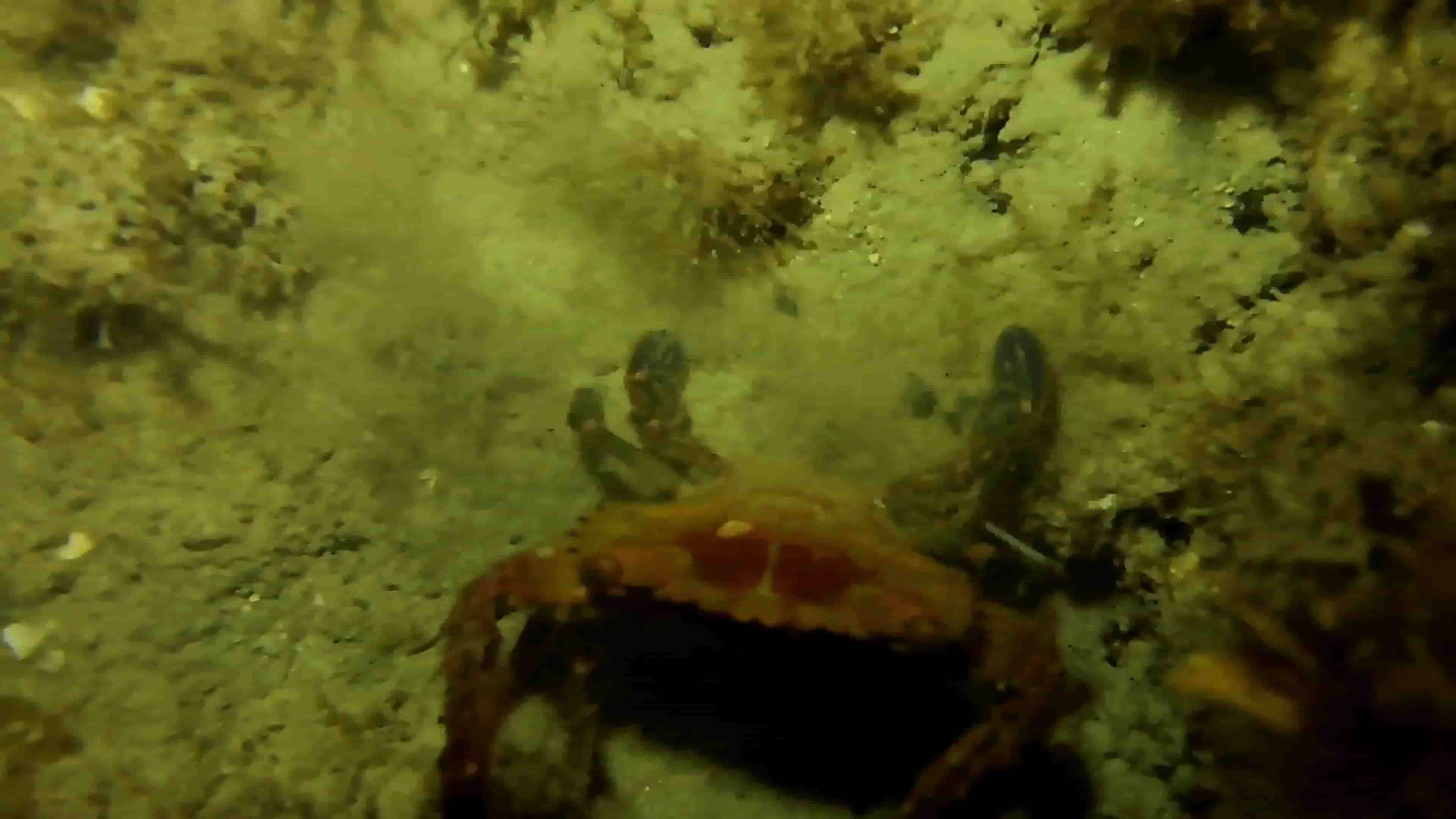}
        \begin{picture}(-10,-10)
            \put(-26,26){\textcolor{white}{40.71}}
        \end{picture}
    \end{minipage}
    \begin{minipage}[b]{0.1185\linewidth}
        \includegraphics[width=\linewidth]{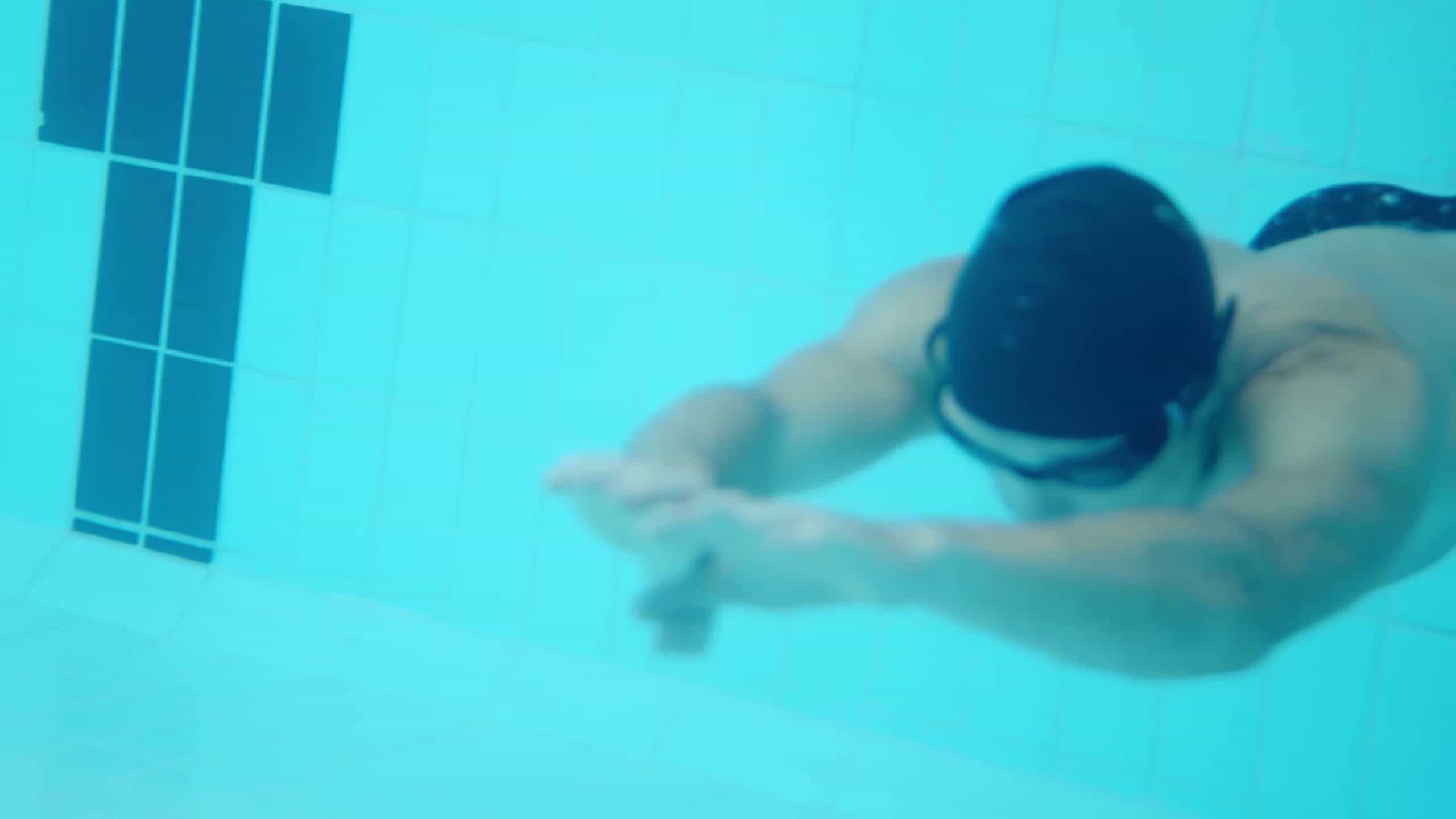}
        \begin{picture}(-10,-10)
            \put(-26,26){\textcolor{white}{43.73}}
        \end{picture}
    \end{minipage}
    \begin{minipage}[b]{0.1185\linewidth}
        \includegraphics[width=\linewidth]{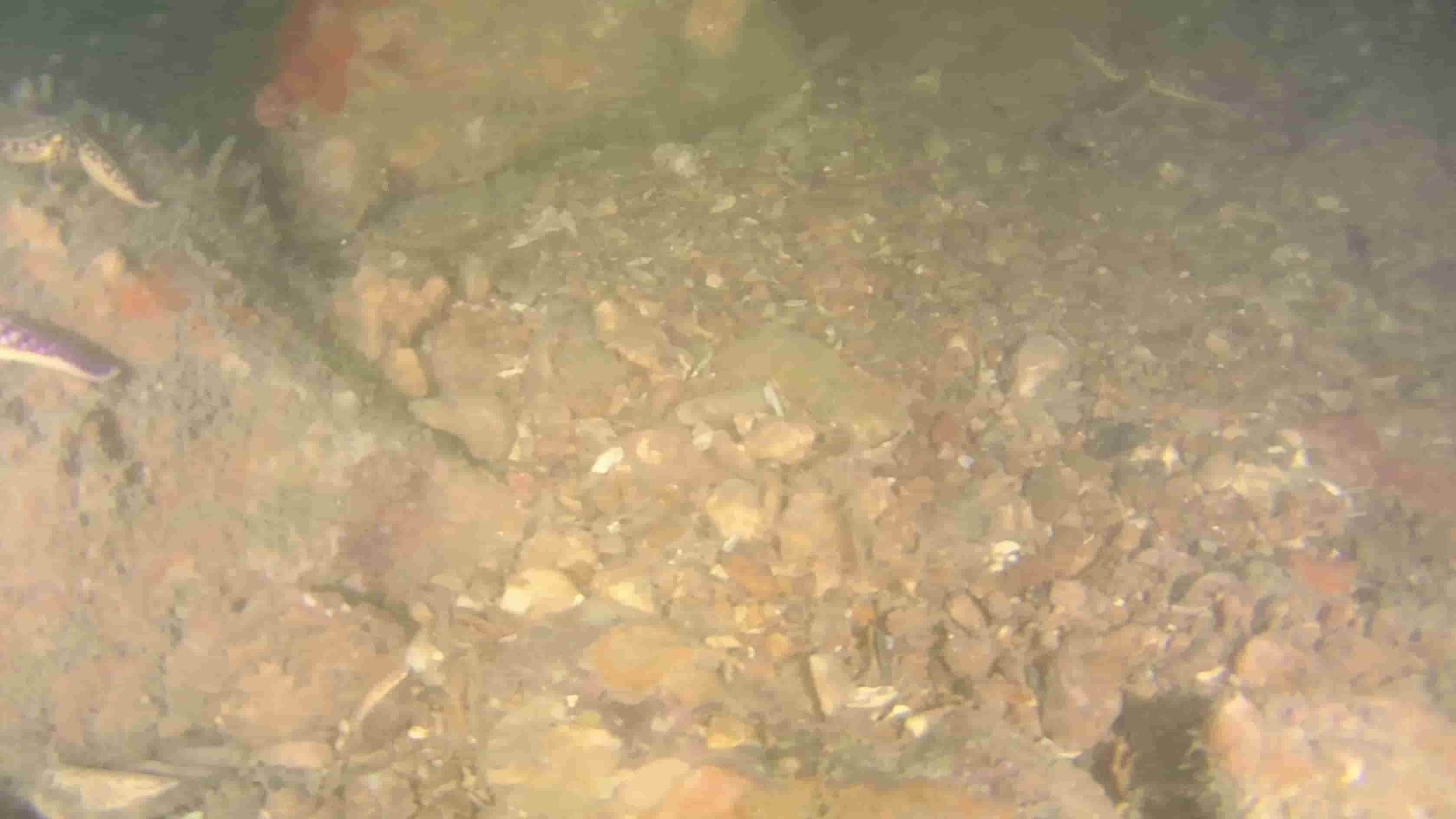}
    \begin{picture}(-10,-10)
            \put(-26,26){\textcolor{white}{25.07}}
        \end{picture}
    \end{minipage}
\rotatebox{90}{\scriptsize{~~~~~~GT}}
\begin{minipage}[b]{0.1185\linewidth}
    \includegraphics[width=\linewidth]{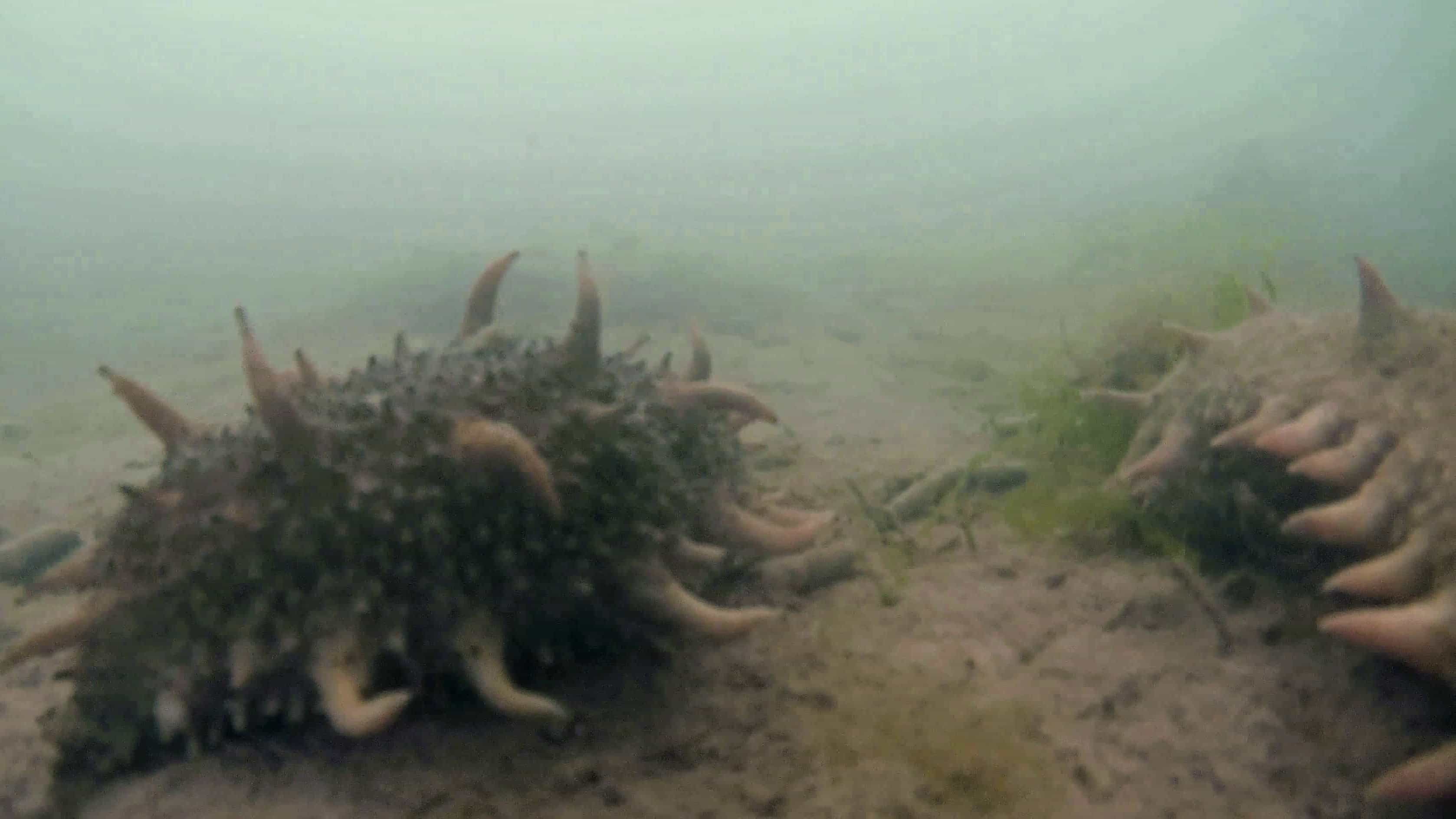}
    \begin{picture}(-10,-10)
        \put(-26,26){\textcolor{white}{24.25}}
    \end{picture}
\end{minipage}
\begin{minipage}[b]{0.1185\linewidth}
    \includegraphics[width=\linewidth]{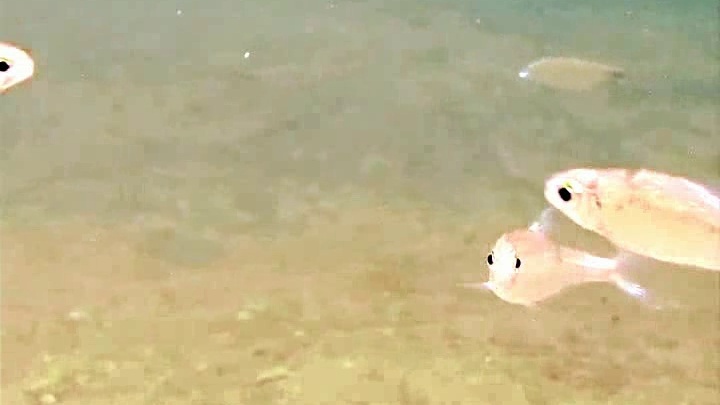}
    \begin{picture}(-10,-10)
        \put(-26,26){\textcolor{white}{52.31}}
    \end{picture}
\end{minipage}
\begin{minipage}[b]{0.1185\linewidth}
    \includegraphics[width=\linewidth]{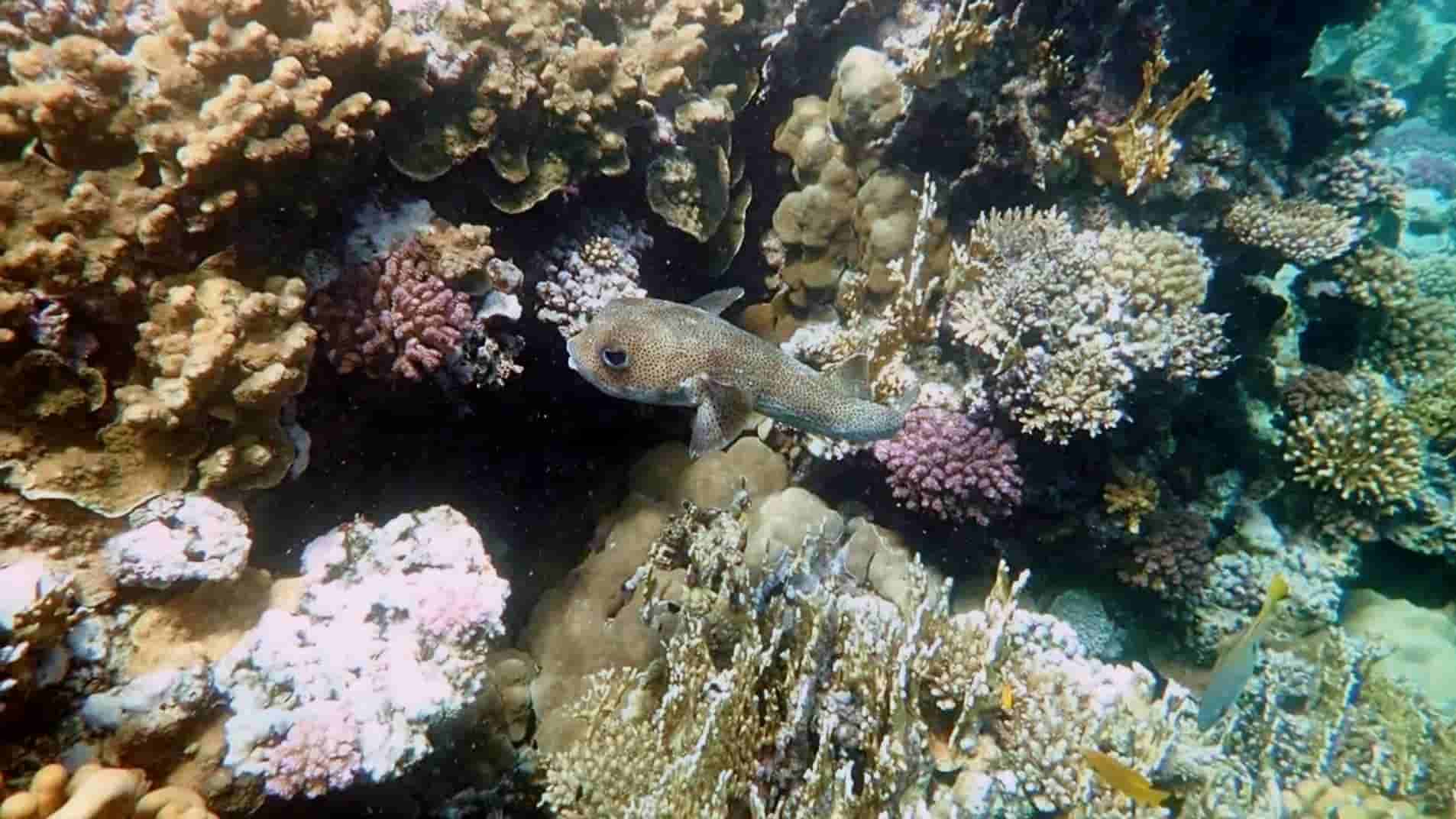}
    \begin{picture}(-10,-10)
        \put(-26,26){\textcolor{white}{72.45}}
    \end{picture}
\end{minipage} 
\begin{minipage}[b]{0.1185\linewidth}
    \includegraphics[width=\linewidth]{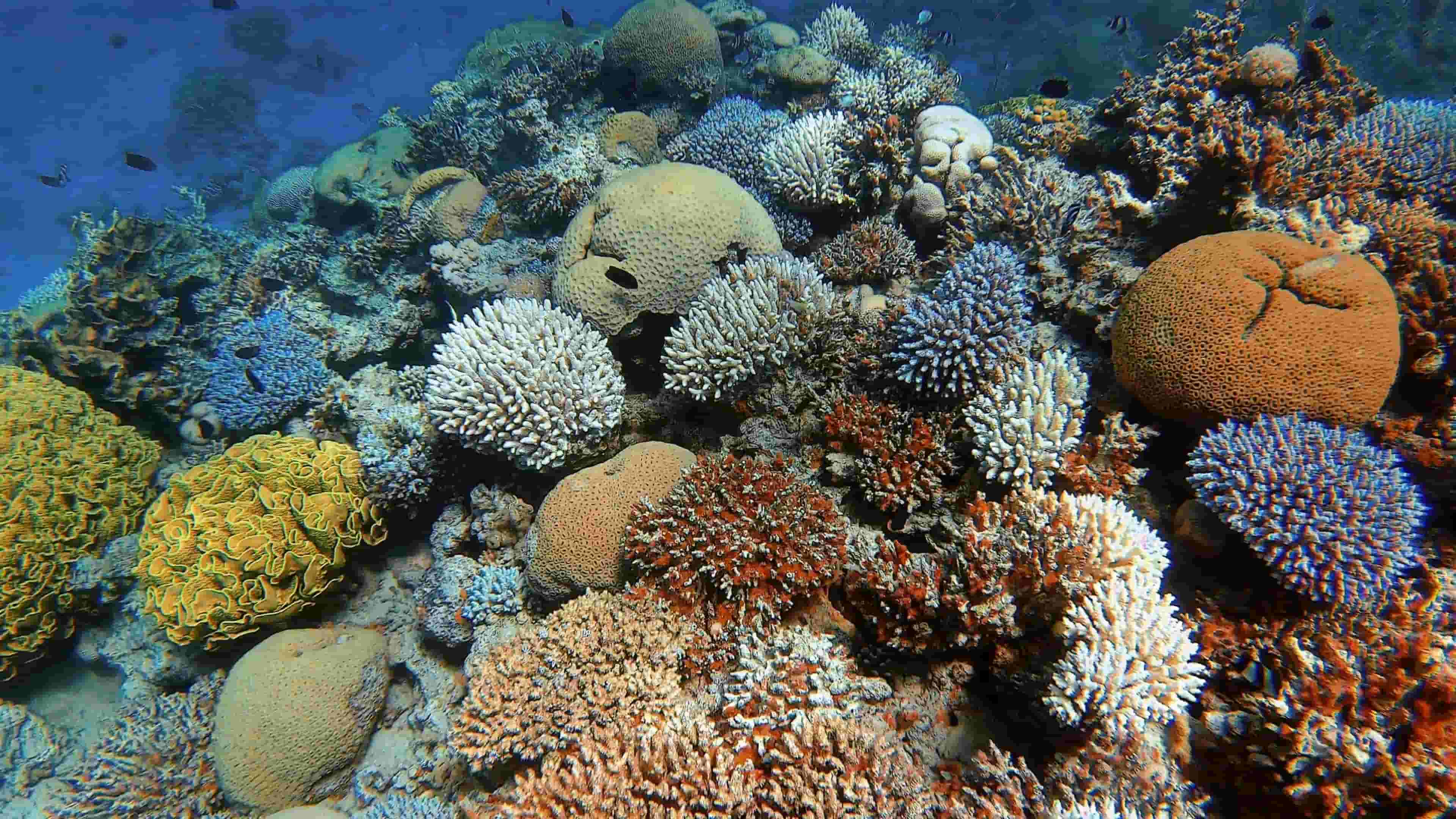}
    \begin{picture}(-10,-10)
        \put(-26,26){\textcolor{white}{72.62}}
    \end{picture}
\end{minipage}
\begin{minipage}[b]{0.1185\linewidth}
    \includegraphics[width=\linewidth]{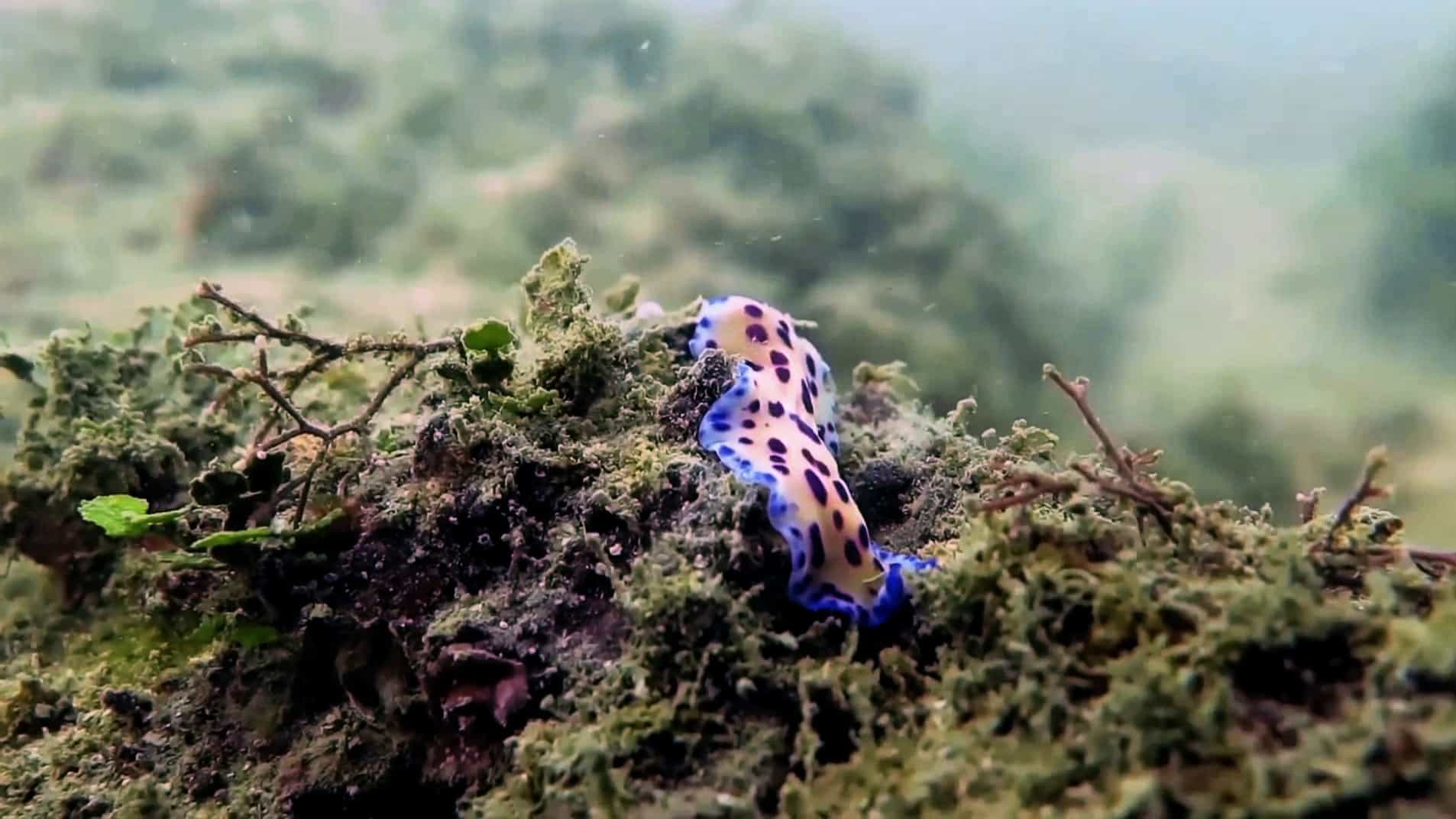}
    \begin{picture}(-10,-10)
        \put(-26,26){\textcolor{white}{70.57}}
    \end{picture}
\end{minipage}
\begin{minipage}[b]{0.1185\linewidth}
    \includegraphics[width=\linewidth]{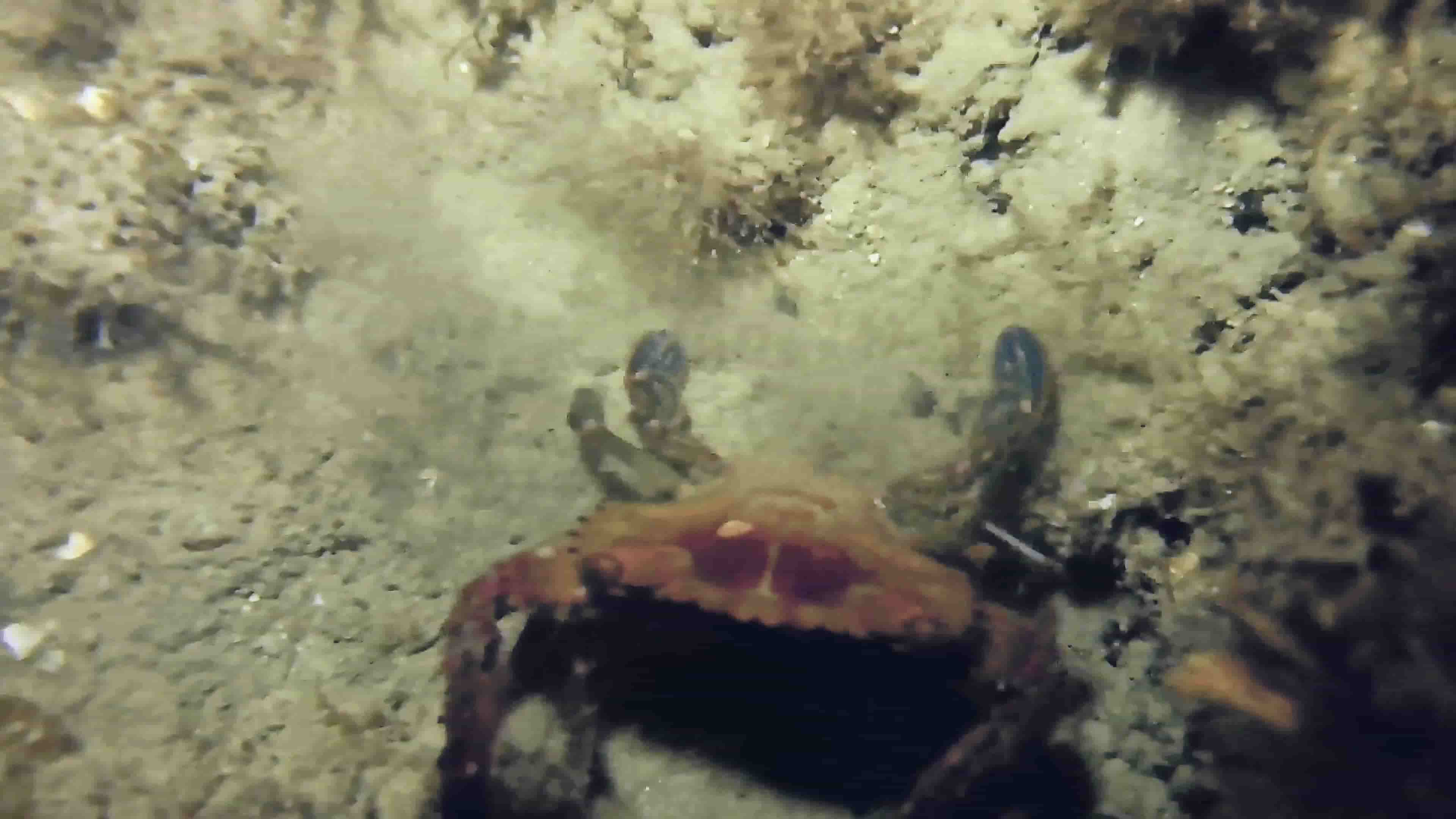}
    \begin{picture}(-10,-10)
        \put(-26,26){\textcolor{white}{51.29}}
    \end{picture}
\end{minipage}
\begin{minipage}[b]{0.1185\linewidth}
    \includegraphics[width=\linewidth]{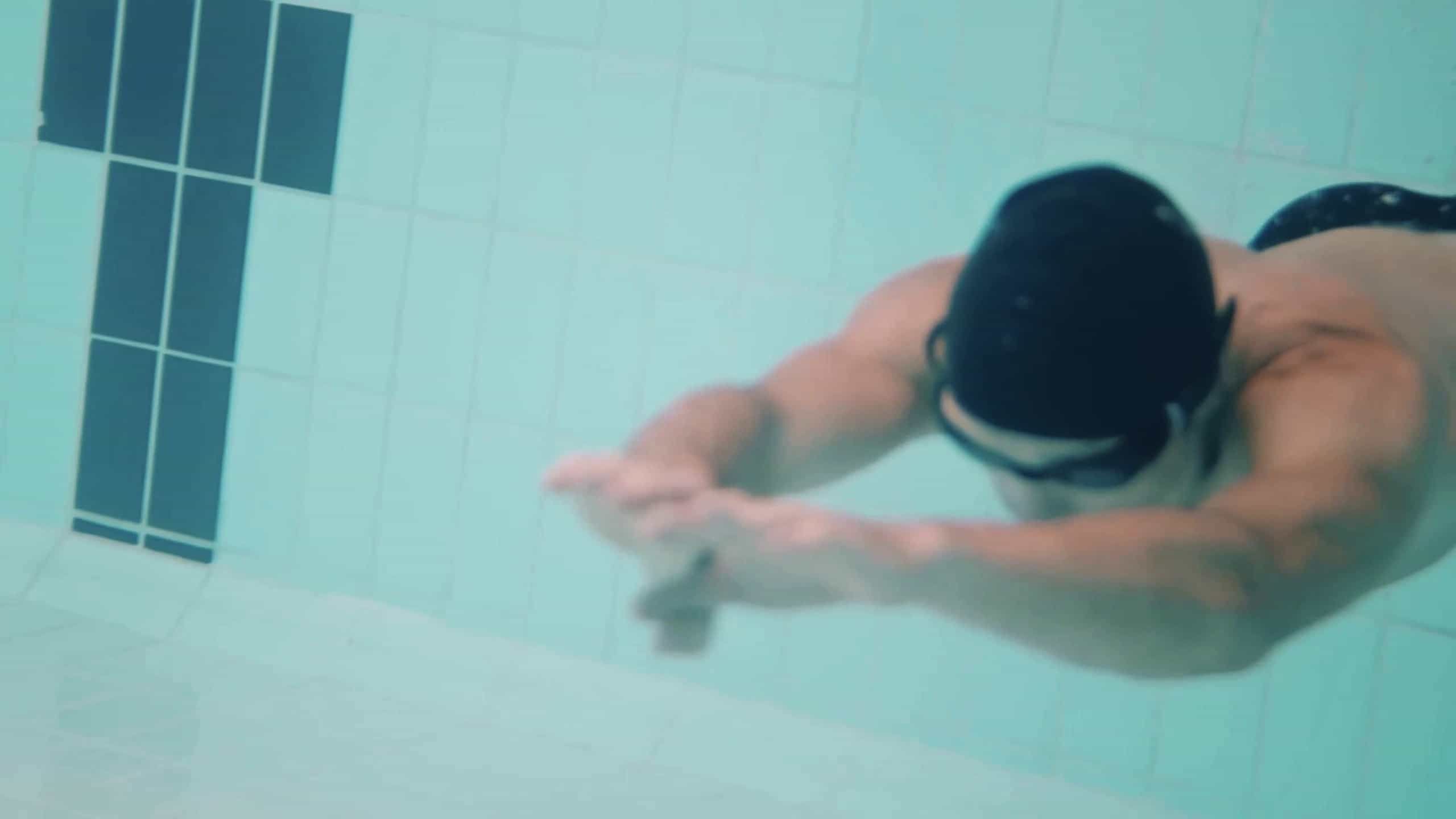}
    \begin{picture}(-10,-10)
        \put(-26,26){\textcolor{white}{54.29}}
    \end{picture}
\end{minipage}
\begin{minipage}[b]{0.1185\linewidth}
    \includegraphics[width=\linewidth]{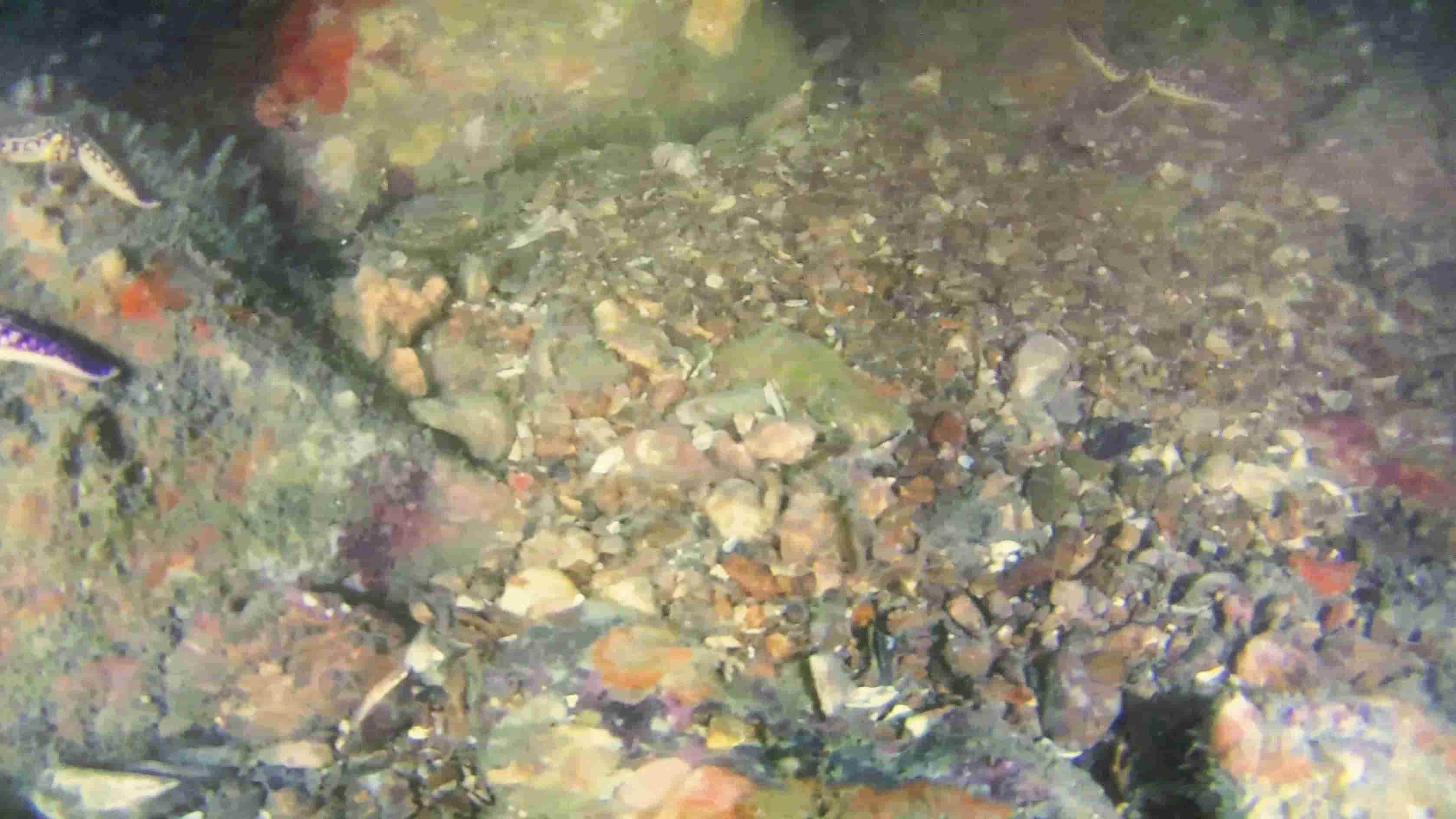}
    \begin{picture}(-10,-10)
        \put(-26,26){\textcolor{white}{36.21}}
    \end{picture}
\end{minipage}
\rotatebox{90}{\scriptsize{~~~~~~Raw}}
\begin{minipage}[b]{0.1185\linewidth}
    \includegraphics[width=\linewidth]{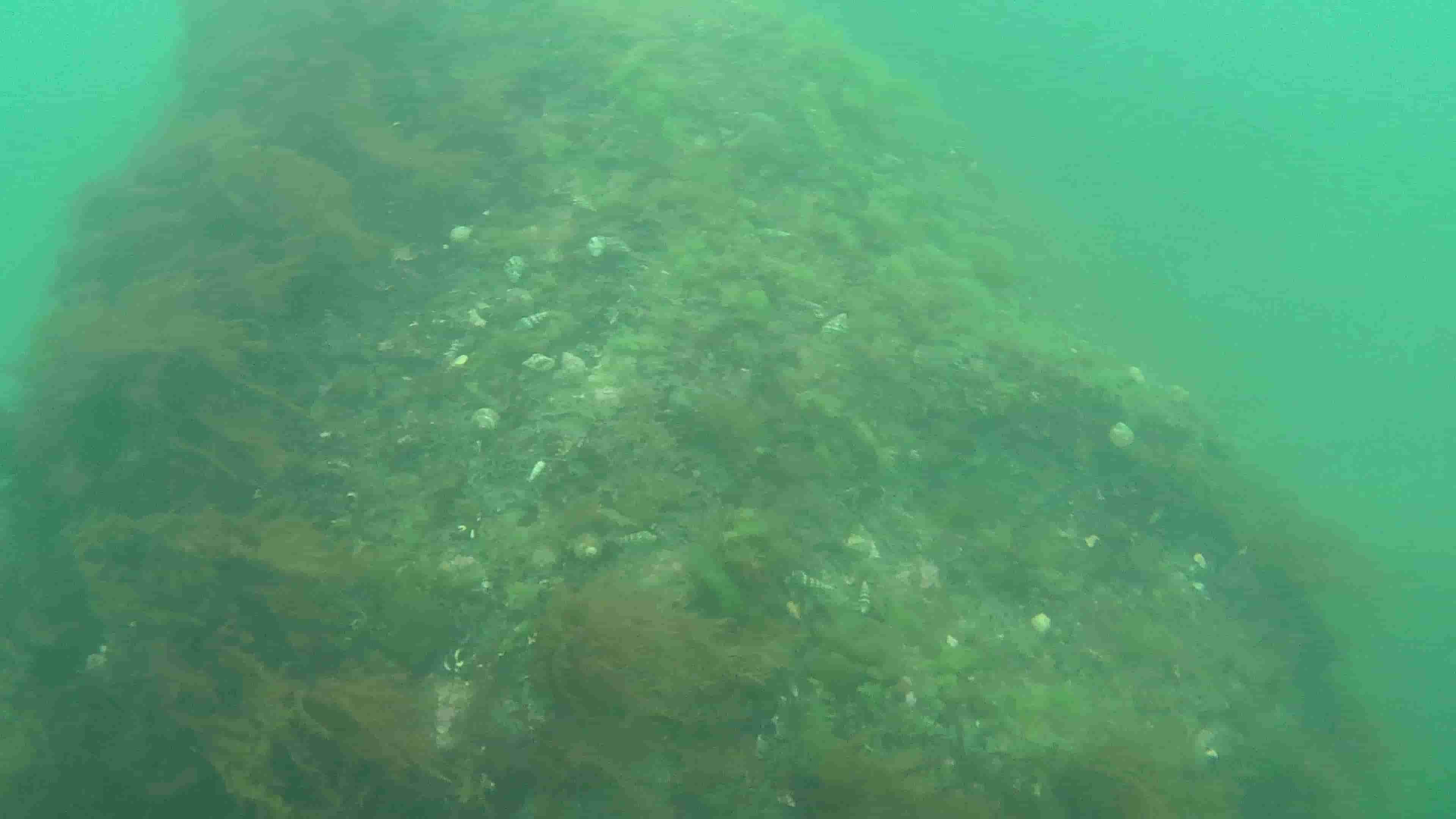}
    \begin{picture}(-10,-10)
        \put(-26,26){\textcolor{white}{22.86}}
    \end{picture}
\end{minipage}
\begin{minipage}[b]{0.1185\linewidth}
    \includegraphics[width=\linewidth]{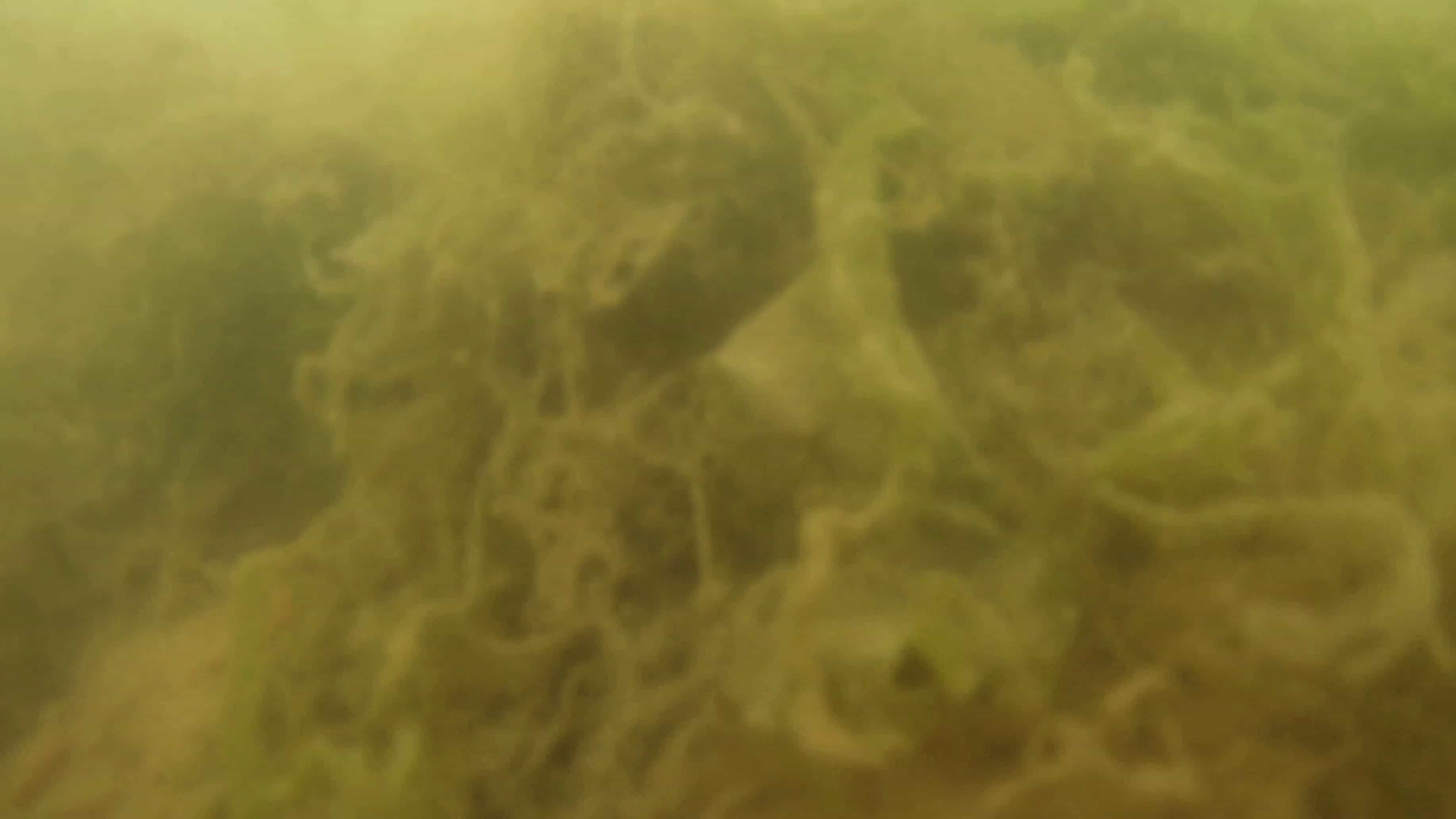}
    \begin{picture}(-10,-10)
        \put(-26,26){\textcolor{white}{15.27}}
    \end{picture}
\end{minipage}
\begin{minipage}[b]{0.1185\linewidth}
    \includegraphics[width=\linewidth]{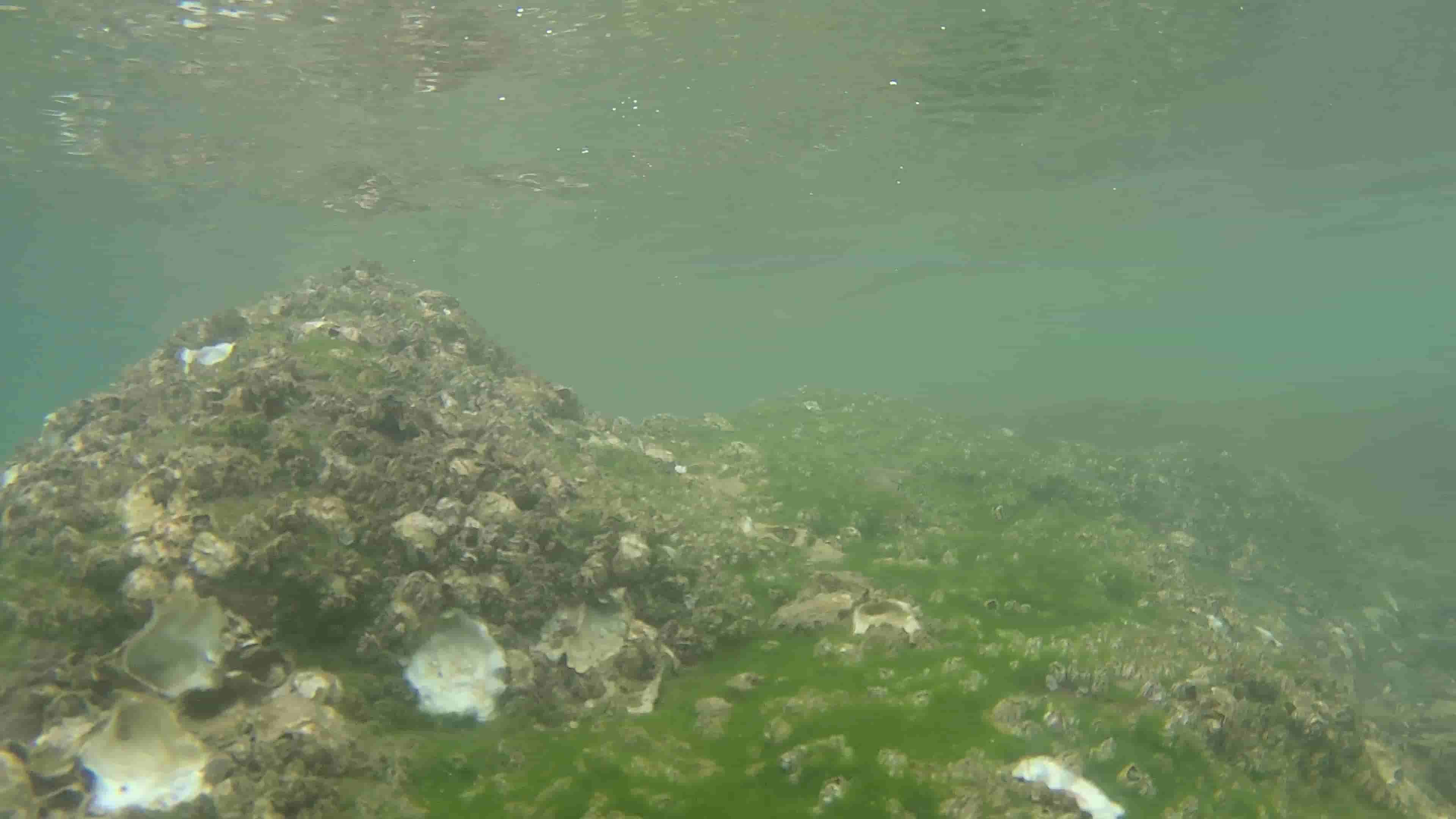}
    \begin{picture}(-10,-10)
        \put(-26,26){\textcolor{white}{35.87}}
    \end{picture}
\end{minipage} 
\begin{minipage}[b]{0.1185\linewidth}
    \includegraphics[width=\linewidth]{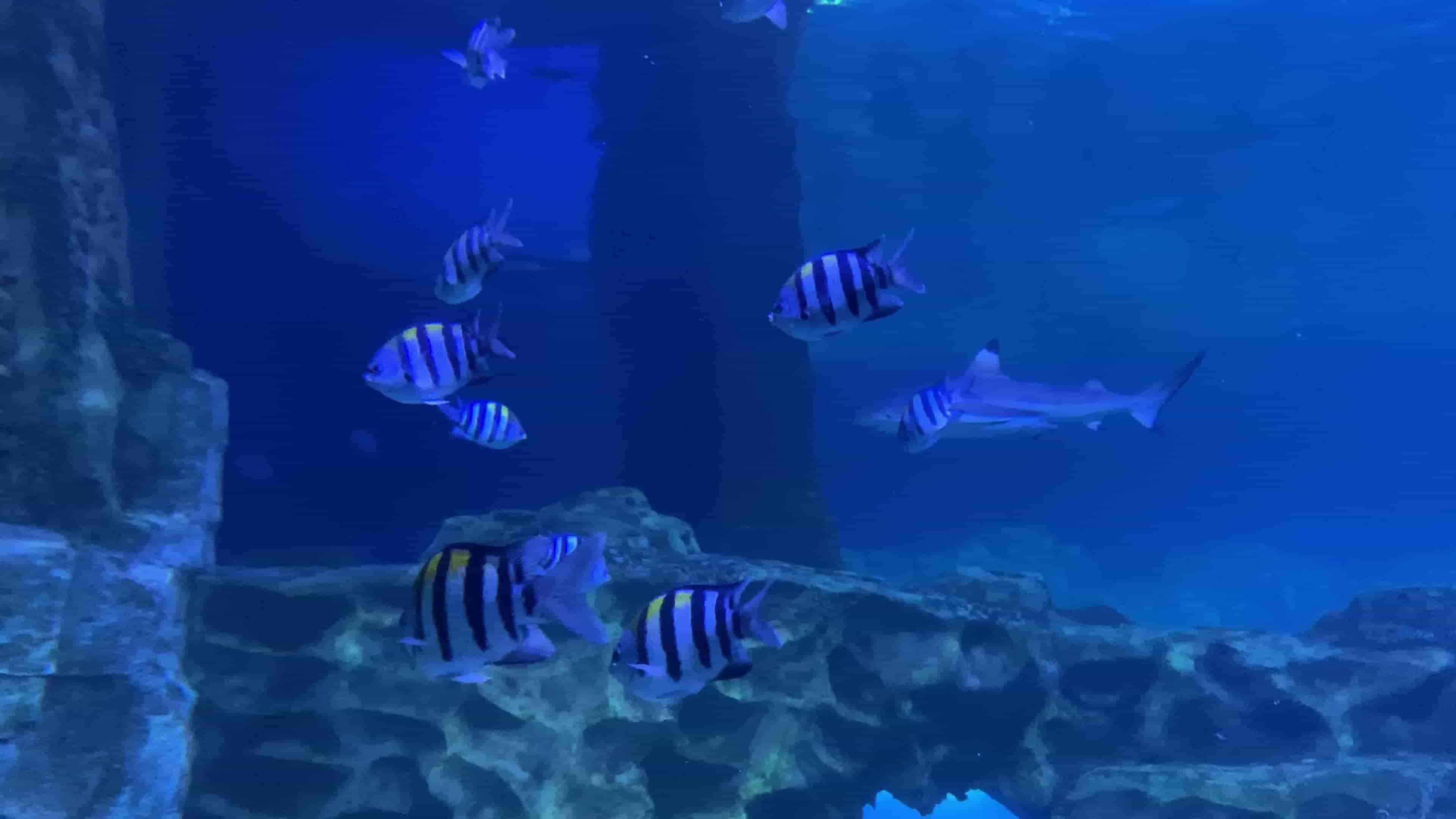}
    \begin{picture}(-10,-10)
        \put(-26,26){\textcolor{white}{48.87}}
    \end{picture}
\end{minipage}
\begin{minipage}[b]{0.1185\linewidth}
    \includegraphics[width=\linewidth]{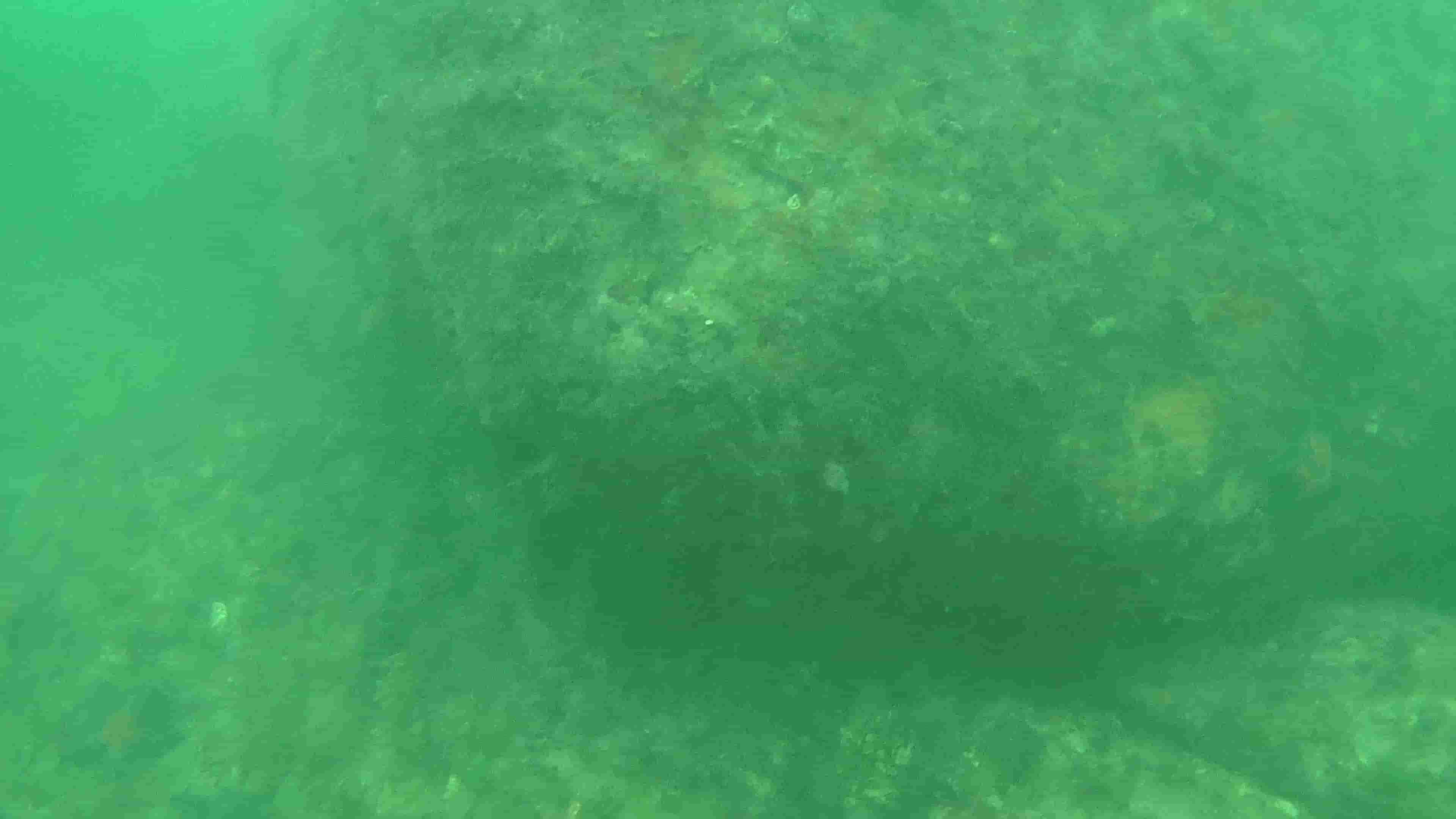}
    \begin{picture}(-10,-10)
        \put(-26,26){\textcolor{white}{19.13}}
    \end{picture}
\end{minipage}
\begin{minipage}[b]{0.1185\linewidth}
    \includegraphics[width=\linewidth]{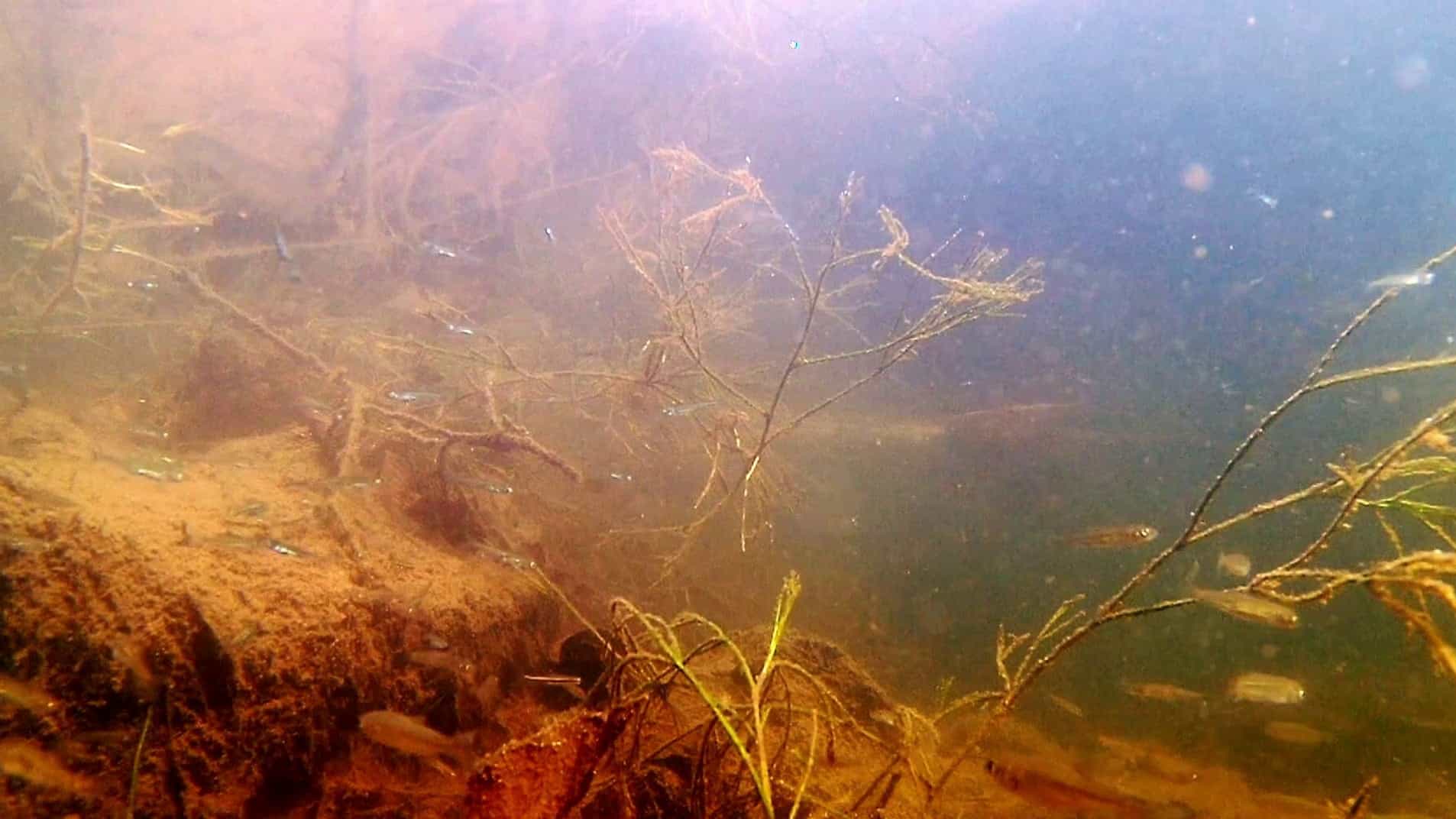}
    \begin{picture}(-10,-10)
        \put(-26,26){\textcolor{white}{30.87}}
    \end{picture}
\end{minipage}
\begin{minipage}[b]{0.1185\linewidth}
    \includegraphics[width=\linewidth]{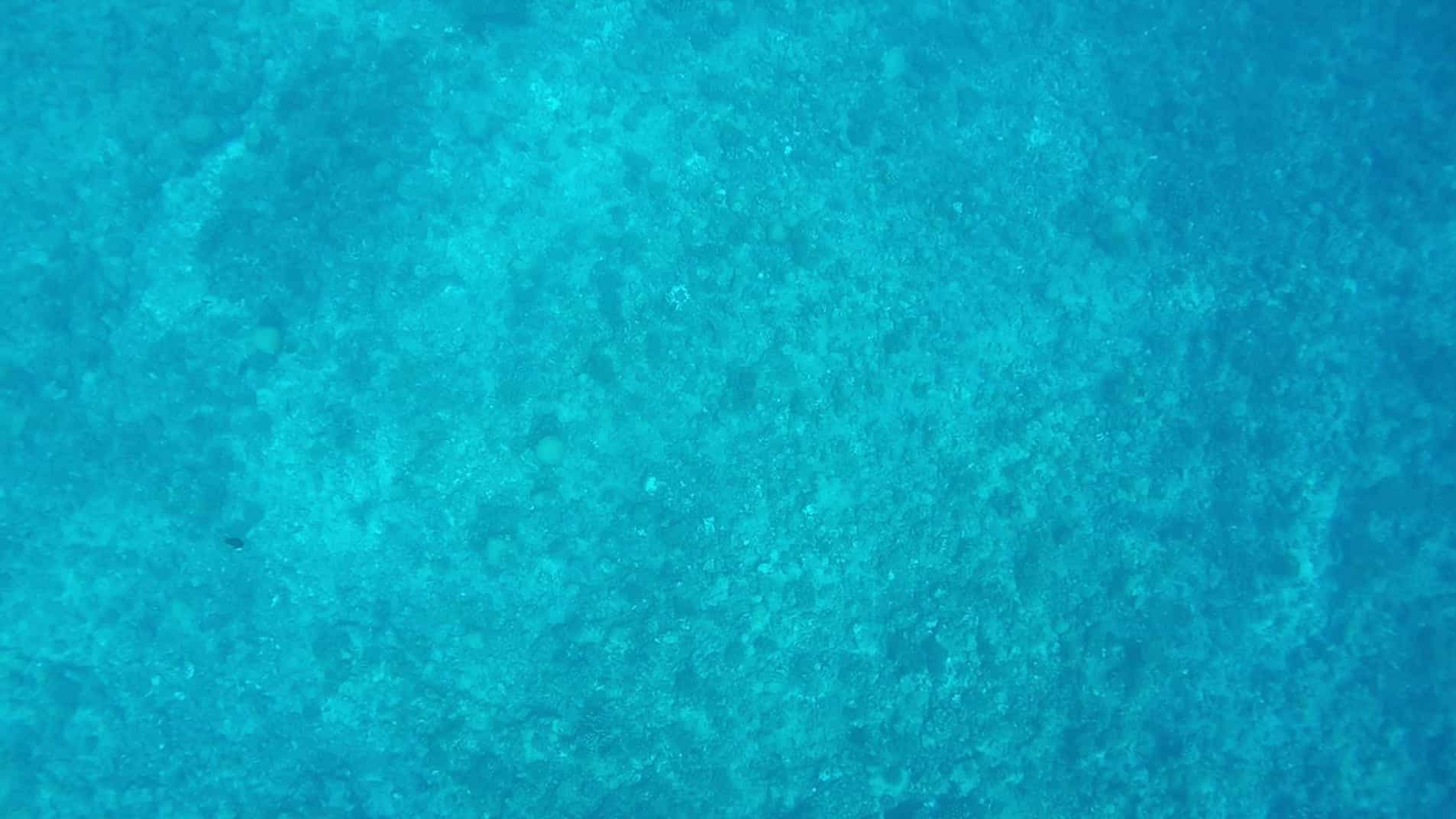}
    \begin{picture}(-10,-10)
        \put(-26,26){\textcolor{white}{35.00}}
    \end{picture}
\end{minipage}
\begin{minipage}[b]{0.1185\linewidth}
    \includegraphics[width=\linewidth]{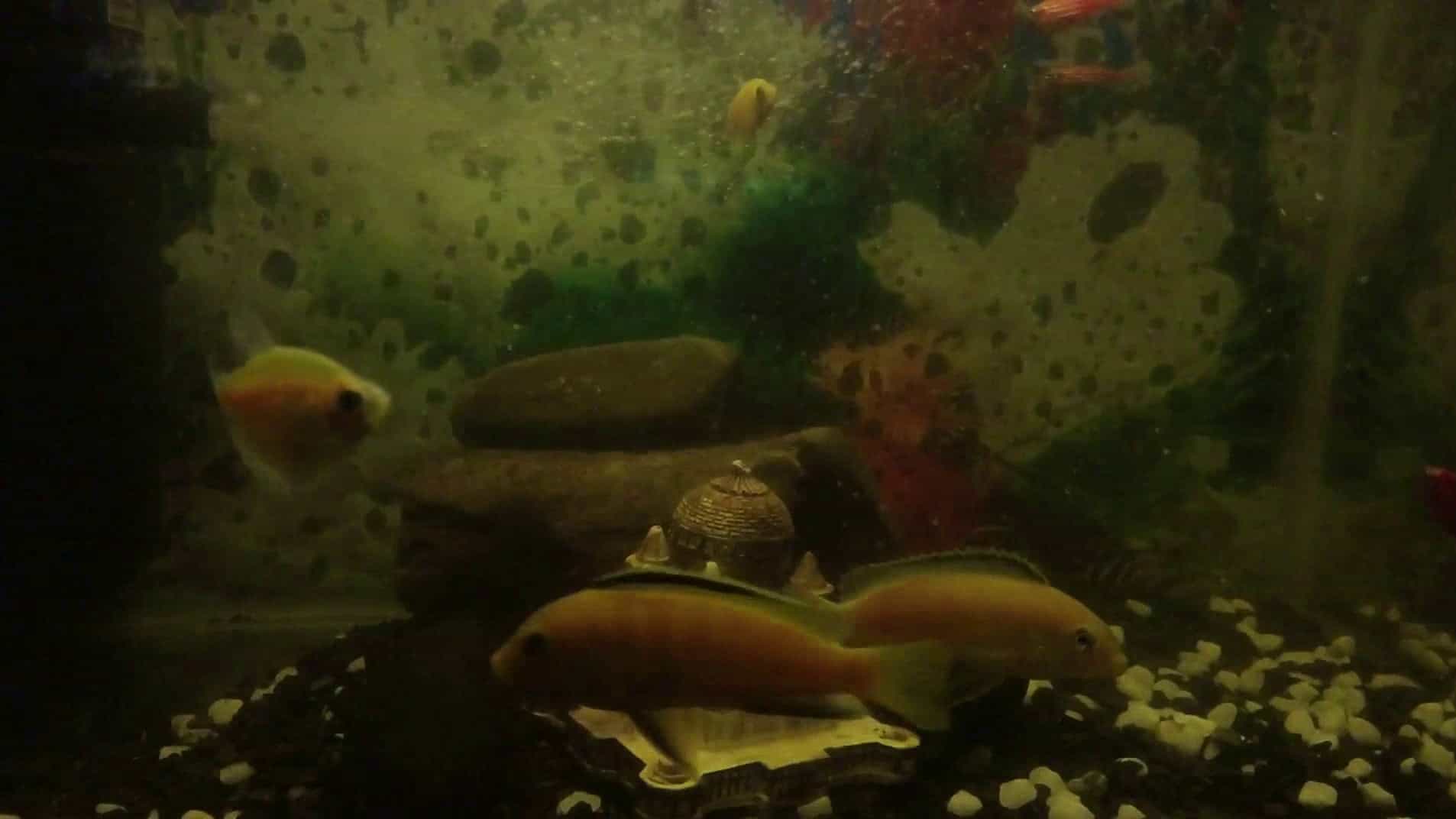}
    \begin{picture}(-10,-10)
        \put(-26,26){\textcolor{white}{31.27}}
    \end{picture}
\end{minipage}
\rotatebox{90}{\scriptsize{~~~~~~~~~~~~~GT}}
\begin{minipage}[b]{0.1185\linewidth}
    \includegraphics[width=\linewidth]{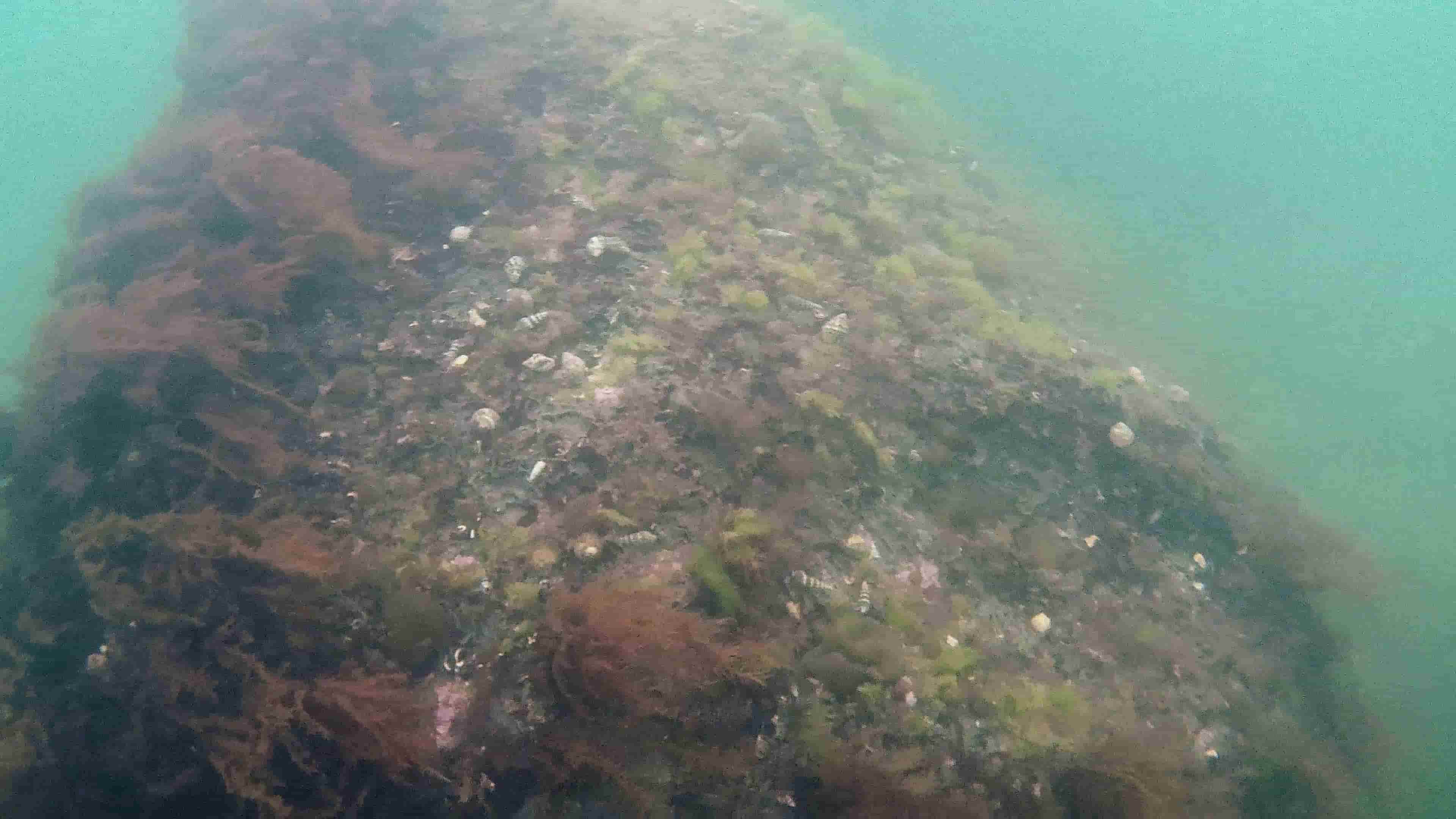}
    \begin{picture}(-10,-10)
        \put(-26,26){\textcolor{white}{33.16}}
    \end{picture}
    \subcaption*{green}
\end{minipage}
\begin{minipage}[b]{0.1185\linewidth}
    \includegraphics[width=\linewidth]{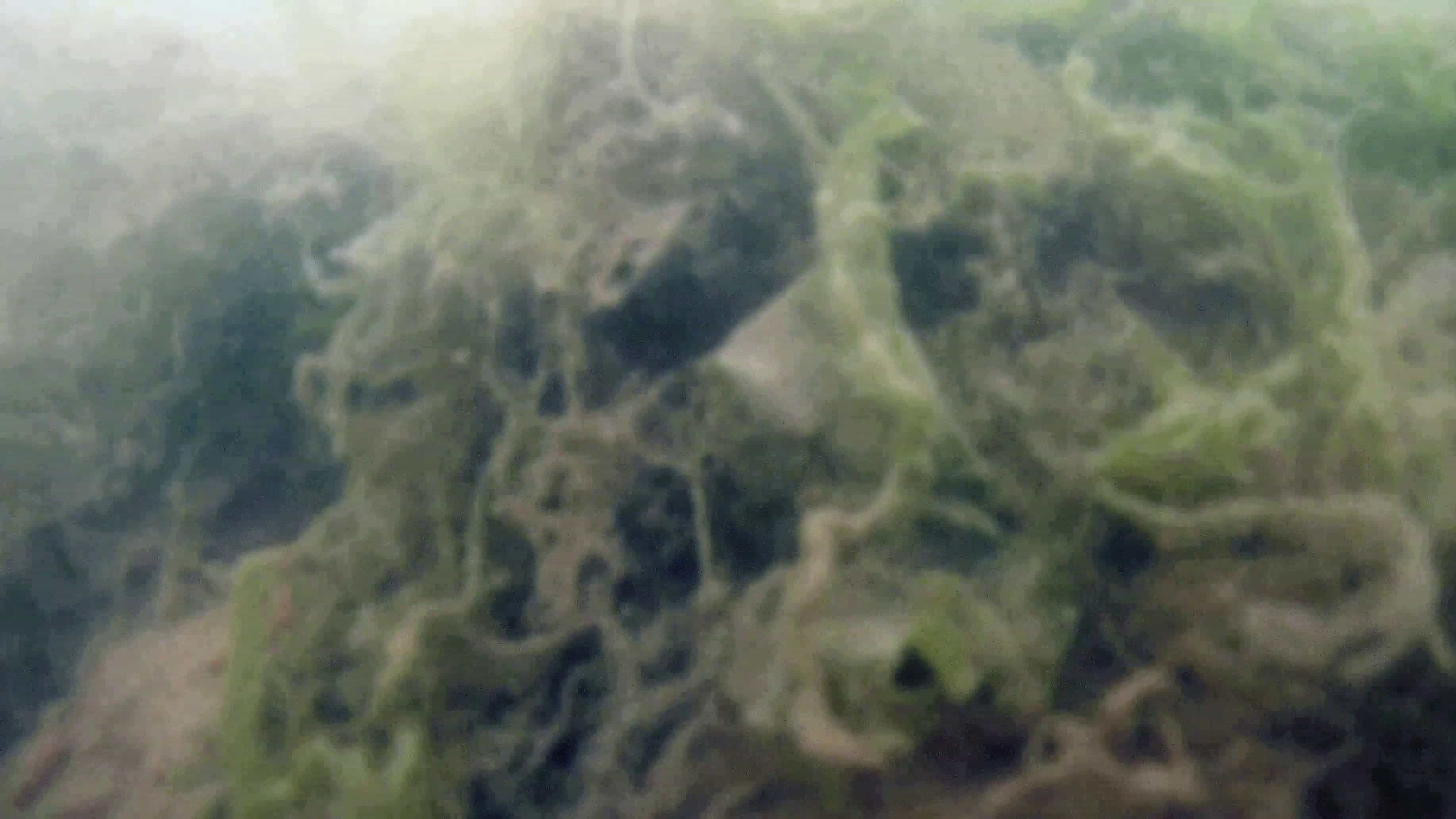}
    \begin{picture}(-10,-10)
        \put(-26,26){\textcolor{white}{22.39}}
    \end{picture}
    \subcaption*{yellow}
\end{minipage}
\begin{minipage}[b]{0.1185\linewidth}
    \includegraphics[width=\linewidth]{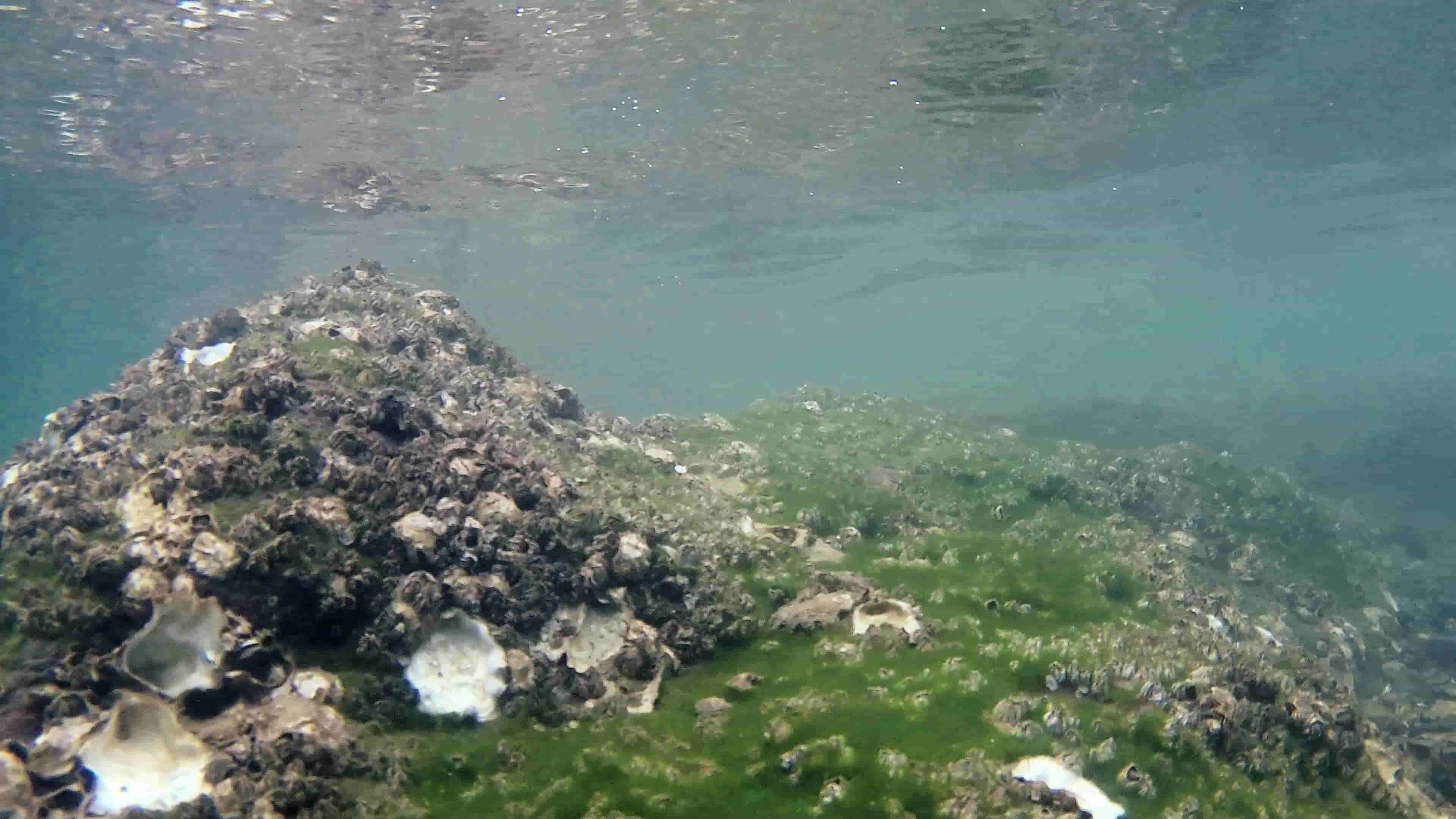}
    \begin{picture}(-10,-10)
        \put(-26,26){\textcolor{white}{46.96}}
    \end{picture}
    \subcaption*{white}
\end{minipage}
\begin{minipage}[b]{0.1185\linewidth}
    \includegraphics[width=\linewidth]{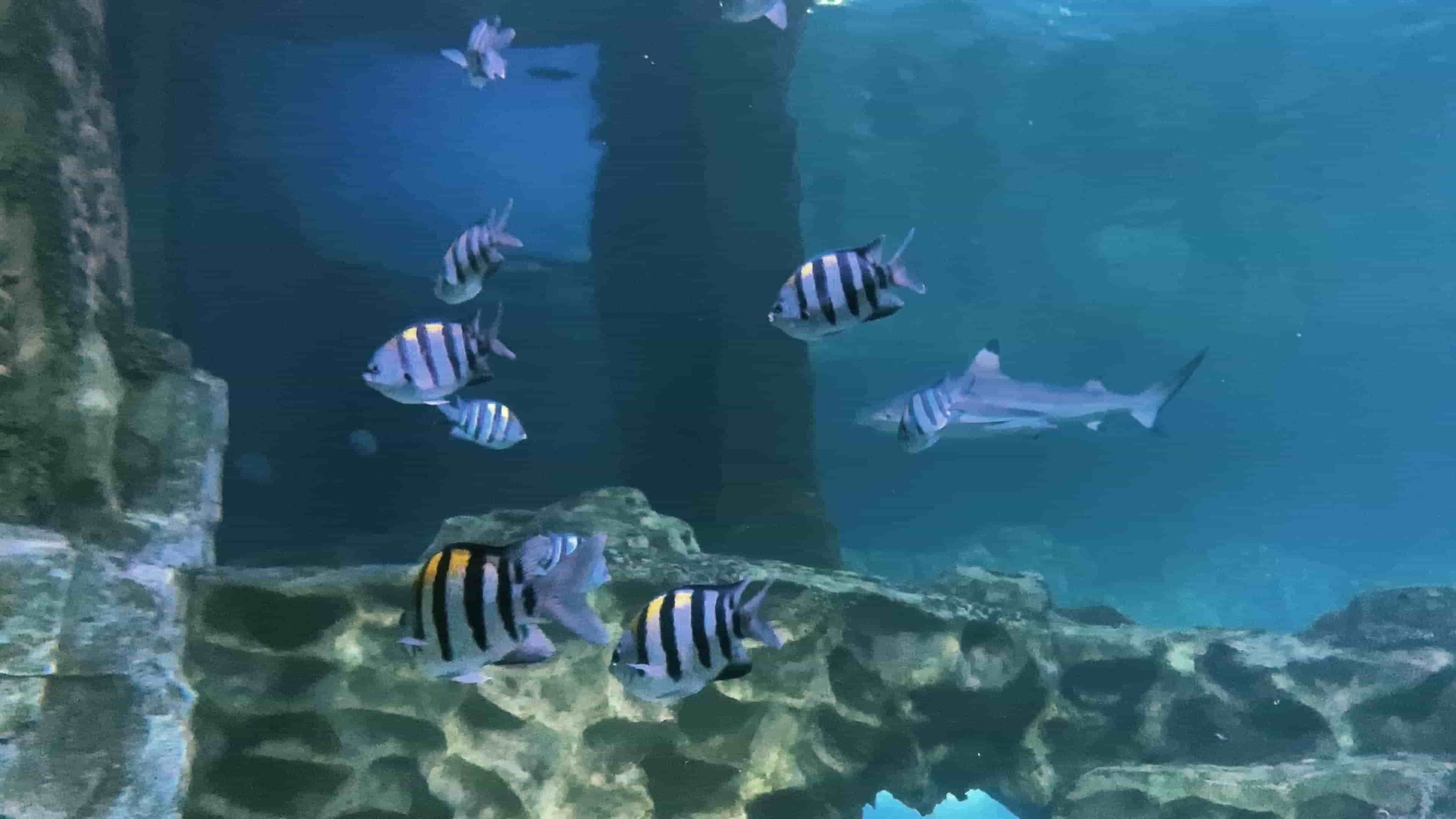}
    \begin{picture}(-10,-10)
        \put(-26,26){\textcolor{white}{59.64}}
    \end{picture}
    \subcaption*{blue}
\end{minipage}
\begin{minipage}[b]{0.1185\linewidth}
    \includegraphics[width=\linewidth]{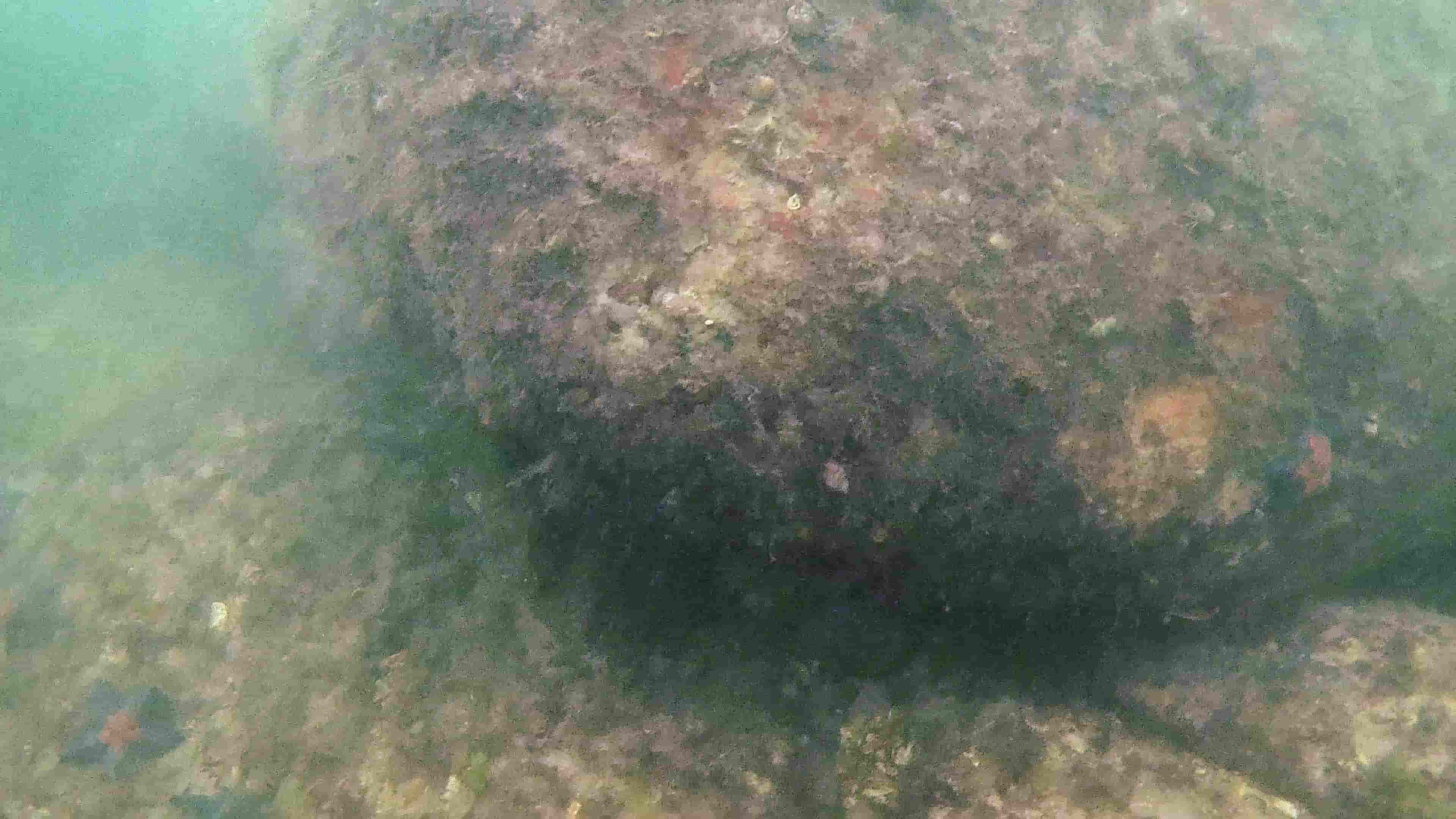}
    \begin{picture}(-10,-10)
        \put(-26,26){\textcolor{white}{29.55}}
    \end{picture}
    \subcaption*{green}
\end{minipage}
\begin{minipage}[b]{0.1185\linewidth}
    \includegraphics[width=\linewidth]{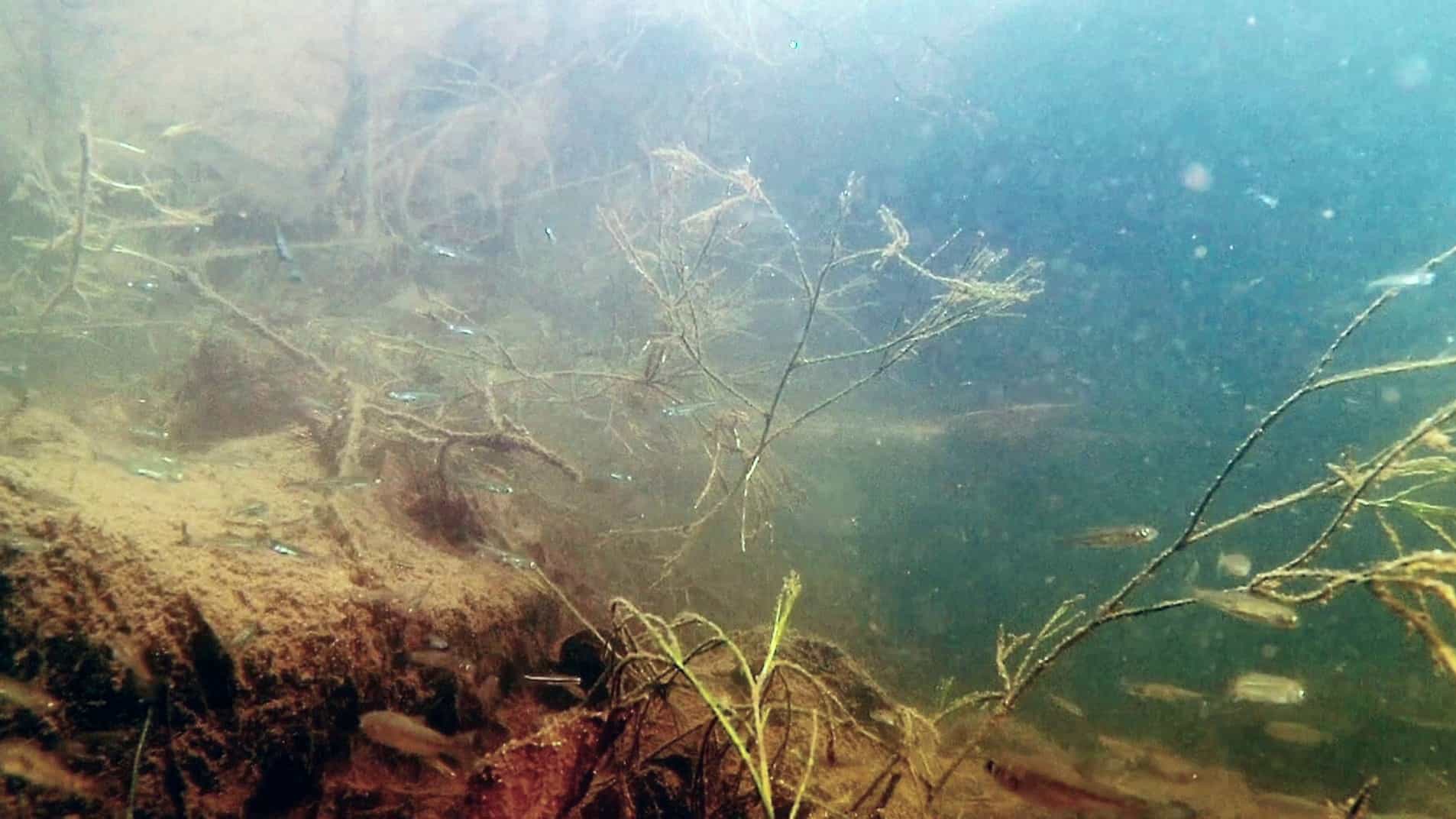}
    \begin{picture}(-10,-10)
        \put(-26,26){\textcolor{white}{37.03}}
    \end{picture}
    \subcaption*{other}
\end{minipage}
\begin{minipage}[b]{0.1185\linewidth}
    \includegraphics[width=\linewidth]{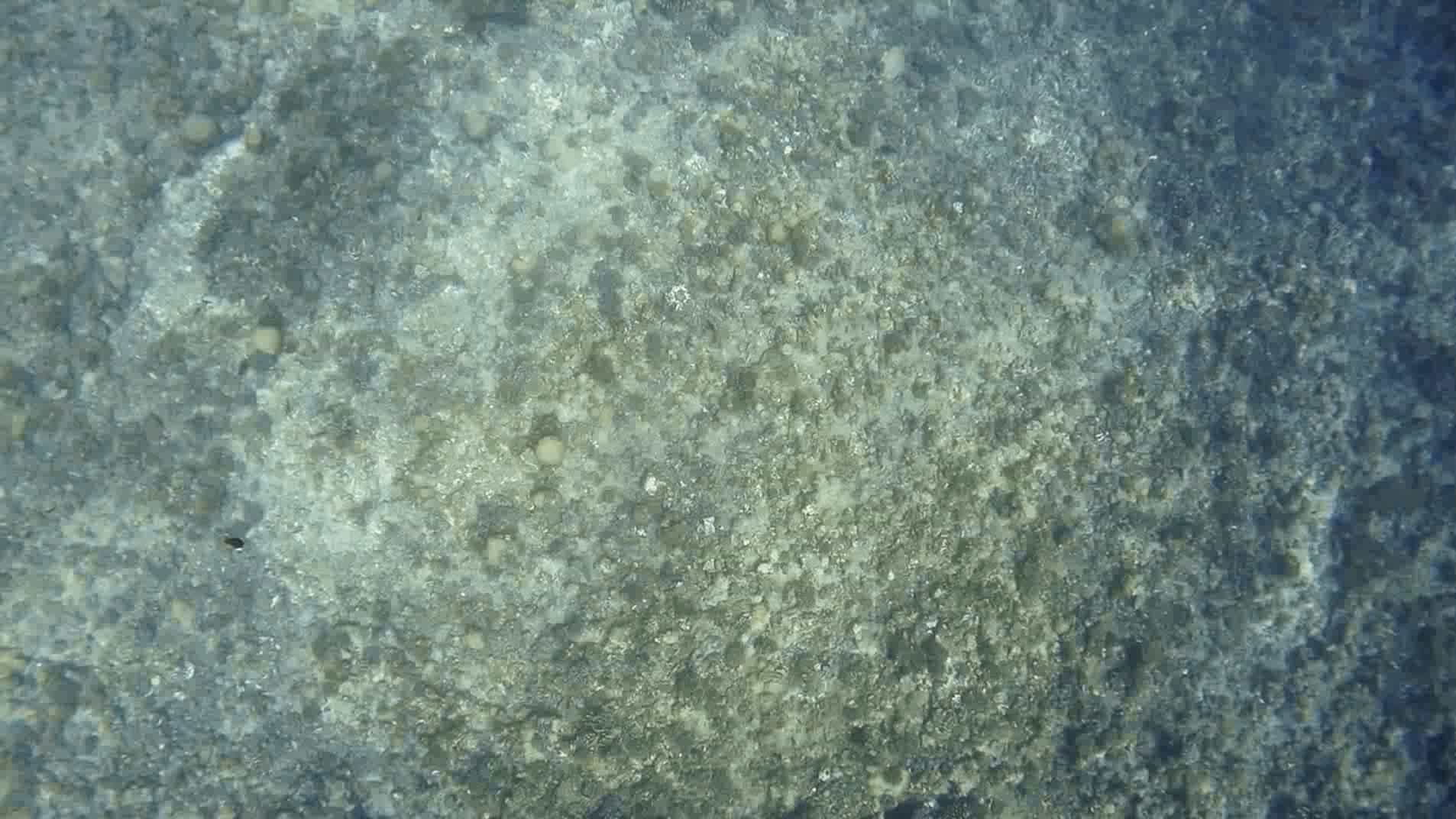}
    \begin{picture}(-10,-10)
        \put(-26,26){\textcolor{white}{42.12}}
    \end{picture}
    \subcaption*{blue}
\end{minipage}
\begin{minipage}[b]{0.1185\linewidth}
    \includegraphics[width=\linewidth]{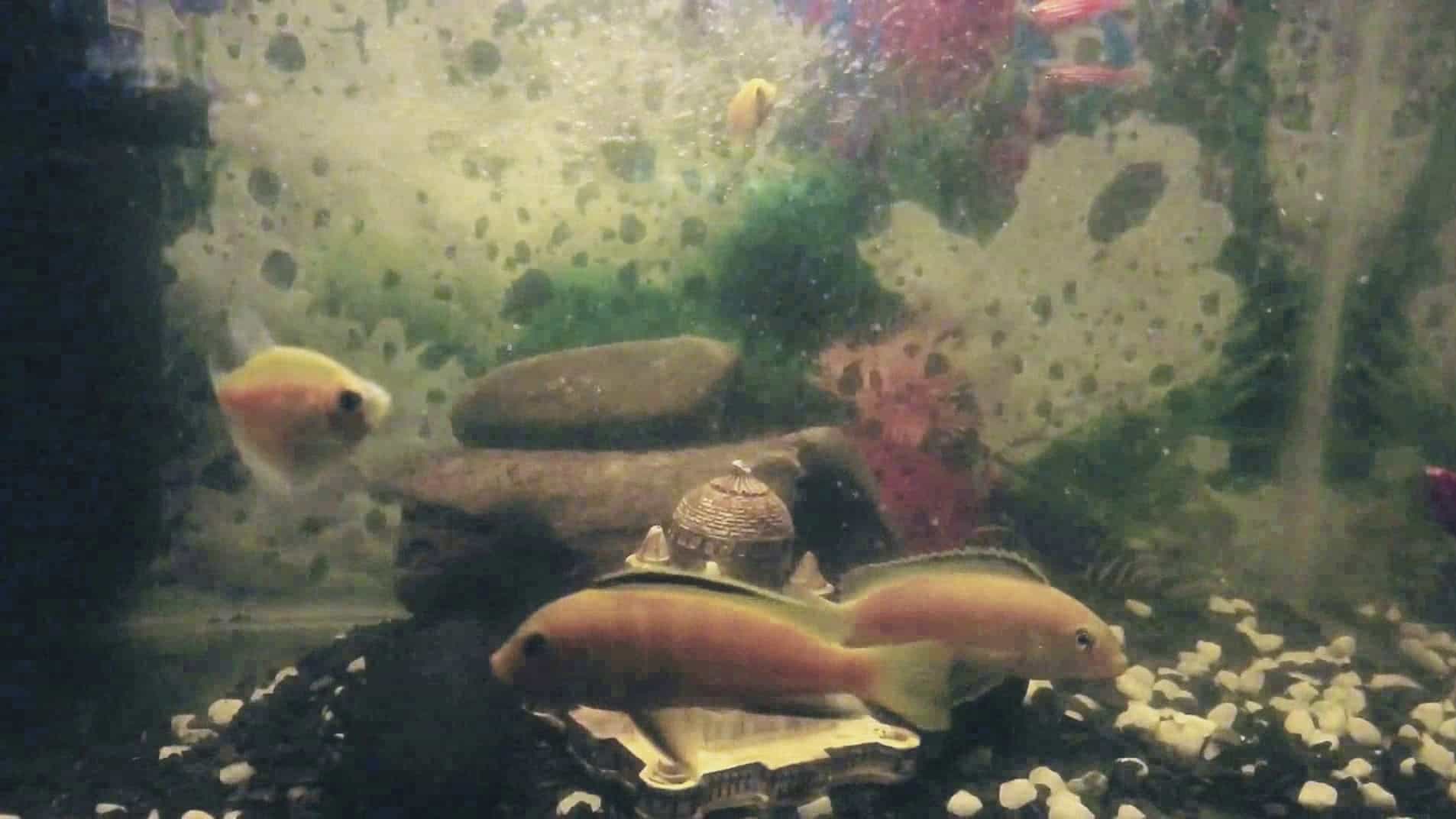}
    \begin{picture}(-10,-10)
        \put(-26,26){\textcolor{white}{35.98}}
    \end{picture}
    \subcaption*{insufficient light}
\end{minipage}
    \caption{The diversity of UVEB samples with data from different color distortions in multiple scenes. Each column provides the water color deviation, raw video quality score, and GT quality score.}
    \label{fig:diversity}
\end{figure*}
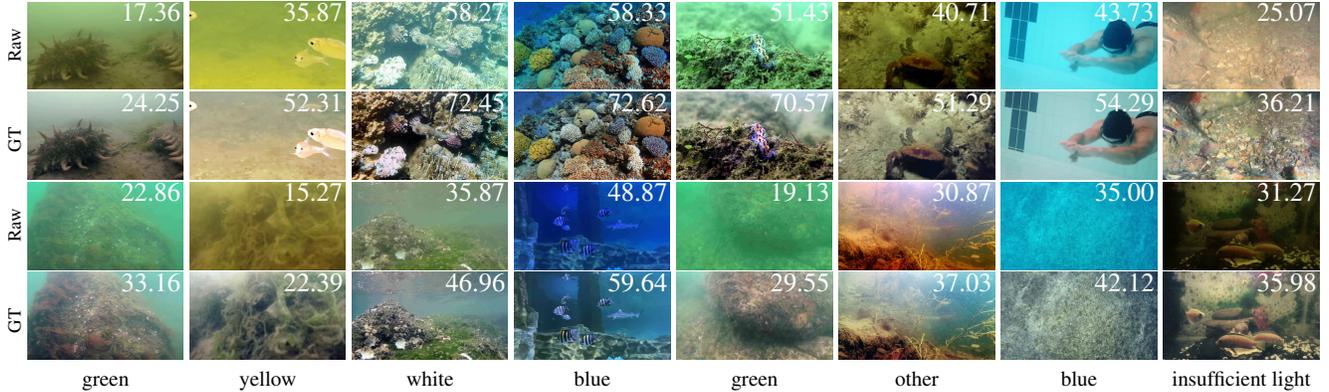
\begin{figure*}[htbp]
    \centering
\begin{minipage}[b]{0.39\linewidth}
    \includegraphics[width=\linewidth]{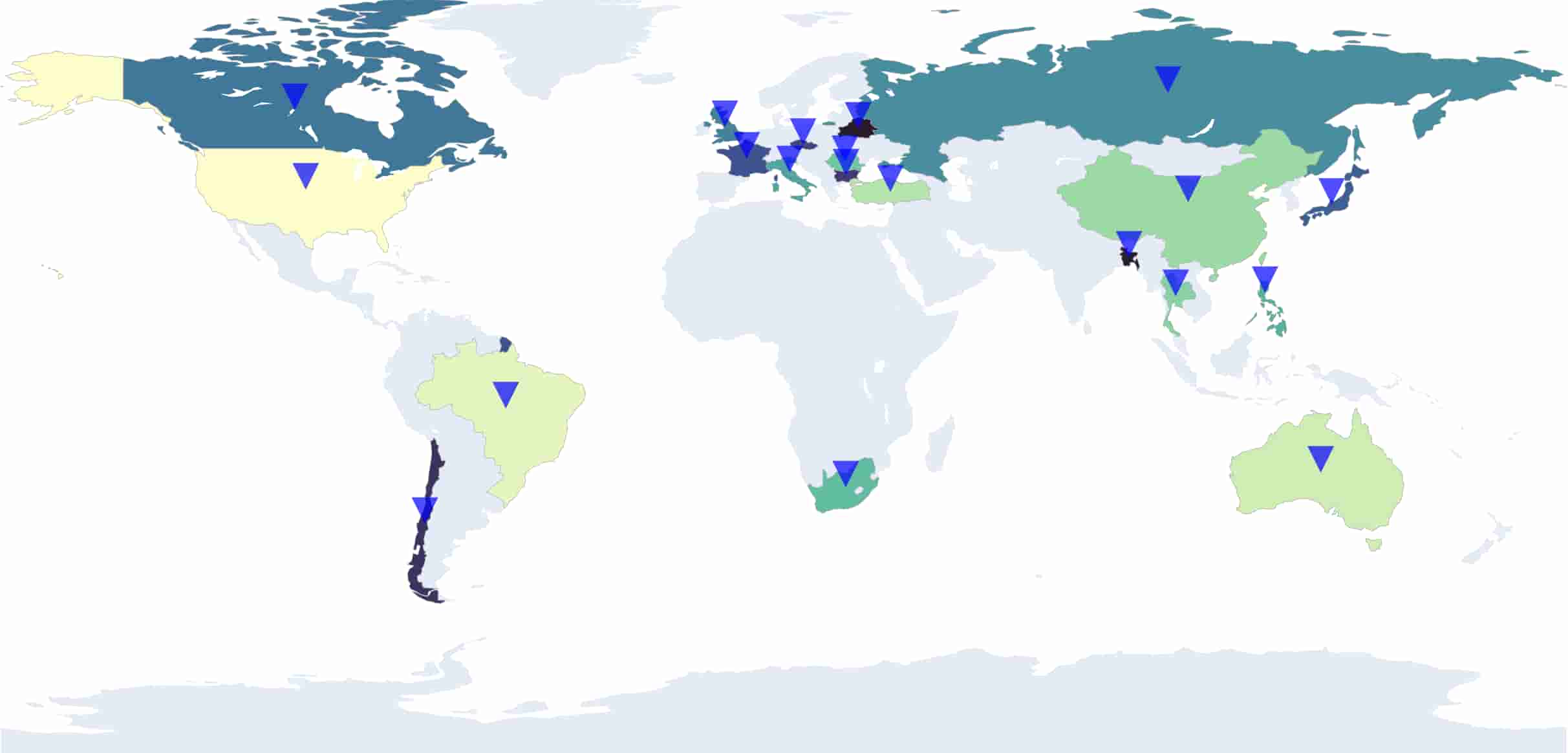}
    \subcaption{Distribution}
    \label{fig:distribution}
\end{minipage}
\begin{minipage}[b]{0.29\linewidth}
    \includegraphics[width=\linewidth]{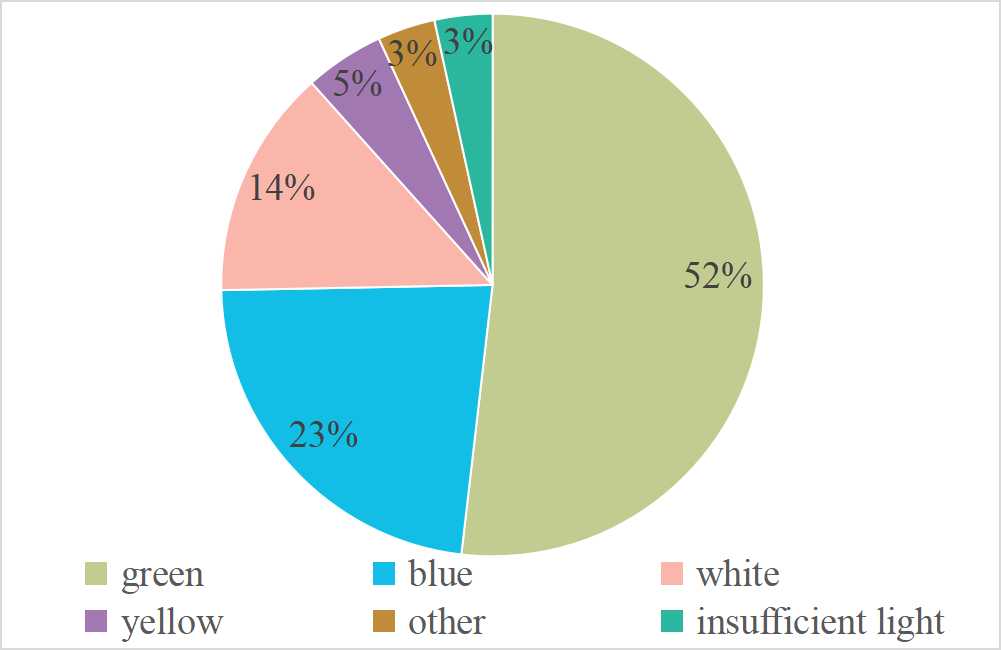}
    \subcaption{Degradation types}
    \label{fig:classify}
\end{minipage}
\begin{minipage}[b]{0.29\linewidth}
    \includegraphics[width=\linewidth]{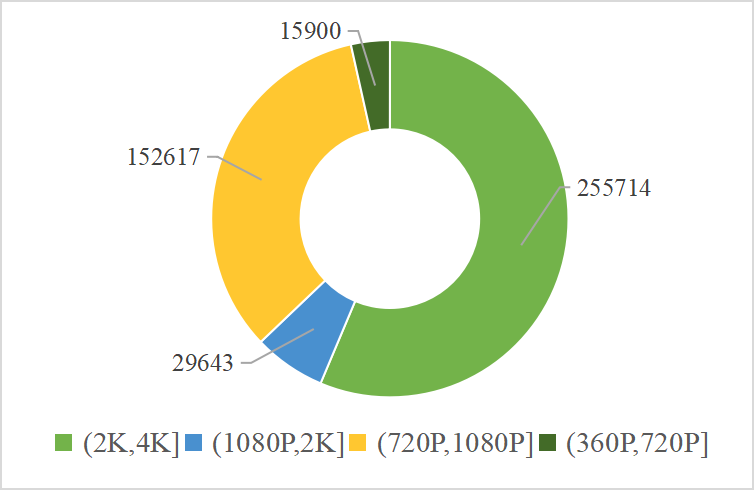}
    \subcaption{Resolution}
    \label{fig:resolution}
\end{minipage}
    \caption{(a) The spatial distribution of videos collected in our dataset. (b) The proportion of six types of underwater video degradation in UVEB. (c)The resolution information of UVEB.}
    \label{fig:2}
\end{figure*}

\hspace{0.16667in}We use FIFISH V6 and FIFISH V-EVO equipped with 4K resolution cameras to collect underwater videos from multiple sea areas and ports in China. We also collect internet underwater videos shared by underwater photographers from many countries to enrich our dataset. 
Due to the difficulty in obtaining clear underwater GT images, the existing real-world UIE datasets UIEB~\cite{li2019underwater} and LSUI~\cite{peng2023u} utilize 12 and 18 methods to enhance the raw images and select the best results from the enhancement results as GT. Practices \cite{fu2022uncertainty,huang2023contrastive,liu2022adaptive,guan2023fast} have proven that this strategy is currently the best way to build paired UIE datasets. We follow this strategy to construct the UVEB dataset.

We select 20 UIE methods (including 10 methods published in the last two years) that can process underwater videos of different sizes to enhance the raw videos and obtain GT. After processing more than 9,000,000 high-resolution frames, we obtained 20 enhancement results of the raw videos. We evaluate the quality score of each enhancement result and choose the optimal enhancement result as the GT. We also provide video quality scores for raw video and GT as supplementary information in the UVEB dataset. The existing UIE methods for both quality assessment and quality enhancement \cite{guo2023underwater,huang2023contrastive} make us believe that future research can utilize the sample quality information for better underwater video enhancement.

\subsection{Labeled Sample Generation}
\label{sec:Labeled Sample Generation}
\begin{table}
\caption{Total score of all methods on the test videos.}
\centering
    \begin{tabular}{cc|cc}
    \toprule
    Method & score & Method & score \\
    \midrule
    PUIE~\cite{fu2022uncertainty} & 60071 & CLUIE~\cite{li2022beyond} & 42376\\
    LANet~\cite{liu2022adaptive} & 59779 & fusion-based~\cite{ancuti2012enhancing} & 40145\\
    CLAHE~\cite{10.5555/180895.180940} & 53196 & MSCNN~\cite{ren2016single} & 38208\\
    FA$^{+}$Net~\cite{jiang2023five} & 49815 & WWPF~\cite{zhang2023underwater} & 32689\\
    URanker~\cite{guo2023underwater} & 49277 & retinex-based~\cite{fu2014retinex} &30894 \\
    FspiralGAN~\cite{guan2023fast} & 48633 & GDCP~\cite{peng2018generalization} & 29547\\
    GC~\cite{schlick1995quantization} & 47975 & HE~\cite{hummel1975image} & 29309\\
    USUIR~\cite{fu2022unsupervised} & 46308 & UDCP~\cite{drews2016underwater} & 19195\\
    MLLE~\cite{zhang2022underwater} & 43745 & MetaUE~\cite{zhang2023metaue} & 19088\\
    Red Channel~\cite{galdran2015automatic} & 43541 & DCP~\cite{5206515} & 13647\\
    \bottomrule
  \end{tabular}
 
  \label{tab:score}
\end{table}

\noindent{\bf Annotation Preparation.}
Selecting the optimal enhancement results involve video quality assessment, thus the whole process is performed under the guidance of ITU-R BT500-13~\cite{series2012methodology} with 15 observers. All observers conducted video quality assessment using the Redmi-27H 4K display under the same experiment setting. Observers can rate the video quality with an integer from 0 to 100, similar to the task setting in ~\cite{gao2023vdpve}. 
Each Observer underwent two days of professional knowledge training to fully understand the physical process and common types of underwater image degradation.

\noindent{\bf Rigorous and Reasonable Assessment Process.}
Different from~\cite{gao2023vdpve,wang2022generation}, underwater video enhancement quality assessment is much more difficult due to the diversity and complexity of video degradation types. Thus, we adopt a more rigorous approach for video quality assessment.


We select 1743 videos (83$\times$21) covering various scenarios and degradation types for assessment test. Although the types and degrees of degradation are diverse and each observer's visual perception is different, each observer's rating of the same data should be stable and reasonably increase with increasing quality. To ensure this point, we prepared 150 videos with various types of videos to construct the example library. We asked observers to score and sort the 150 videos in increasing order of quality. The sorted example libraries obtained by each observer through the process were used as respective video quality scales. Observers could view their scales if they needed reference on their ratings and watch the videos many times to give more definitive ratings. To avoid visual fatigue, observers took a mandatory half-hour break after a half-hour evaluation, and the labeling task was allowed to be completed within 30 days.

\begin{figure}[htbp]
    \centering
    \includegraphics[width=\linewidth]{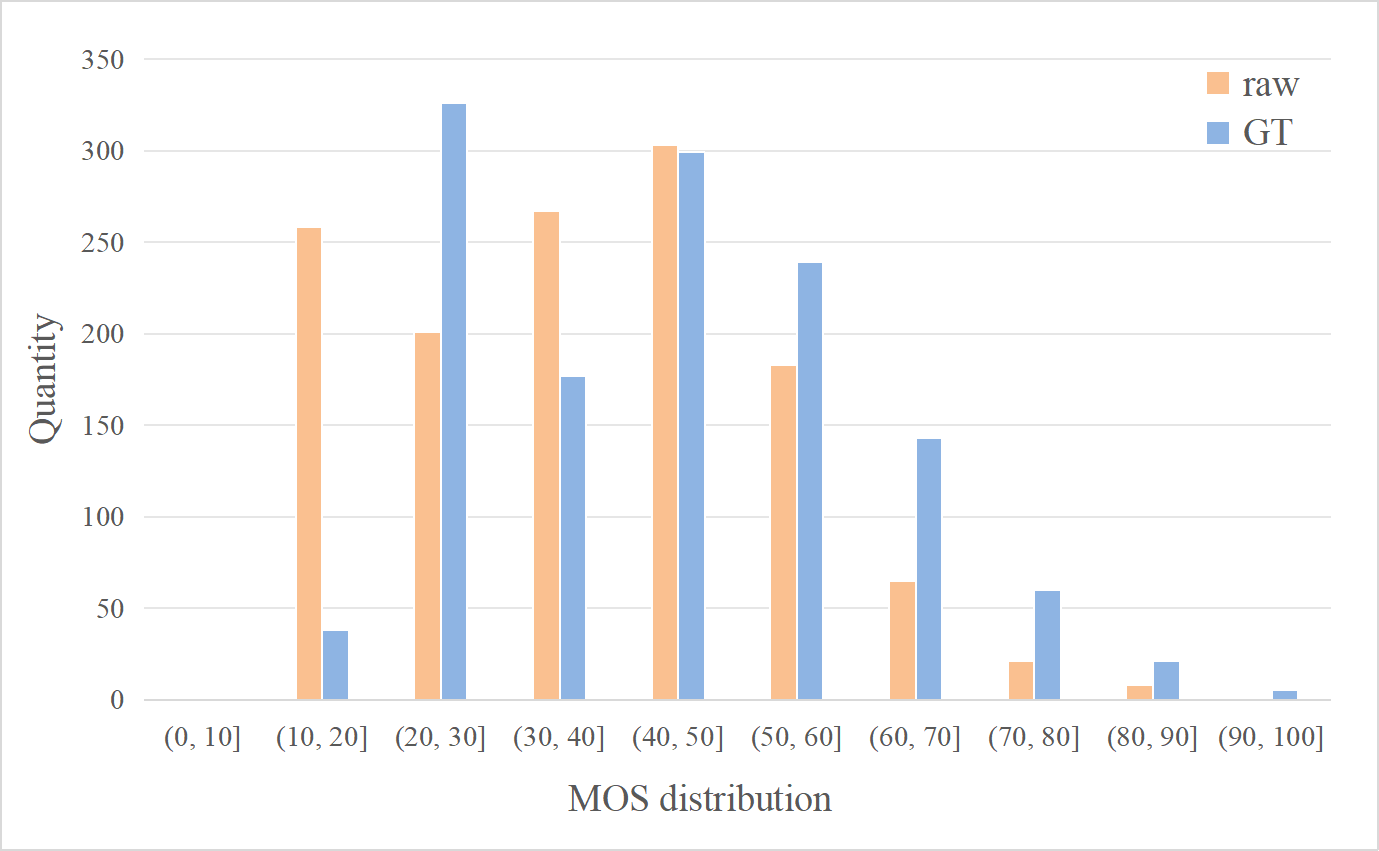}
    \caption{MOS of the samples before and after enhancement.}
    \label{fig:mos}
\end{figure}



\noindent{\bf Processing of Annotated Data}
For each set of videos, the ratings of observers that deviated by two standard deviations were eliminated based on all observers' ratings of the raw video. The remaining ratings were used to select GT. 
Based on the remaining ratings, we selected method $ M $, which received the most votes and the highest score. The remaining ratings of the raw video are averaged as the raw video quality score $ R_{q}$. 

The mean scores of the observers who chose $ M $ as the best enhancement method on the raw video and the enhanced result are $ S_{r}$ and $ S_{e}$, respectively. The difference between the two is  ${\Delta s}$. The GT sample score $ G_{q}$ is obtained by

\begin{equation}\label{1}
G_{q} = R_{q} + {\Delta s}
\end{equation}
where $R_{q}$ is given by more observers' evaluation than $ S_{r}$. Therefore, we set it as the raw video score. Since $ S_{e}$ and the raw video score $ R_{q}$ are given by inconsistent observer groups. Therefore, treating $ S_{e}$ as the quality score of GT is sufficiently credible. The quality improvement degree ${\Delta s}$ of the $ M $ method on the raw video is more reliable. Therefore, we follow the \cref{1} to obtain the quality score of GT. We also deleted 43 sets of samples with $ R_{q}$ less than 40 and enhancement degree ${\Delta s}$ less than 3, as the quality improvement of these samples was not significant.

\subsection{Data Analysis}

\begin{figure}[t]
    \centering
    \rotatebox{90}{\scriptsize{~~~~~~Raw}}
\begin{minipage}[b]{0.3165\linewidth}
    \subfloat{\includegraphics[width=\linewidth]{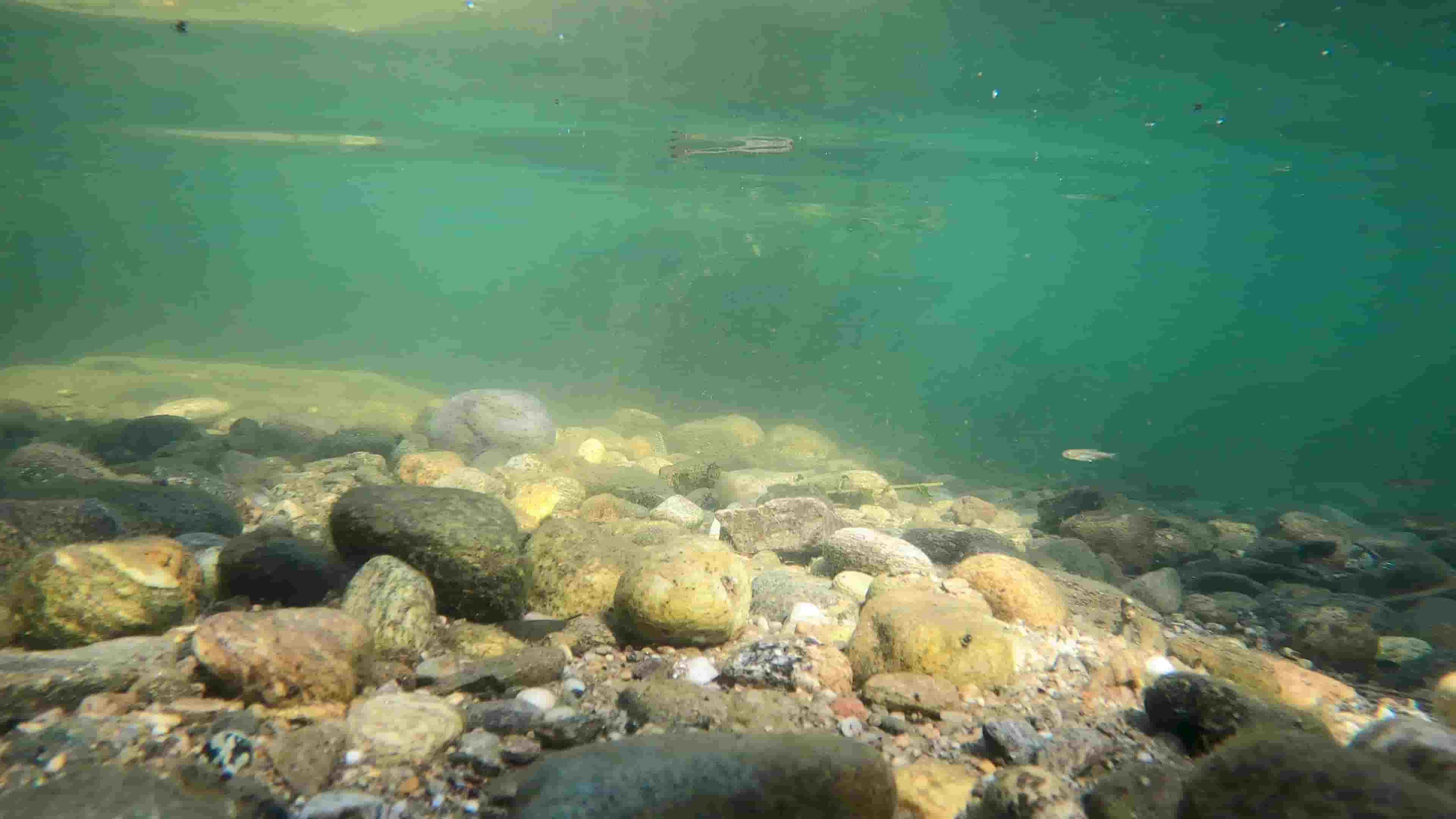}}
\end{minipage}
\begin{minipage}[b]{0.3165\linewidth}
    \subfloat{\includegraphics[width=\linewidth]{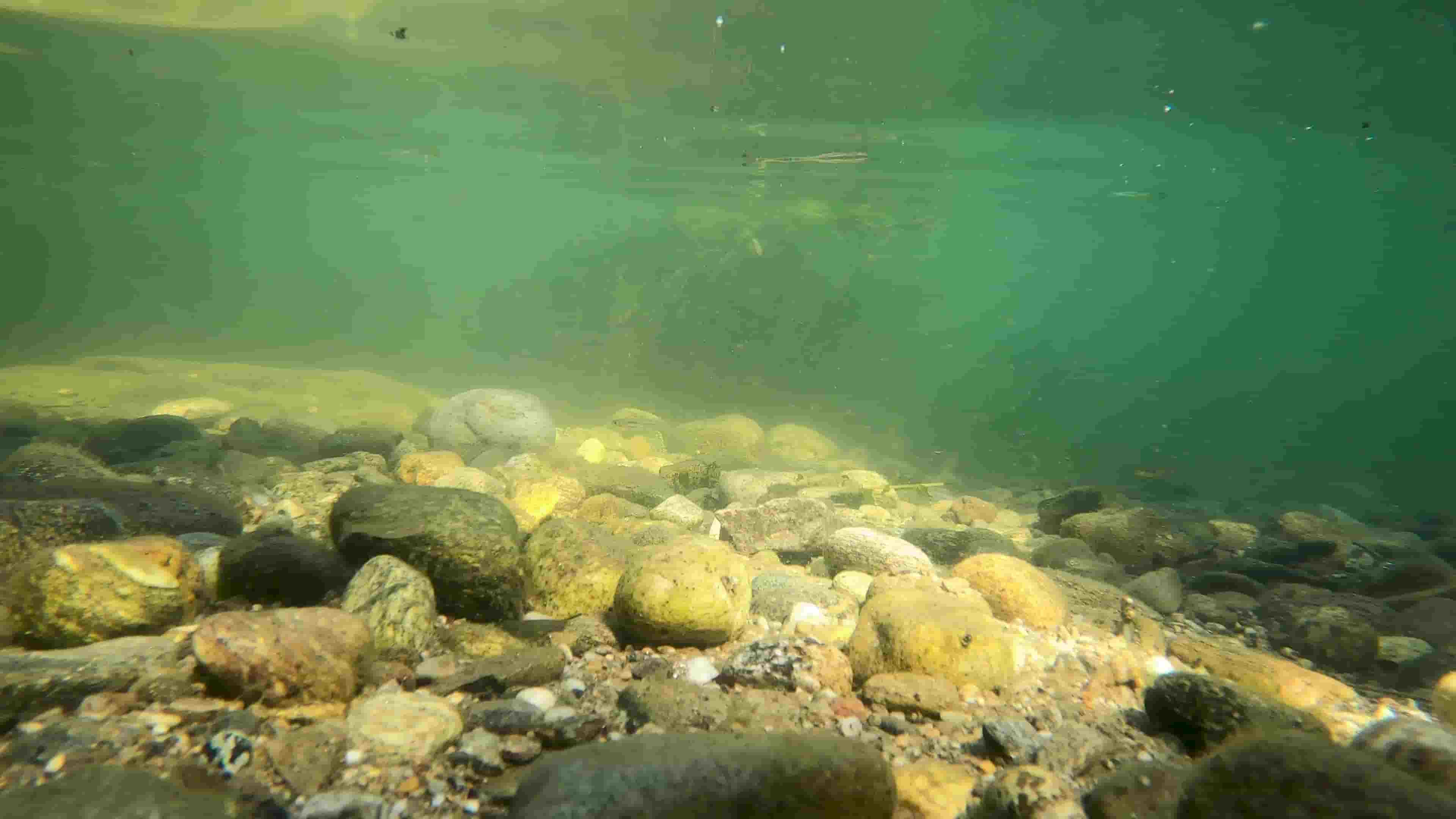}}
\end{minipage}
\begin{minipage}[b]{0.3165\linewidth}
    \subfloat{\includegraphics[width=\linewidth]{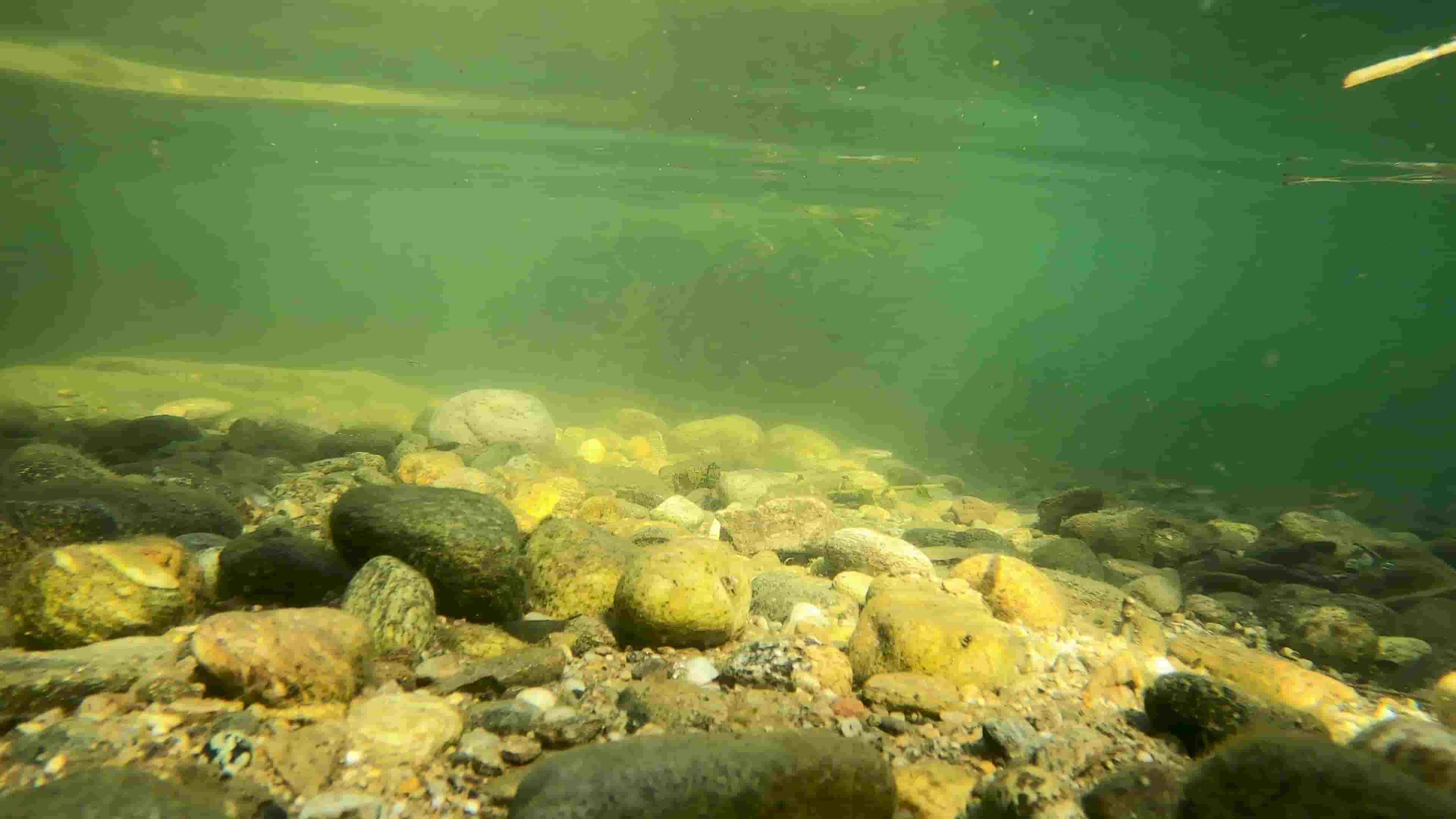}}
\end{minipage}
\vspace{0cm}
\rotatebox{90}{\scriptsize{~~~~~~~~~~~~~~GT}}
\begin{minipage}[b]{0.3165\linewidth}
    \subfloat{\includegraphics[width=\linewidth]{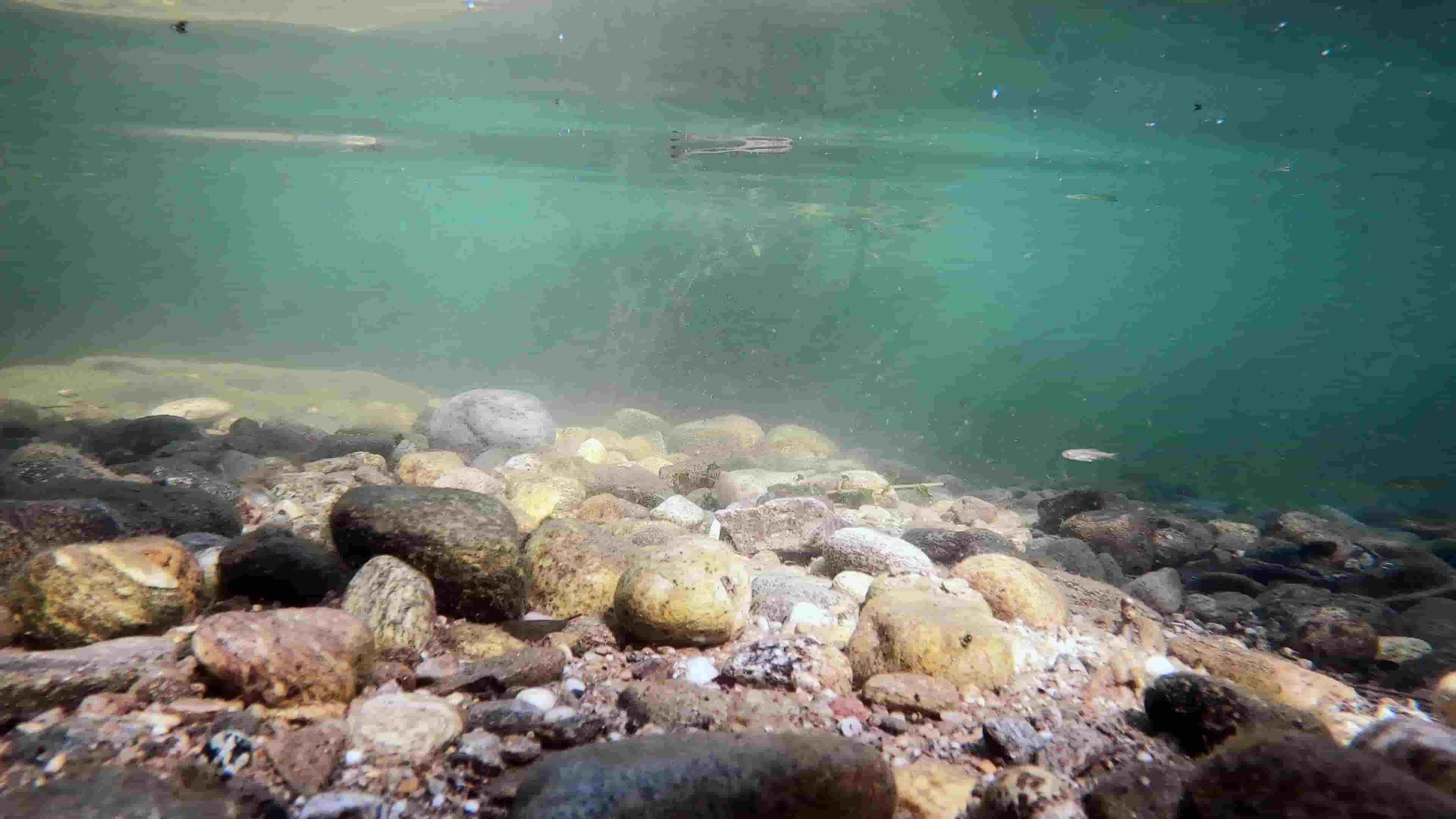}}
    \subcaption*{${1}^{st}$ frame}
\end{minipage}
\begin{minipage}[b]{0.3165\linewidth}
    \subfloat{\includegraphics[width=\linewidth]{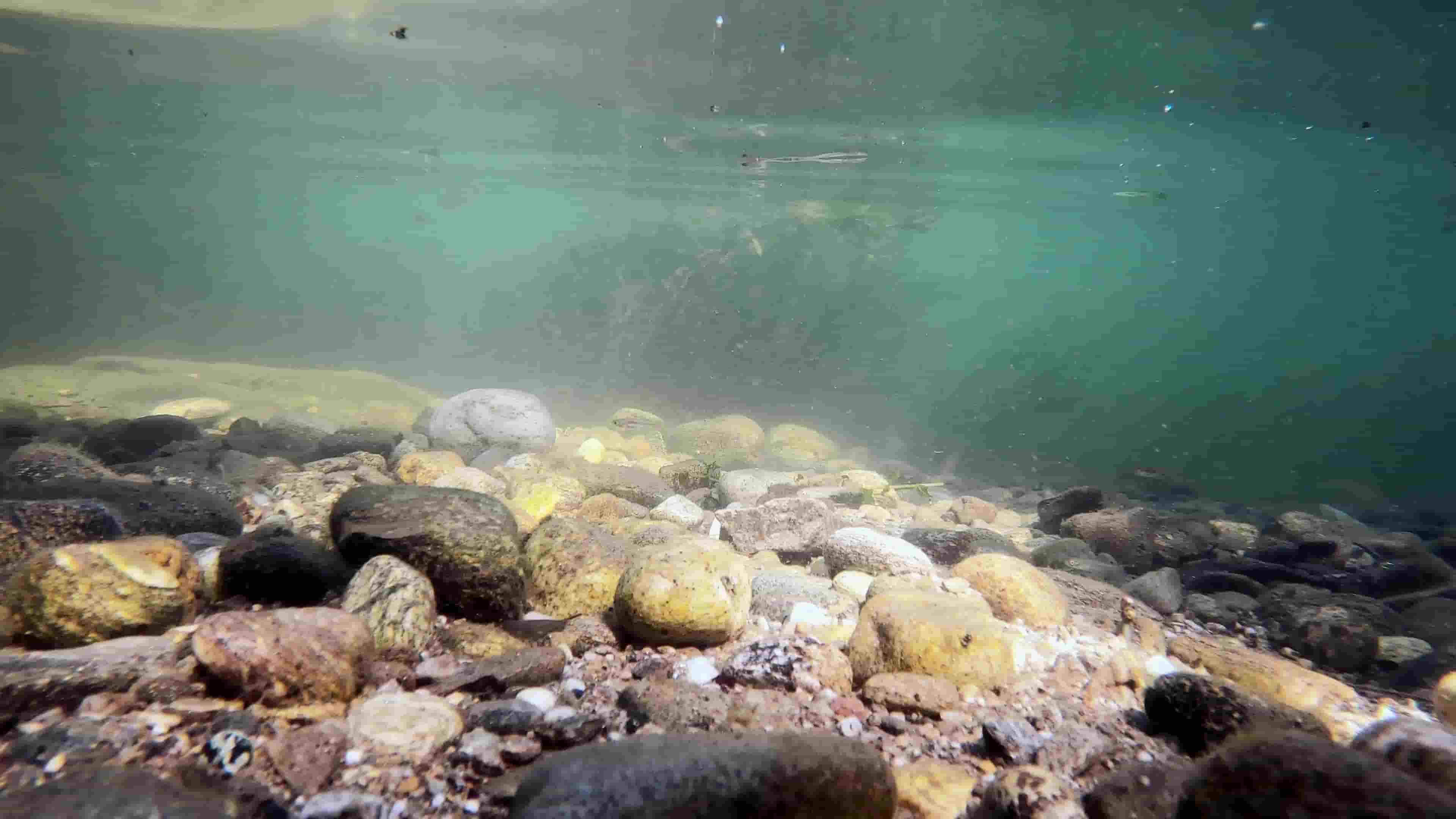}}
    \subcaption*{${90}^{th}$  frame}
\end{minipage}
\begin{minipage}[b]{0.3165\linewidth}
    \subfloat{\includegraphics[width=\linewidth]{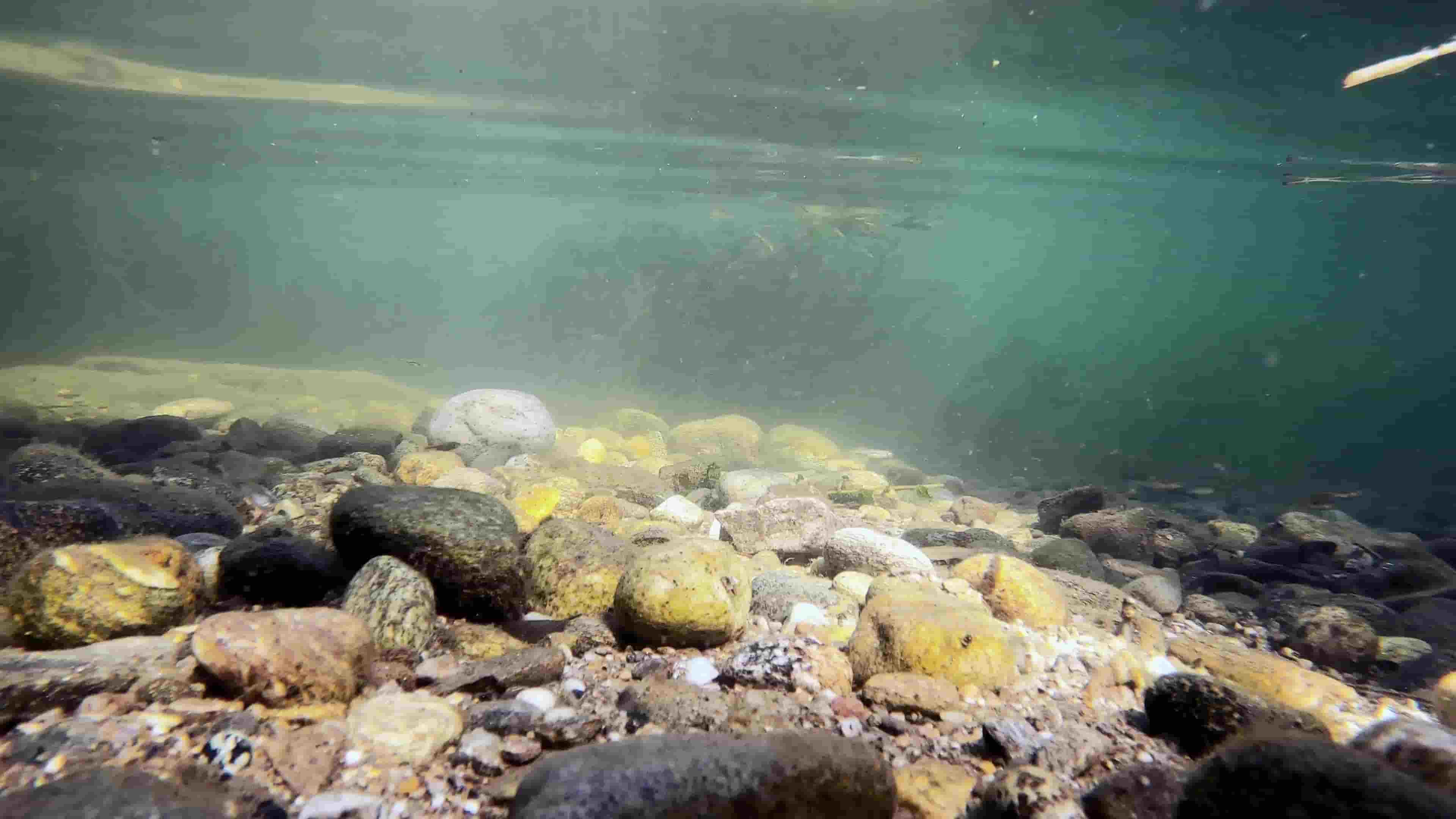}}
    \subcaption*{${600}^{th}$  frame}
\end{minipage}
    \caption{Color deviation changes with ambient light.}
    \label{fig:cc}
\end{figure}
\begin{figure}[t]
    \centering
    \begin{minipage}[b]{0.19\linewidth}
        \subfloat{\includegraphics[width=\linewidth]{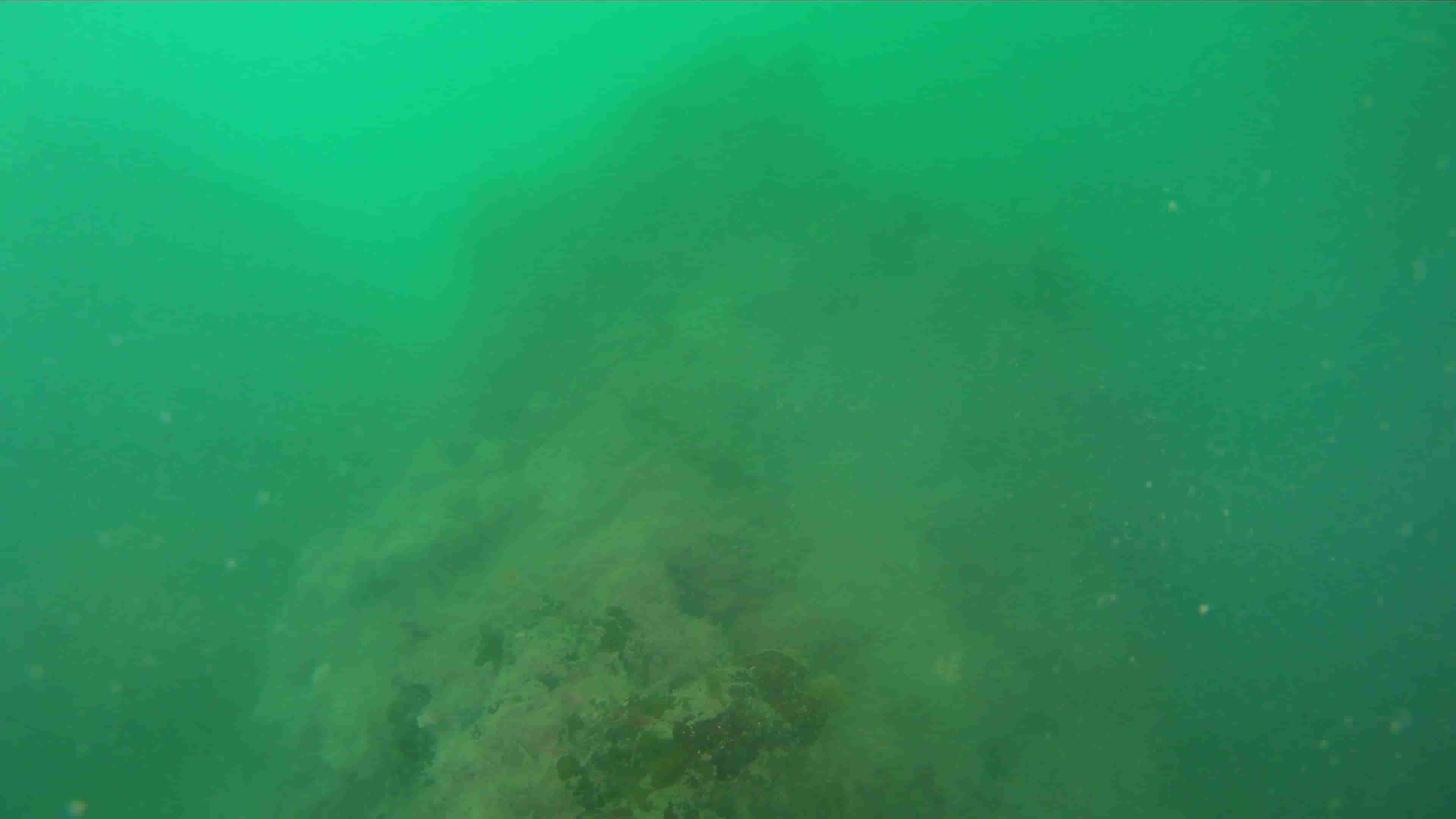}}
        \begin{picture}(0,0)
            \put(-18,19){\textcolor{white}{\footnotesize 15.50}}
        \end{picture}
    \end{minipage}
    \begin{minipage}[b]{0.19\linewidth}
        \subfloat{\includegraphics[width=\linewidth]{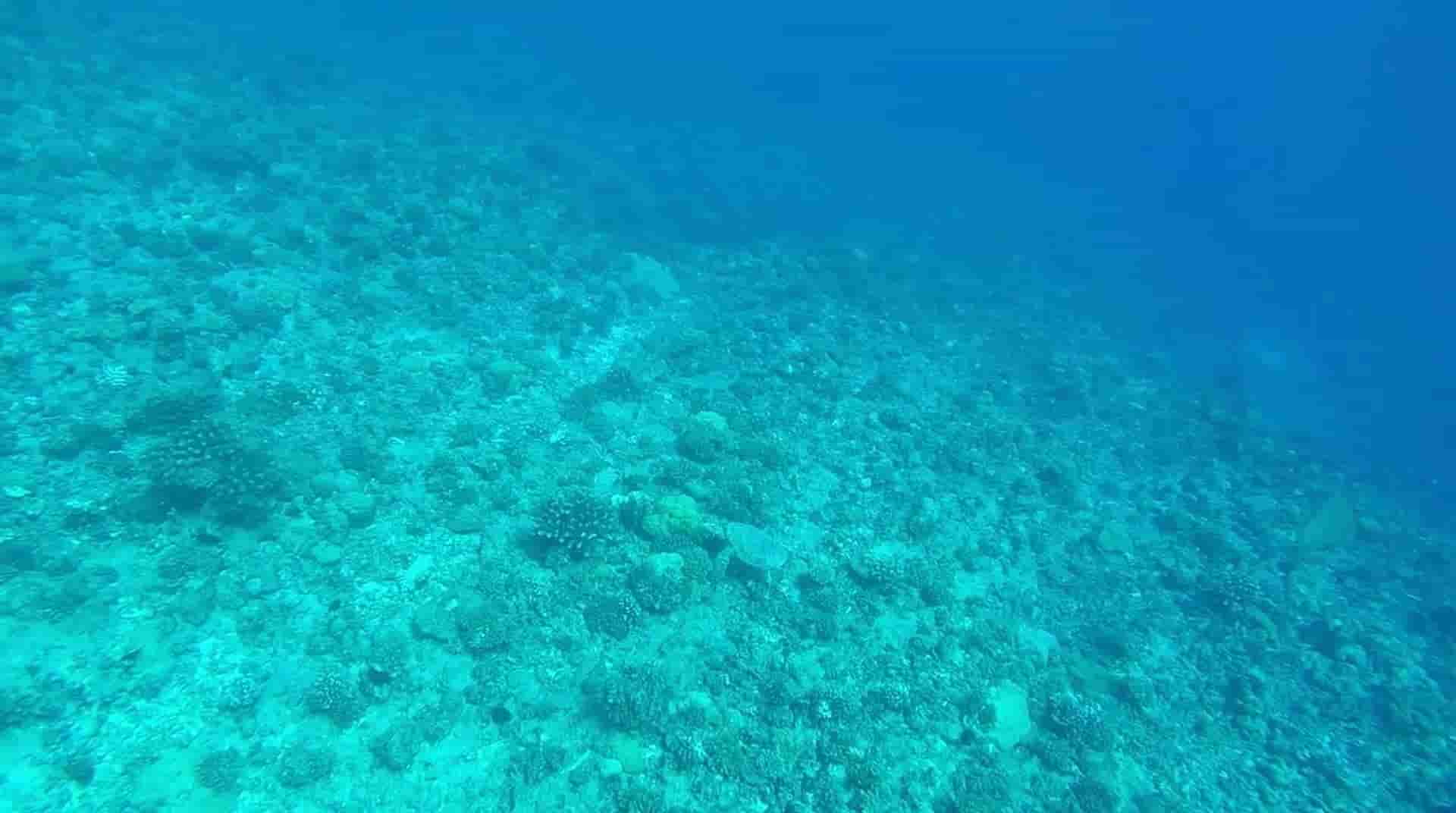}}
        \begin{picture}(0,0)
            \put(-18,19){\textcolor{white}{\footnotesize 35.00}}
        \end{picture}
    \end{minipage}
    \begin{minipage}[b]{0.19\linewidth}
        \subfloat{\includegraphics[width=\linewidth]{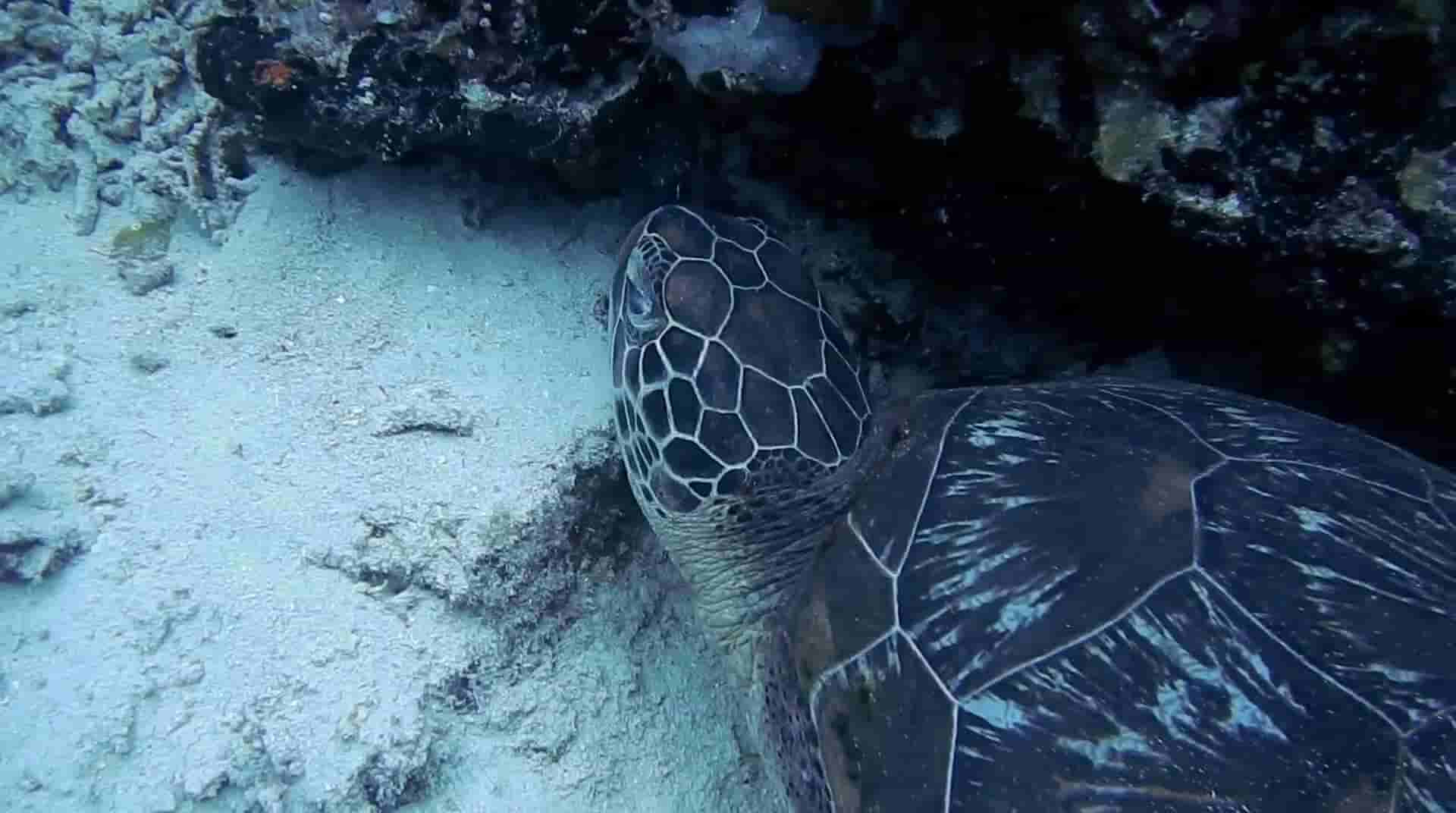}}
        \begin{picture}(0,0)
            \put(-18,19){\textcolor{white}{\footnotesize 55.07}}
        \end{picture}
    \end{minipage}
    \begin{minipage}[b]{0.19\linewidth}
        \subfloat{\includegraphics[width=\linewidth]{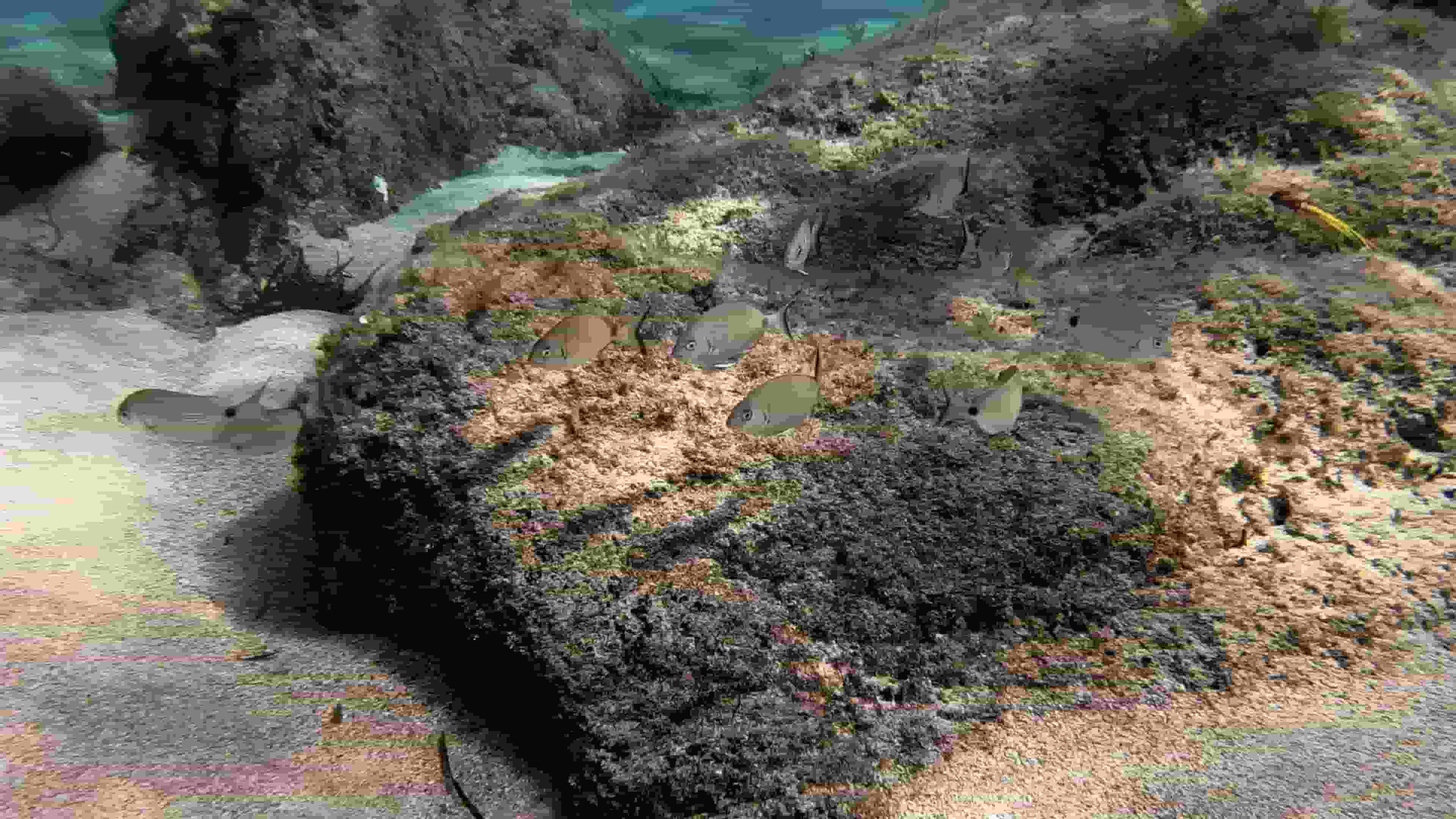}}
        \begin{picture}(0,0)
            \put(-18,19){\textcolor{white}{\footnotesize 75.25}}
        \end{picture}
    \end{minipage}
    \begin{minipage}[b]{0.19\linewidth}
        \subfloat{\includegraphics[width=\linewidth]{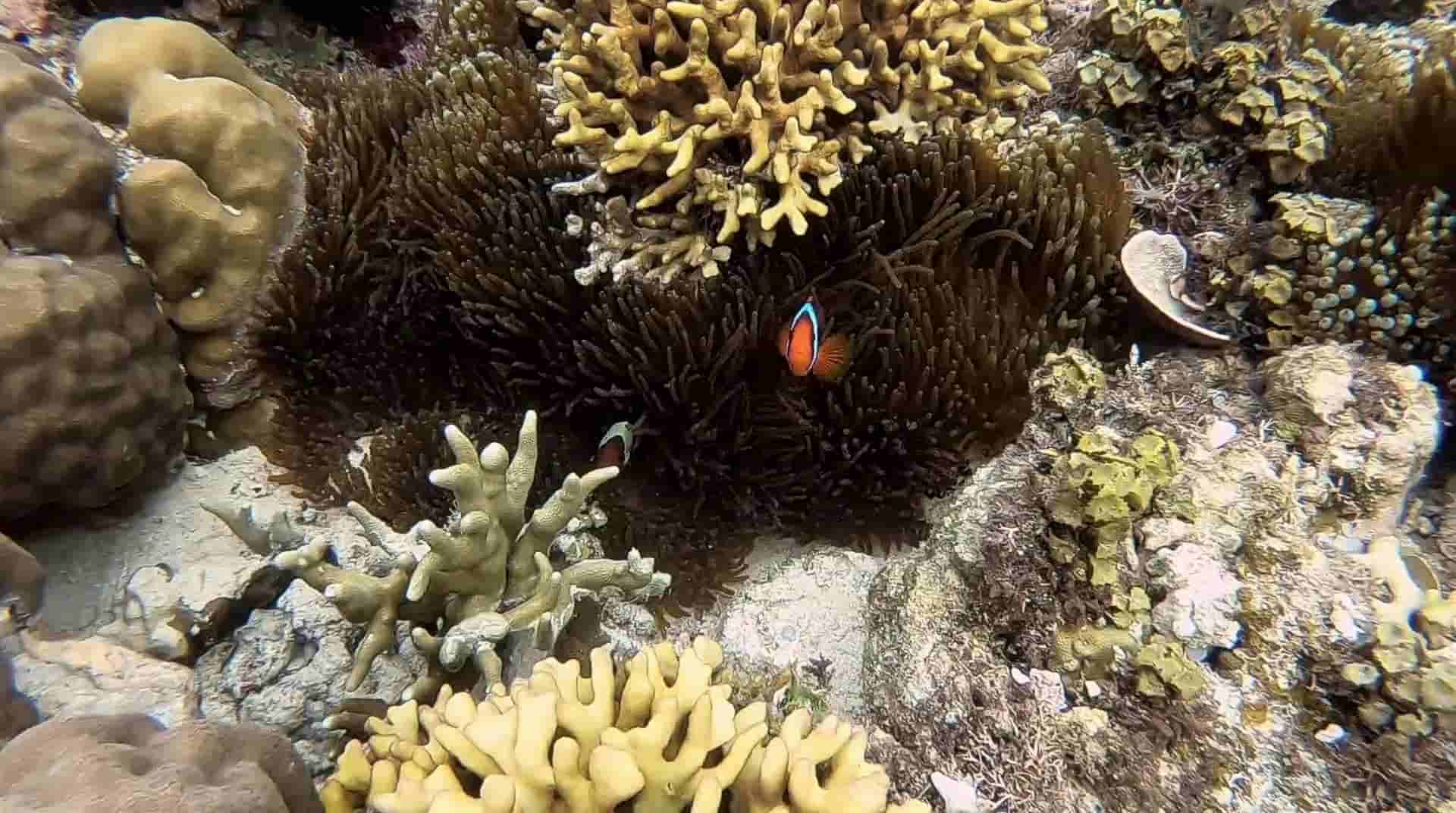}}
        \begin{picture}(0,0)
            \put(-18,19){\textcolor{white}{\footnotesize 95.04}}
        \end{picture}
    \end{minipage}
    \vspace{0cm}
    \begin{minipage}[b]{0.19\linewidth}
        \subfloat{\includegraphics[width=\linewidth]{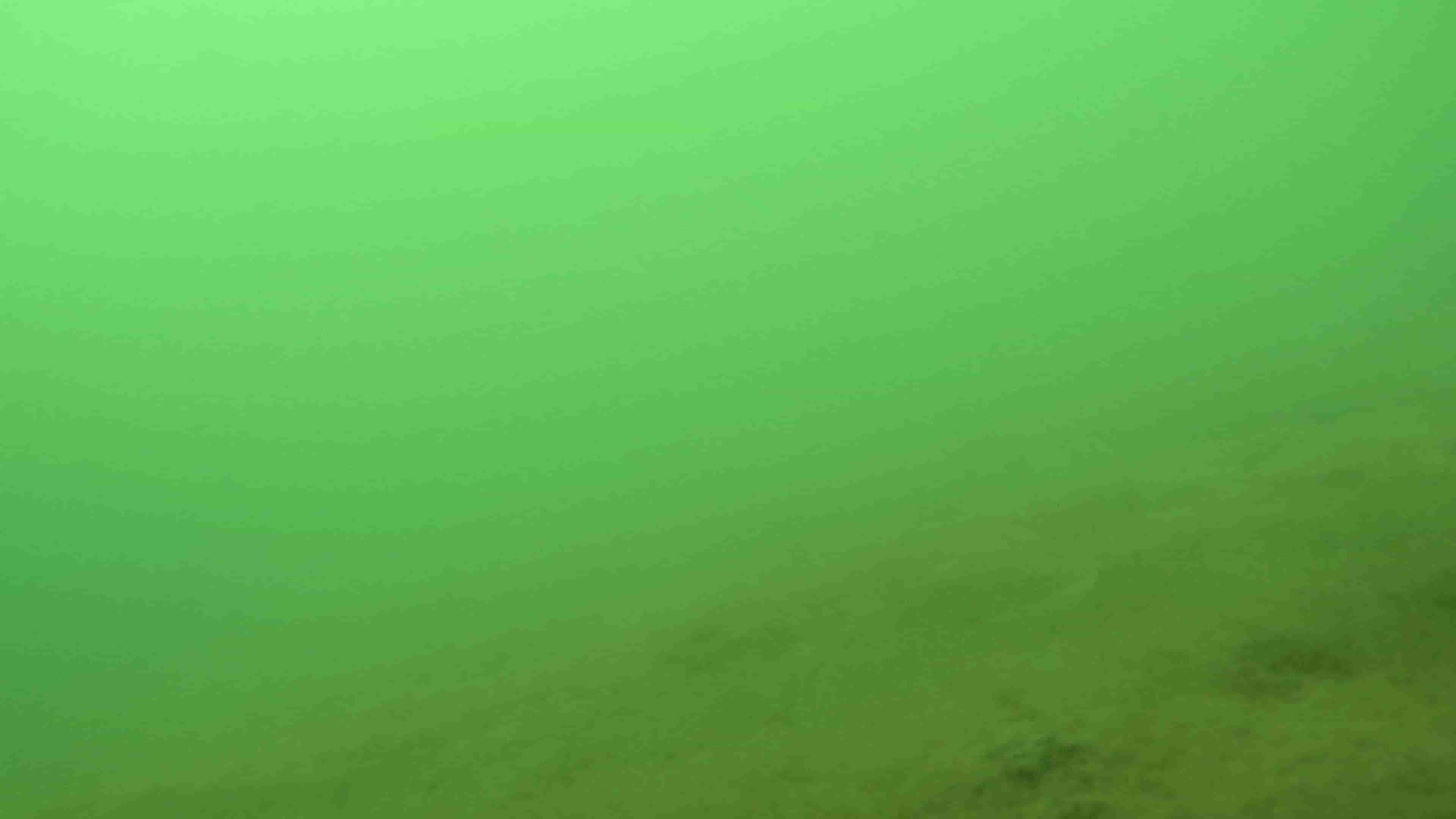}}
        \begin{picture}(0,0)
            \put(-18,19){\textcolor{white}{\footnotesize 14.00}}
        \end{picture}
    \end{minipage}
    \begin{minipage}[b]{0.19\linewidth}
        \subfloat{\includegraphics[width=\linewidth]{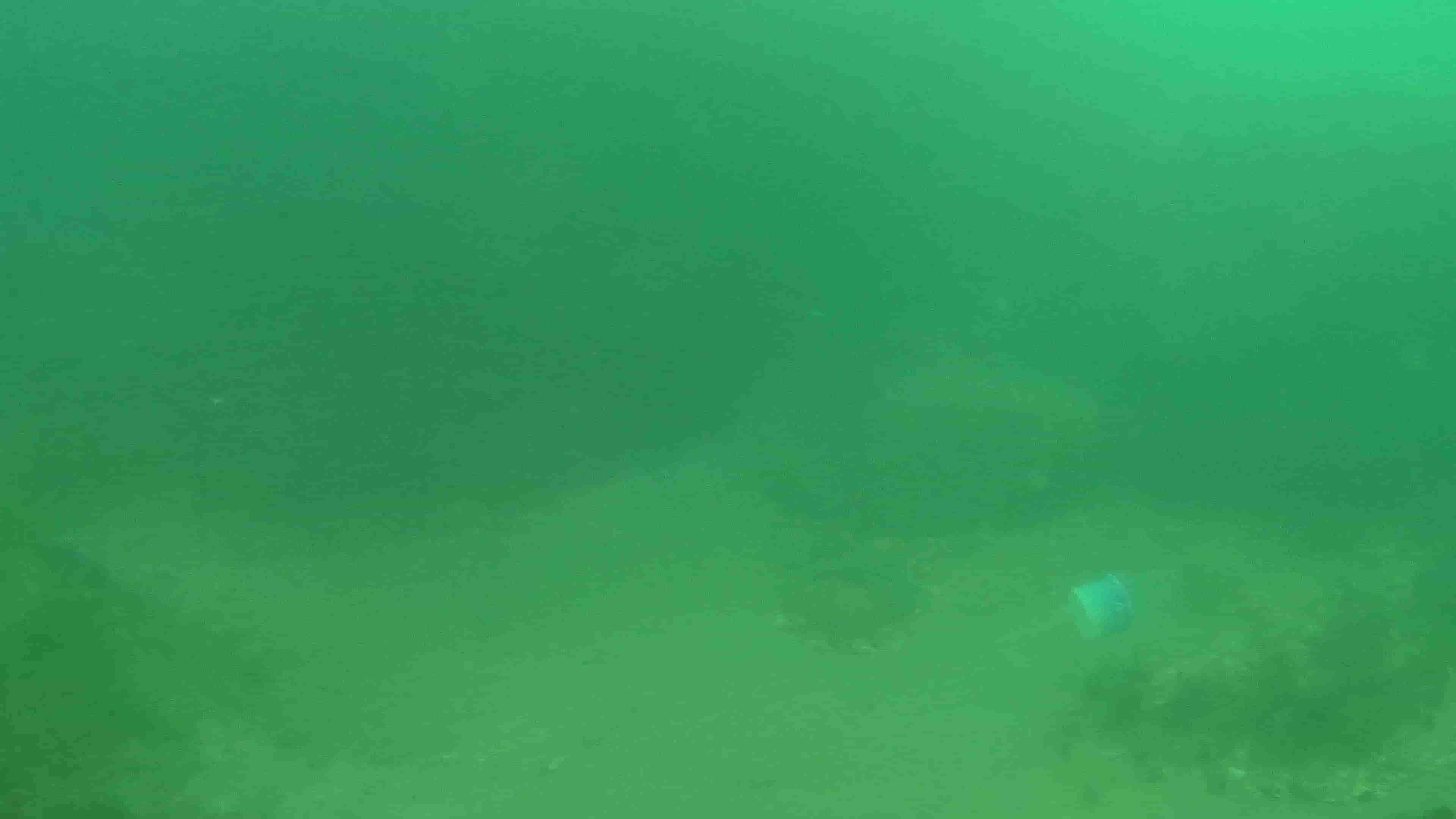}}
        \begin{picture}(0,0)
            \put(-18,19){\textcolor{white}{\footnotesize 16.60}}
        \end{picture}
    \end{minipage}
    \begin{minipage}[b]{0.19\linewidth}
        \subfloat{\includegraphics[width=\linewidth]{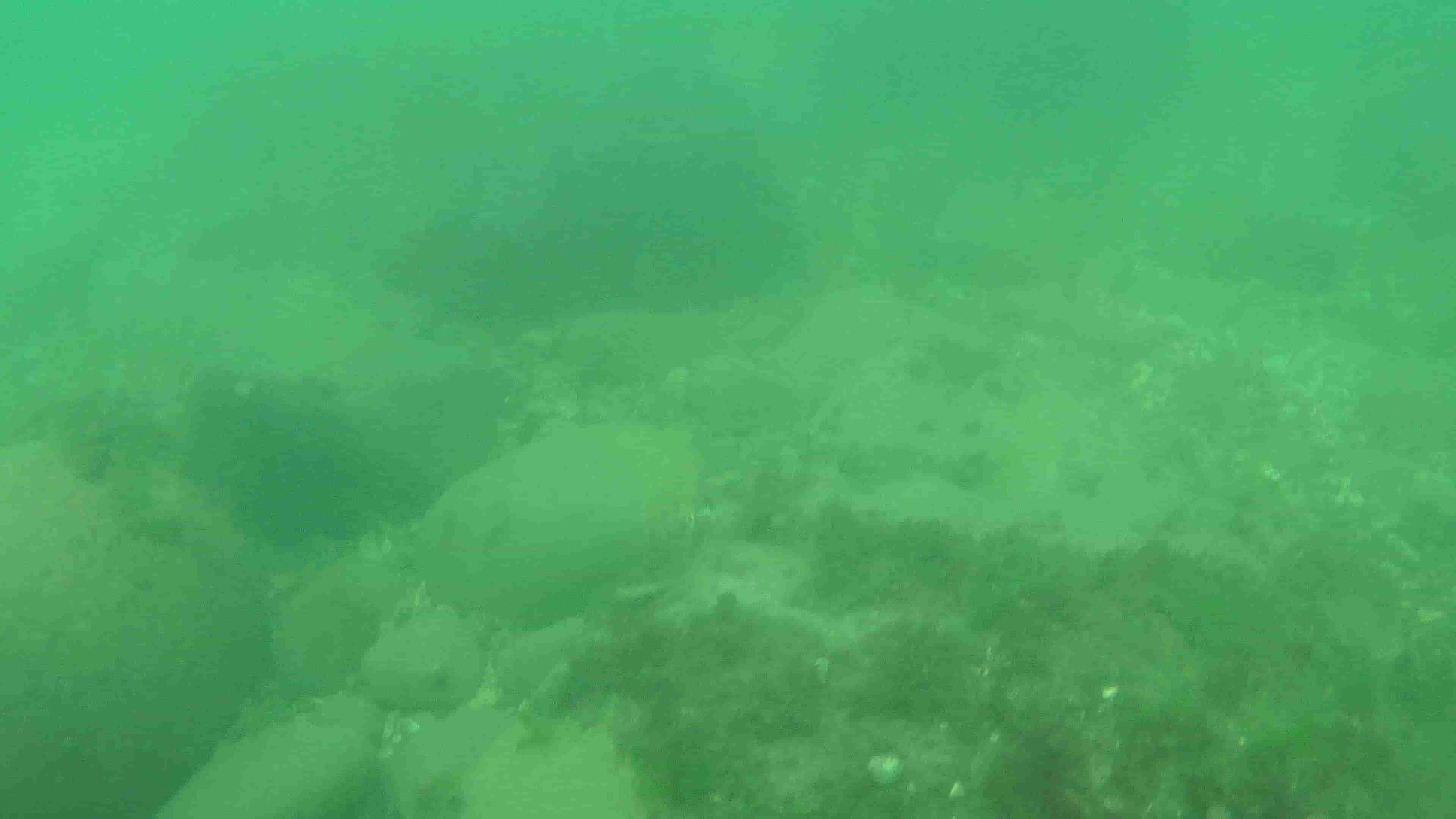}}
        \begin{picture}(0,0)
            \put(-18,19){\textcolor{white}{\footnotesize 18.80}}
        \end{picture}
    \end{minipage}
    \begin{minipage}[b]{0.19\linewidth}
        \subfloat{\includegraphics[width=\linewidth]{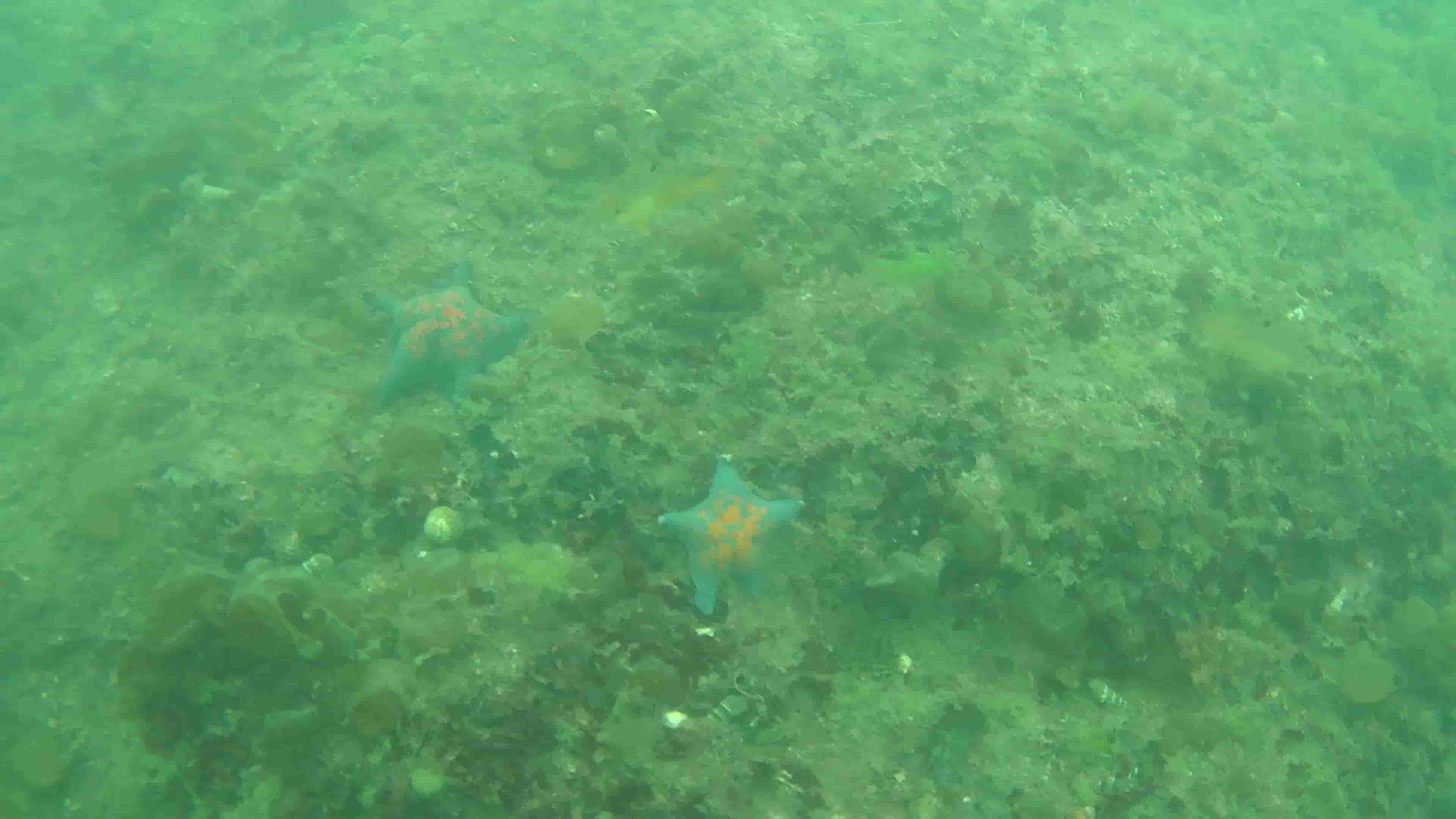}}
        \begin{picture}(0,0)
            \put(-18,19){\textcolor{white}{\footnotesize 22.43}}
        \end{picture}
    \end{minipage}
    \begin{minipage}[b]{0.19\linewidth}
        \subfloat{\includegraphics[width=\linewidth]{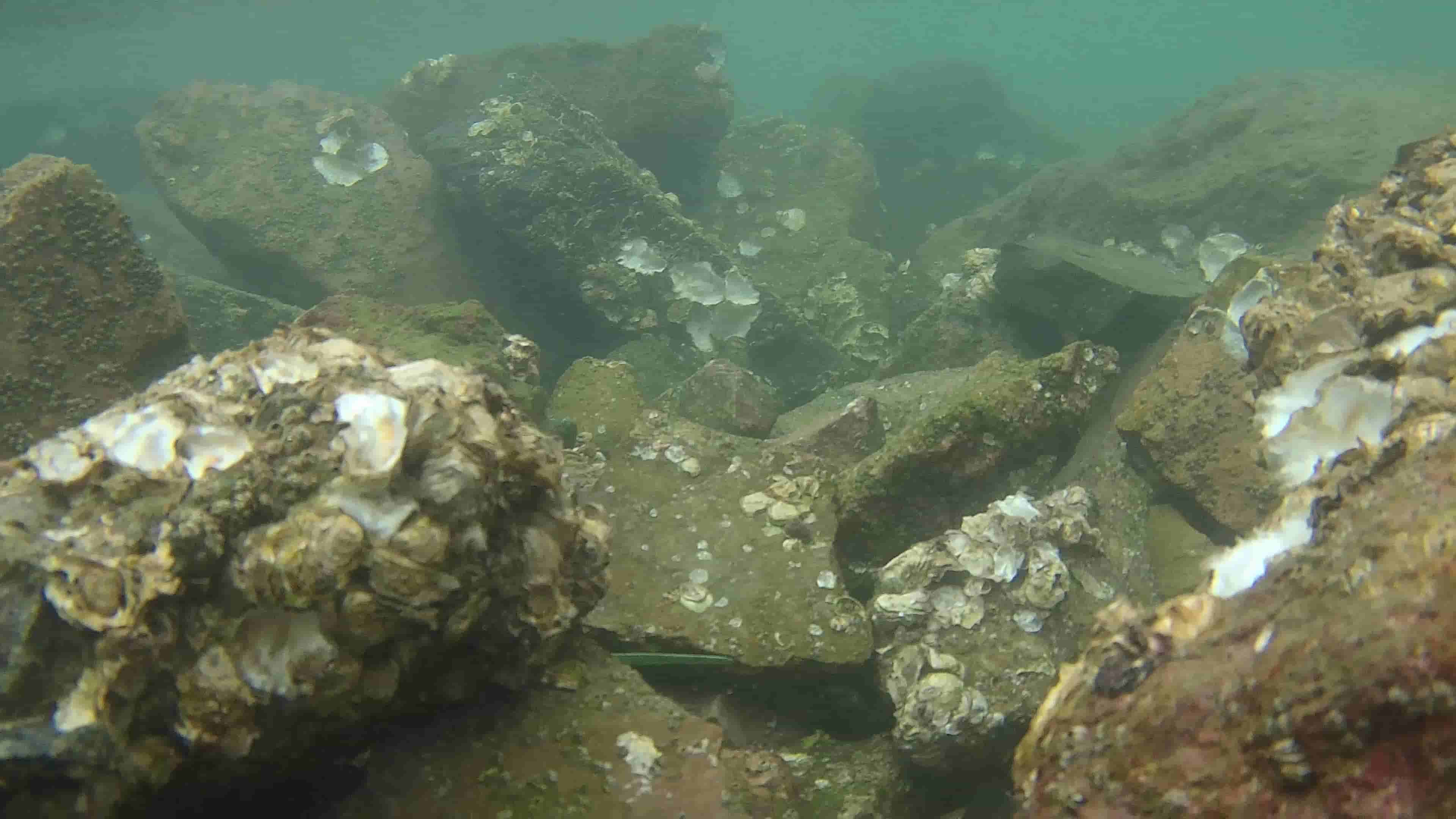}}
        \begin{picture}(0,0)
            \put(-18,19){\textcolor{white}{\footnotesize 47.36}}
        \end{picture}
    \end{minipage}
    \caption{The rating increases as the video quality improves.}
    \label{fig:rating}
\end{figure}
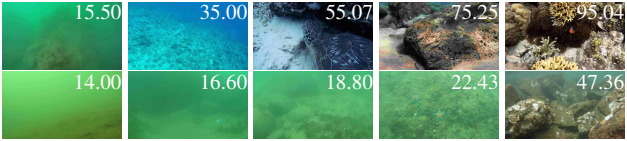
\noindent{\bf Diversity of Dataset.} UVEB includes various underwater scenes such as offshore, open sea, rivers, lakes, ports, aquariums, swimming pools, etc. \cref{fig:diversity} shows the scene diversity and degradation type diversity of UVEB samples. UVEB video degradation mainly includes six types: blue, green, yellow, white, other colors, and insufficient lighting. \cref{fig:2} (b) shows the proportion of six types of underwater video degradation in UVEB. UVEB dataset contains 25\% yellow, white, and other color deviation data, as well as underwater videos with insufficient light, which are rarely mentioned but appear in actual scenes. \cref{fig:2} (a) shows that the distribution of video collectors' source countries in this dataset is diverse, which is more than twenty. \cref{fig:2} (c) shows the resolution information of the overall dataset. The resolutions of most data are larger than 2K. The number of frames in the various resolution intervals in our UVEB dataset totaled 453,874 frames. \cref{fig:mos} shows the mean opinion scores (MOS) of the samples before and after enhancement. We can see that the UVEB includes samples of diverse quality and the GT quality is better than raw videos.

According to observations during the data collection, the diversity of water types, imaging distances, and ambient light contributes to the variety of color deviations in underwater images. From \cref{fig:cc}, we can find that the color deviations may be diverse due to changes in ambient light even in the same video.

\begin{figure}[t]
    \centering
    \includegraphics[width=\linewidth]{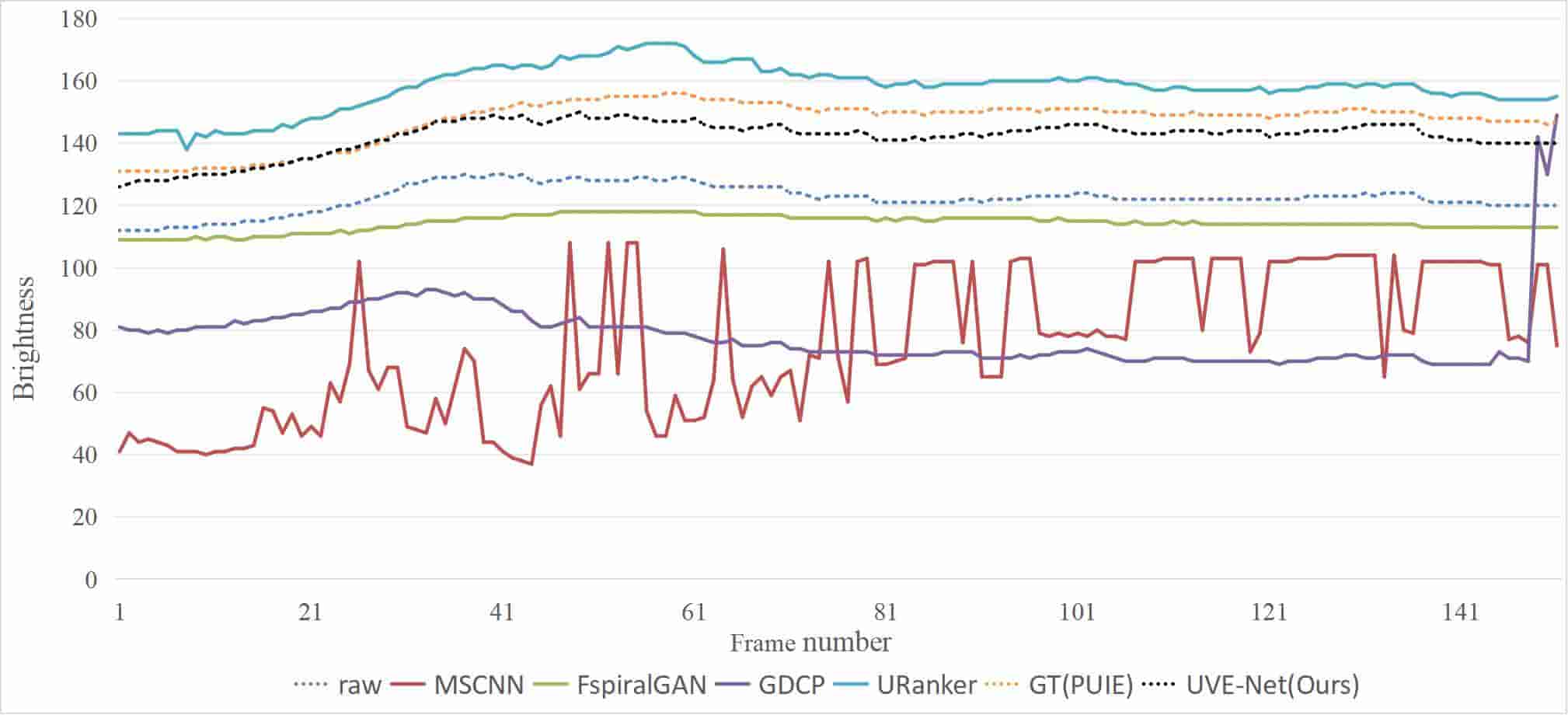}
    \caption{Brightness variation curves for different enhancement results of \#1057 video.}
    \label{fig:brightness}
\end{figure}

\noindent{\bf Reliable Samples Quality.} We calculate the proportion of samples within two standard deviations like~\cite{wang2022generation}, which is 96.99\% and larger than 95\%. According to ~\cite{series2012methodology}, our evaluation process is reliable and the error is controlled within a reasonable range. \cref{fig:rating} shows the partial scoring results of the observers. The first line shows that the ratings of videos with different color deviations increase with quality improvement. The second line shows that the ratings of videos with the same color deviation increase as the degree of color deviation decreases. The better the overall quality of the image, the higher the score. 

Since image enhancement methods do not consider the correlation between video frames and enhance a single frame individually, people may be concerned about the flickering issues among different enhanced frames. To investigate this point, we show the brightness variation curves of different enhancement methods with an video example in~\cref{fig:brightness}. Some brightness curves fluctuate sharply, such as MSCNN~\cite{ren2016single} and GDCP~\cite{peng2018generalization} in \cref{fig:brightness}. According to observations, due to the good fitting ability of neural networks, the enhancement results of most deep learning methods will not encounter this problem, such as the enhancement results of FspiralGAN~\cite{guan2023fast} and PUIE~\cite{fu2022uncertainty} methods used as GT in the sample. Only the enhanced results with stable brightness changes and no frame flicker can be chosen as GT.

\begin{figure*}[h]
  \centering
  \includegraphics[width=\textwidth]{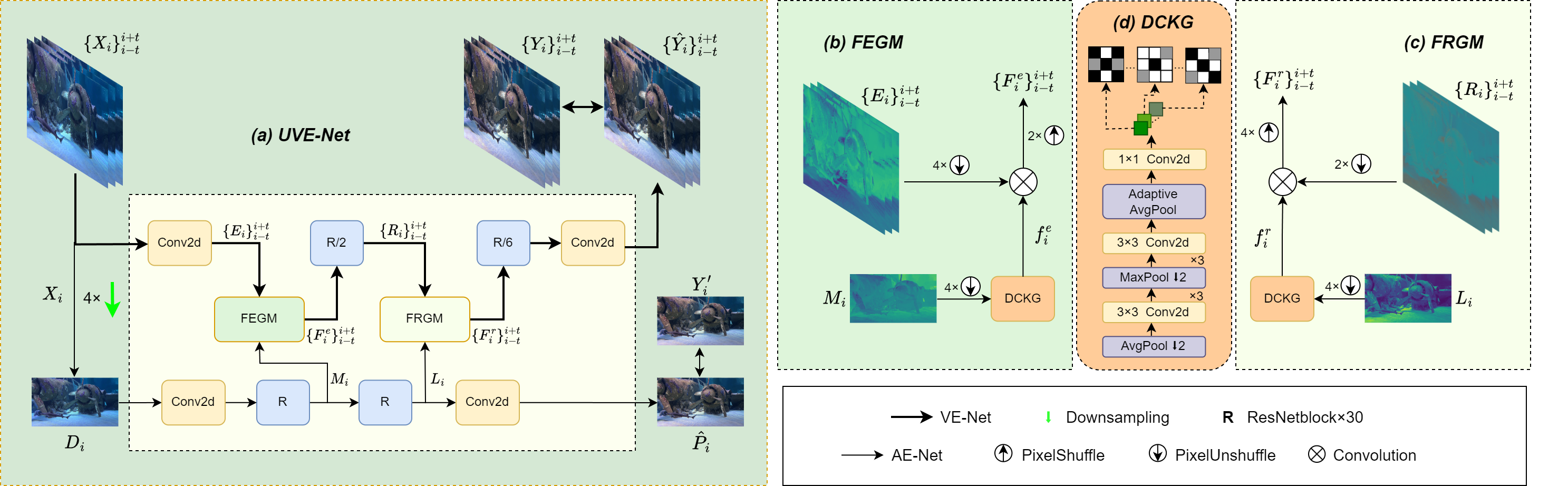}
  \caption{(a) Overall framework of the UVE-Net. UVE-Net includes the upper branch Video Enhancement Network, and the lower branch Auxiliary Enhancement Network. (b) FEGM. (c) FRGM. (d) DCKG. (R, R$/$2, and R$/$6 mean 30, 15, and 5 residual blocks.) }
  \label{fig:network}
\end{figure*}

%% file: sec/4_UVE-Net.tex
\section{UVE-Net}
\label{sec:UVE-Net}
\hspace{0.16667in}
%
In most cases,  the water body and degradation level between adjacent frames are quite similar. Thus, we can make use of this fact and let the adjacent frames follow similar feature extraction and enhancement processes to accelerate the inference speed. 
The downsampled frame and its original frame have similar contents and degradation process, which makes them follow a similar enhancement process. Therefore, we can use the enhancement process of the low-resolution downsampled frame to guide the original frame to complete enhancement more directional efficiently.

In UVE-Net, we first use an auxiliary network to understand and solve the image enhancement problem of the downsampled middle frame. Then, the auxiliary network converts the problem-solving process (enhancement process) into action instructions (convolutional kernels) and passes them on to the main network. The main network completes middle frame enhancement more directionally and efficiently based on the guidance information.
Based on the strong correlation of degradation in adjacent frames, these convolutional kernels (action instructions) are also transmitted to adjacent frames to help them complete enhancement more efficiently.
\subsection{UVE-Net Overall framework}
\hspace{0.16667in}
As shown in \cref{fig:network} (a), UVE-Net comprises the upper branch video enhancement network (VE-Net) and the lower branch auxiliary enhancement network (AE-Net).  AE-Net completes the quality enhancement of ${D_i}\in\mathbb{R}^{\frac{H}{4} \times \frac{W}{4} \times 3}$. VE-Net completes the enhancement of the current frames $\{X_i\in\mathbb{R}^{H \times W \times 3}\}_{i-t}^{i+t}$. The overall framework is aim to efficiently transfer the enhancement process of low resolution middle frame to the current frame to be restored, helping the current frame to better complete the image enhancement process efficiently.

$\{X_i\}_{i-t}^{i+t}$ are the degraded frames, where ${X_i}$ is the middle frame. ${X_i}$ gets down-sampled low-resolution representation ${D_i}$ through $\times$4 downsample operation. 
AE-Net uses the enhancement process of ${D_i}$ to guide $\{X_i\}_{i-t}^{i+t}$ complete enhancement process.
The guidance process is carried out at low resolution, bringing less computational costs.

AE-Net and VE-Net complete preliminary feature extraction through $3 \times 3$ convolution. VE-Net converts the middle extraction feature ${M_i}\in\mathbb{R}^{\frac{H}{4} \times \frac{W}{4} \times C}$ and clean restoration feature ${L_i}\in\mathbb{R}^{\frac{H}{4} \times \frac{W}{4} \times C}$ in the enhancement process of ${D_i}$ into convolutional kernel sequences ${f_i^e}\in\mathbb{R}^{3 \times 3 \times 16C}$ and ${f_i^r}\in\mathbb{R}^{3 \times 3 \times 16C}$ through the feature extraction guidance module (FEGM) and feature restoration guidance module (FRGM).
By enlighteningly performing more efficient information exchange through the transfer of convolutional kernels without feature alignment or aggregation, we reduce the computational costs required 
in FRGM and FEGM modules. We also use group convolution in FEGM and FERM to further reduce their computational costs. Subsequent experiments show that FRGM and FEGM can significantly improve the learning performance of VE-Net with low computational costs.

$f_i^e$ and $f_i^r$ serve as the action guidance for the feature extraction and enhancement of the current frame ${X_i}$, helping VE-Net performs more efficient feature transformations. Based on the strong correlation of degradation in adjacent frames,  $f_i^e$ and $f_i^r$ are also transmitted to adjacent frames to help them complete enhancement better and faster. For frames $\{X_{i}\}_{i-t}^{i+t}$, the auxiliary network AE-Net is only activated once in $X_i$, which also makes the guidance process is carried out in an efficient way. 
\subsection{FEGM}
\hspace{0.16667in} FEGM shown in \cref{fig:network} (b) converts the intermediate extracting feature during the enhancement process of ${D_i}$ into convolutional kernels $f_i^e$ and delivers them to the current frames to help VE-Net complete feature extraction better without frame alignment and inter-frame information aggregation. The process of FEGM can be expressed as:
\begin{equation}\label{2}
    \{F_i^e\}_{i-t}^{i+t} = FEGM(\{E_i\}_{i-t}^{i+t} , M_i)
\end{equation}
where $M_i$ represents the coarse feature extraction of the middle frame in the lower branch. $\{E_i\in\mathbb{R}^{H \times W \times C}\}_{i-t}^{i+t}$ represents the initial feature extraction of $\{X_i\}_{i-t}^{i+t}$.

The initial extracted feature ${D^{\prime}_i\in\mathbb{R}^{\frac{H}{4} \times \frac{W}{4} \times C}}$ of the upper branch passes through 30 residual blocks to generate the middle extracted feature ${M_i}$. 
${M_i}$ and $\{E_i\}_{i-t}^{i+t}$ from $\{X_i\}_{i-t}^{i+t}$ are respectively processed by PixelUnshuffle(4 $\times$ $\downarrow$) and be converted to ${M_i^4\in\mathbb{R}^{\frac{H}{16} \times \frac{W}{16} \times 16C}}$ and $\{E_i^4\in\mathbb{R}^{\frac{H}{4} \times \frac{W}{4} \times 16C}\}_{i-t}^{i+t}$ for channel adjustment. The purpose of these adjustments is to generate convolutional kernels while avoiding drastic changes in network channels. ${M_i^4}$ is converted into convolutional kernels $f_i^e$ through multiple pooling and group convolution operations like~\cite{pan2023deep}, which is described as dynamic convolutional kernels generation (DCKG) in \cref{fig:network} (d).
After this stage, the information of $H/16 \times W/16$ pixels is converted into a $3\times 3$ convolutional kernel. 
VE-Net performs more directional feature extraction with the help of $f_i^e$. $\{E_{i}^4\}_{i-t}^{i+t}$
is convolved with $f_i^e$ and pass through PixelShuffle(2$\times$ $\uparrow$) to get the guided extraction features $\{F_{i}^e\in\mathbb{R}^{\frac{H}{2} \times \frac{W}{2} \times 4C}\}_{i-t}^{i+t}$. To summarize the above process, \cref{2} can be 
be expressed in detail as:
\begin{equation}\label{3}
    \{F_i^e\}_{i-t}^{i+t} = (DCKG(M_i\downarrow_{4}) \ast \{E_i\}_{i-t}^{i+t}\downarrow_{4})\uparrow_{2}
\end{equation}
where $\downarrow_{4}$ represents the PixelUnshuffle rate as 4 , $\uparrow_{2}$ represents the PixelShuffle rate as 2, $(\ast)$ represents convolution.
\subsection{FRGM}
\hspace{0.16667in}Under the guidance of low-resolution clean features ${L_i}$ obtained before ${D_i}$ completes enhancement, the current frame $\{X_i\}_{i-t}^{i+t}$ can complete the mapping transformation to the clear image $\{Y_i\}_{i-t}^{i+t}$. The mathematical expression is as follows:
\begin{equation}\label{4}
    \{F_i^r\}_{i-t}^{i+t} = FRGM(\{R_i\}_{i-t}^{i+t} , L_i)
\end{equation}
where $\{R_i\in\mathbb{R}^{\frac{H}{4} \times \frac{W}{4} \times 4C}\}_{i-t}^{i+t}$ represents the input frames refined extracted feature of the upper branch, $L_i$ represents the clean features obtained before ${D_i}$ completes enhancement in the lower branch.

Thus, we design FRGM to convert the ${L_i}$ into convolutional kernels $f_i^r$ and deliver them to $\{X_i\}_{i-t}^{i+t}$. VE-Net performs more directional feature restoration with the help of $f_i^r$. \cref{fig:network} (c) shows the architecture of FRGM. Specifically, ${L_i}$ is reshaped to ${L_i^4\in\mathbb{R}^{\frac{H}{16} \times \frac{W}{16} \times 16C}}$ by PixelUnshuff(4 $\times$ $\downarrow$) and $\{R_{i}\}_{i-t}^{i+t}$ is reshaped to $\{R_{i}^2\in\mathbb{R}^{\frac{H}{4} \times \frac{W}{4} \times 16C}\}_{i-t}^{i+t}$ by PixelUnshuff(2$\times$$\downarrow$). Then ${L_i^4}$ is transformed into $f_i^r$ through pooling and convolutional layers like FEGM. 
$\{R_{i}^2\}_{i-t}^{i+t}$ is convolved with $f_i^r$ and pass through PixelShuffle(4 $\times$ $\uparrow$) to obtain the guided restoration features $\{F_i^r\in\mathbb{R}^{H \times \ W \times C}\}_{i-t}^{i+t}$. \cref{4}  is transformed into:
\begin{equation}\label{5}
    \{F_i^r\}_{i-t}^{i+t} = (DCKG(L_i\downarrow_{4}) \ast \{R_i\}_{i-t}^{i+t}\downarrow_{2})\uparrow_{4}
\end{equation}
where $\downarrow_{4}$ represents the PixelUnshuffle rate as 4 , $\downarrow_{2}$ represents the PixelUnshuffle rate as 2, $\uparrow_{4}$ represents the PixelShuffle rate as 4, $(\ast)$ represents convolution.
\subsection{Loss Function}
\hspace{0.16667in}
We calculate the loss of the upper and lower branches as the total loss function $\mathcal{L}$.
\begin{equation}
     \mathcal{L}={\mathcal{L}}_{pix}\{(\widehat{Y_i} , {Y_i})\}_{i-t}^{i+t}+{\mathcal{L}}_{pix}({\widehat{P}_{i}} , {Y^{\prime}}_{i})
\end{equation}
where ${\mathcal{L}}_{pix}$ refers to Charbonnier Loss~\cite{lai2018fast}.
$\{Y_i\}_{i-t}^{i+t}$ is the GT of sample $\{X_i\}_{i-t}^{i+t}$, and ${Y^{\prime}}_{i}$ is the GT of $D_i$. 

%% file: sec/5_Experiments.tex
\section{Experiments}
\begin{figure*}[htbp]
    \centering
\begin{minipage}[b]{0.1208\linewidth}
    \includegraphics[width=\linewidth]{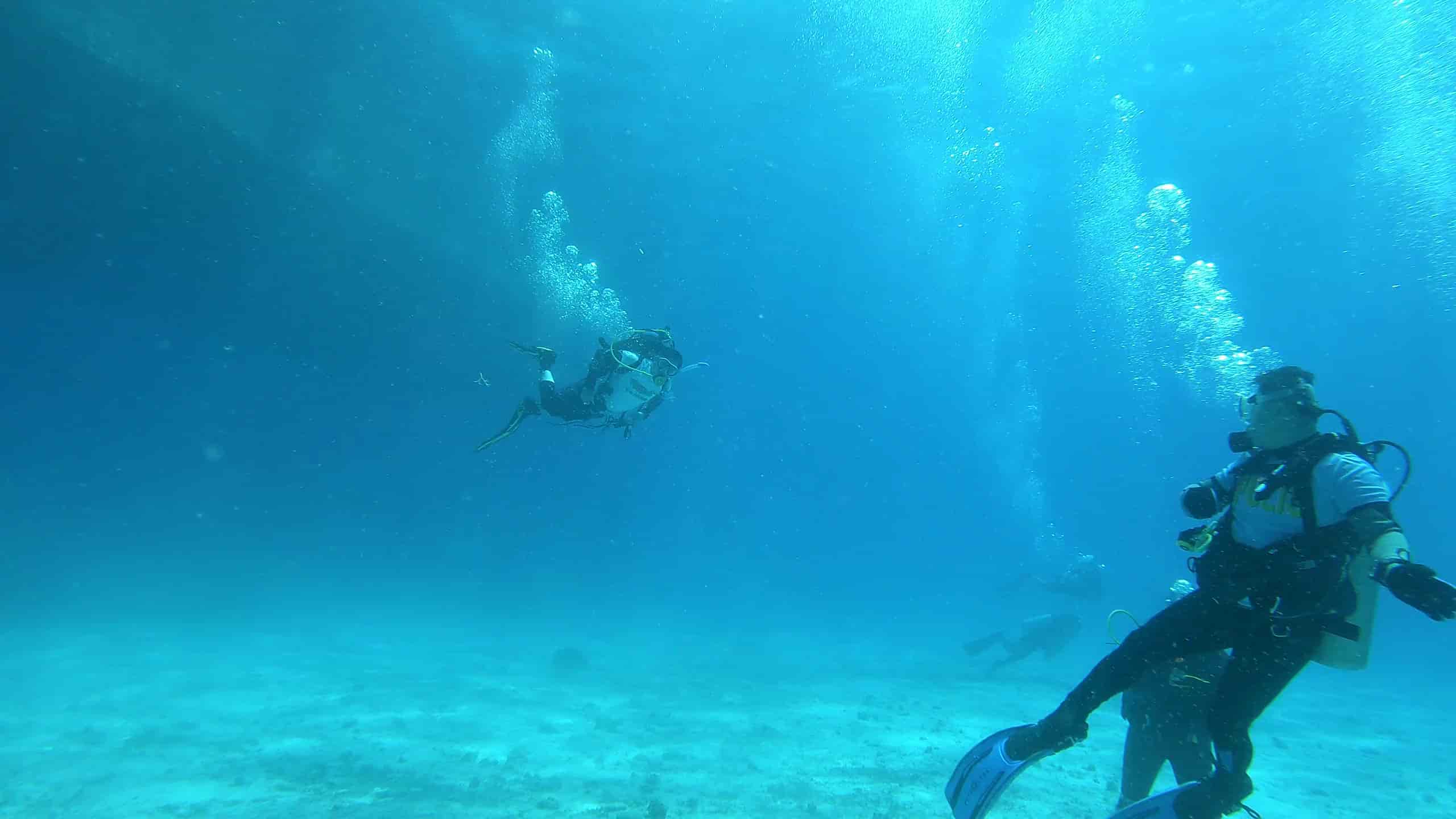}
\end{minipage}
\begin{minipage}[b]{0.1208\linewidth}
    \includegraphics[width=\linewidth]{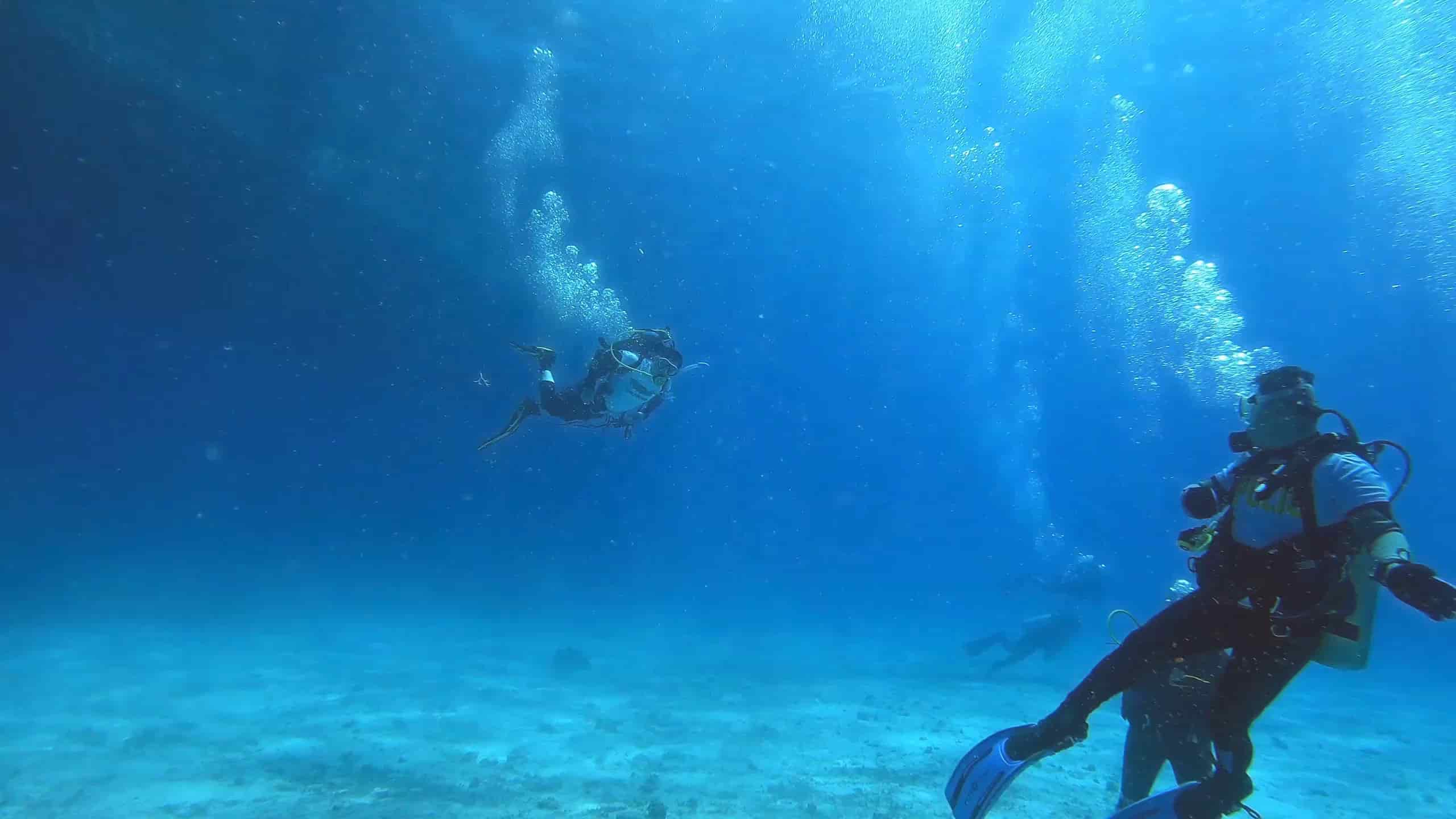}
\end{minipage}
\begin{minipage}[b]{0.1208\linewidth}
    \includegraphics[width=\linewidth]{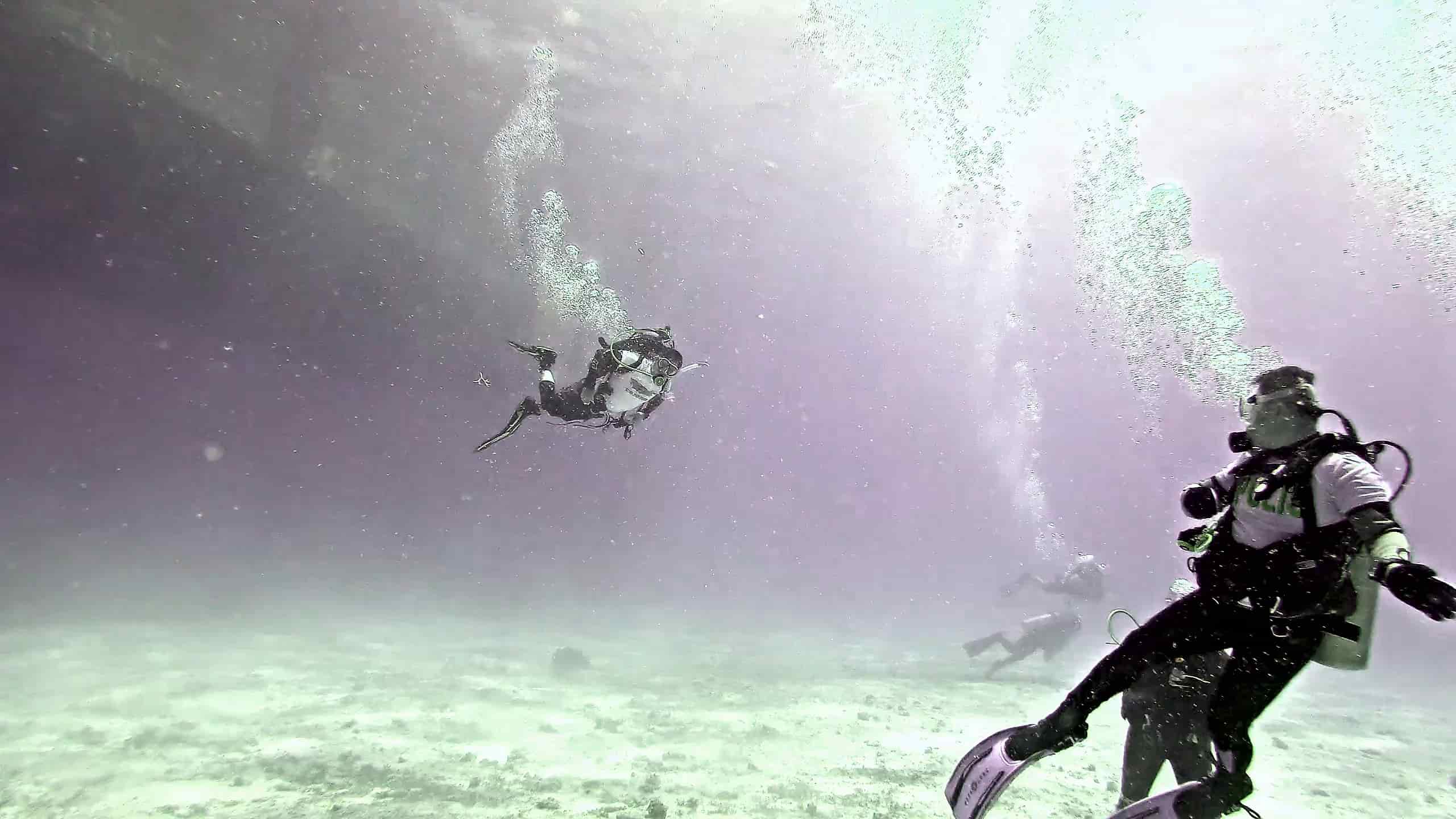}
\end{minipage}
\begin{minipage}[b]{0.1208\linewidth}
    \includegraphics[width=\linewidth]{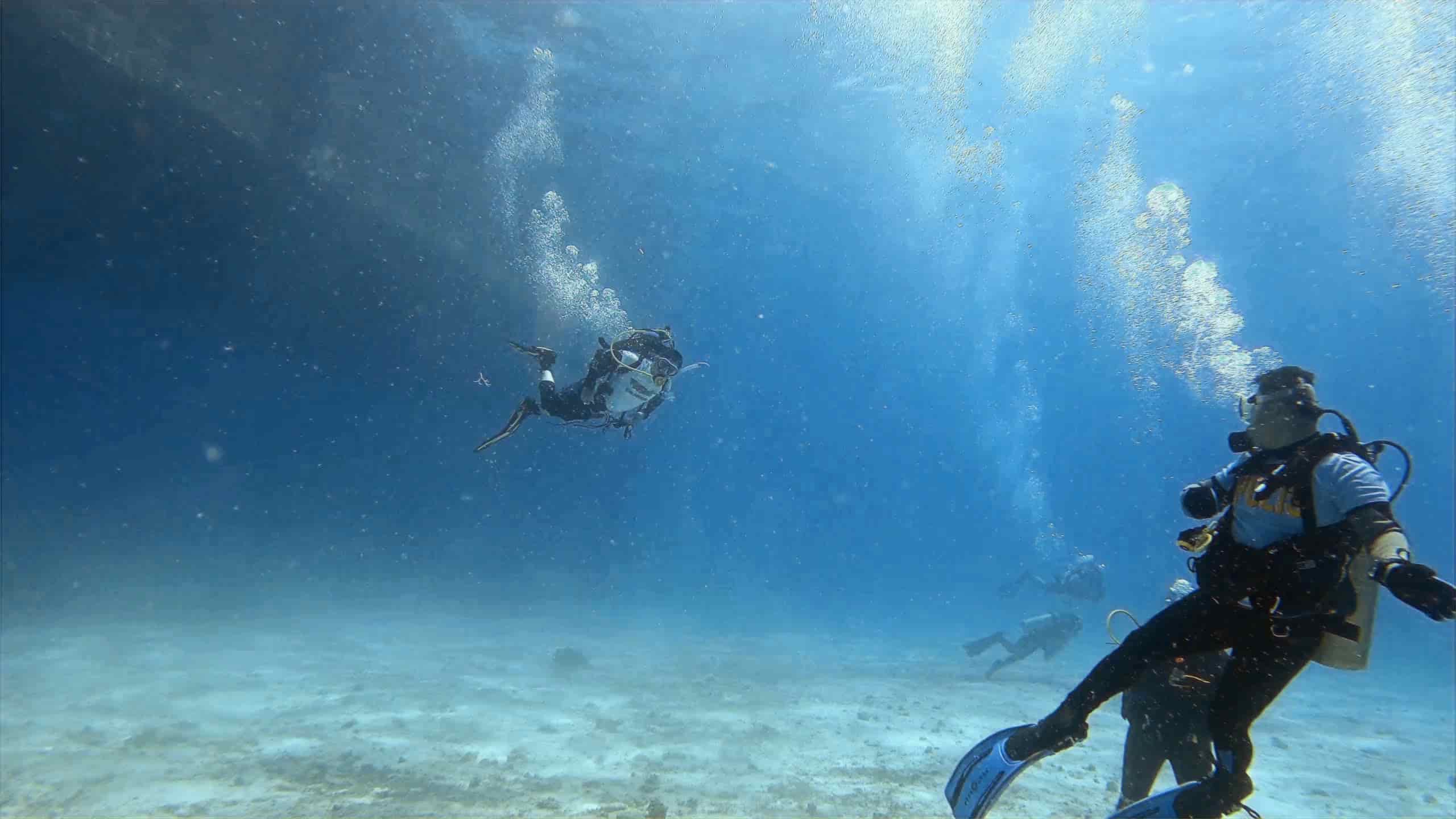}
\end{minipage} 
\begin{minipage}[b]{0.1208\linewidth}
    \includegraphics[width=\linewidth]{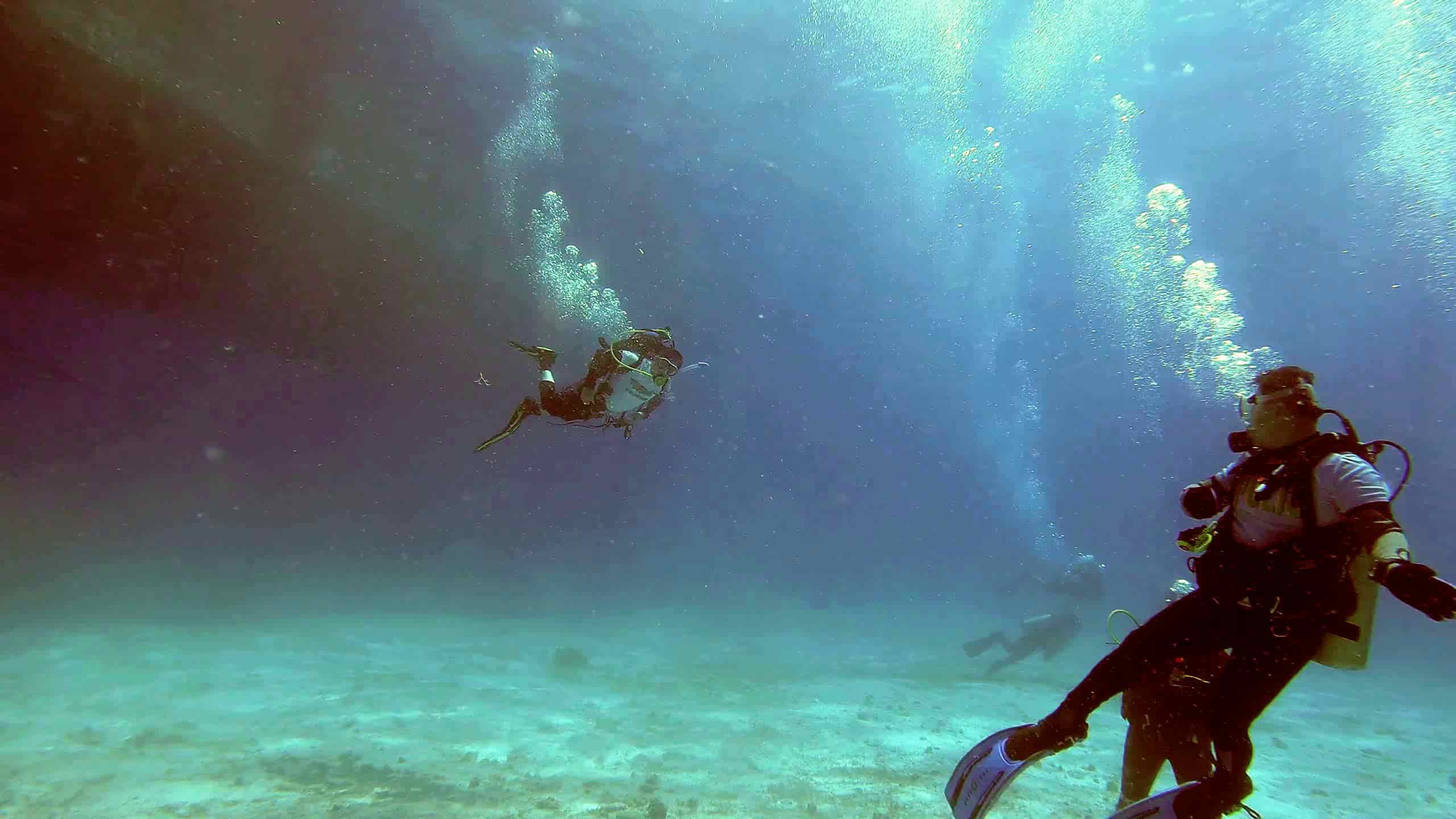}
\end{minipage}
\begin{minipage}[b]{0.1208\linewidth}
    \includegraphics[width=\linewidth]{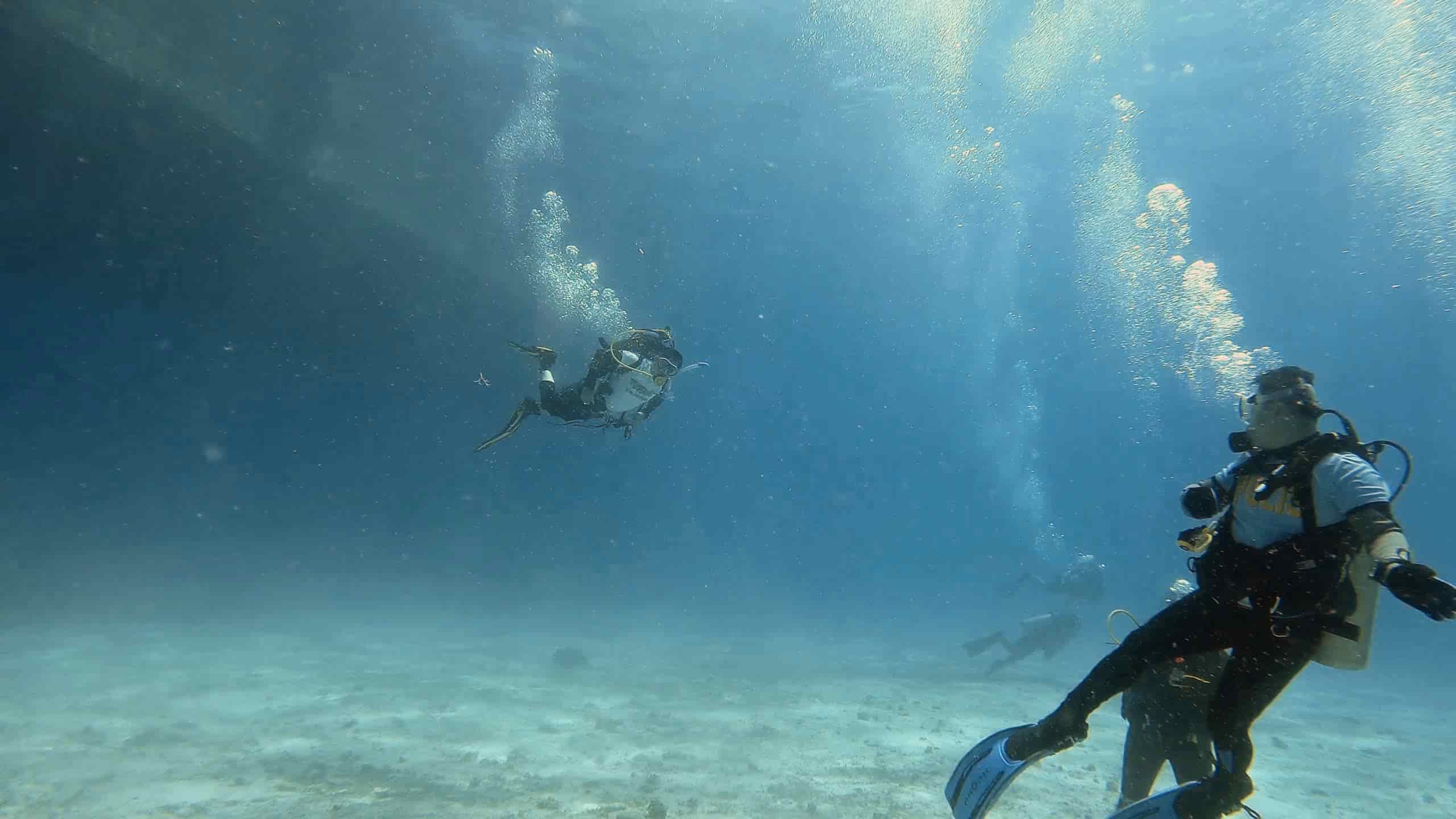}
\end{minipage}
\begin{minipage}[b]{0.1208\linewidth}
    \includegraphics[width=\linewidth]{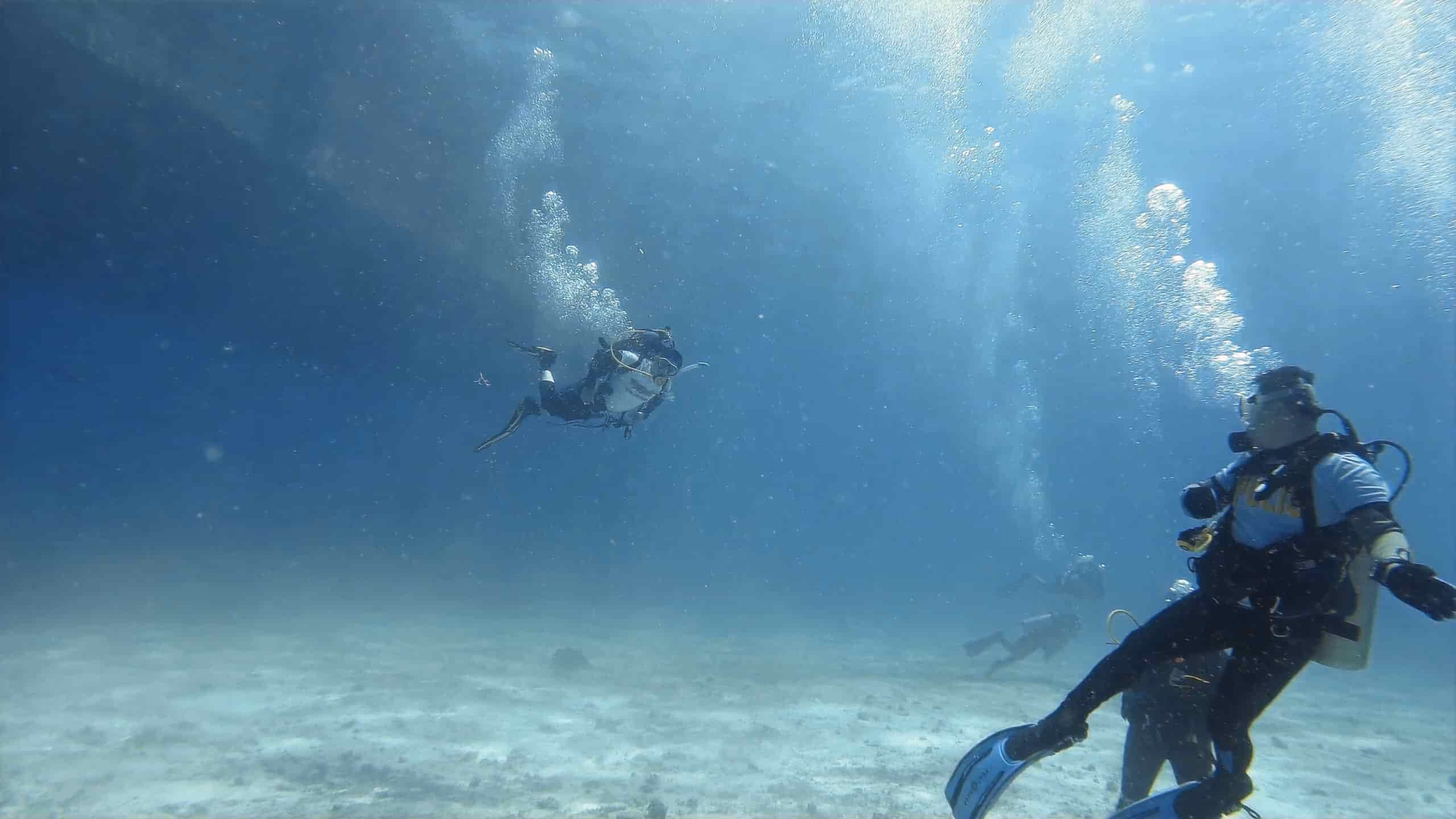}
\end{minipage}
\begin{minipage}[b]{0.1208\linewidth}
    \includegraphics[width=\linewidth]{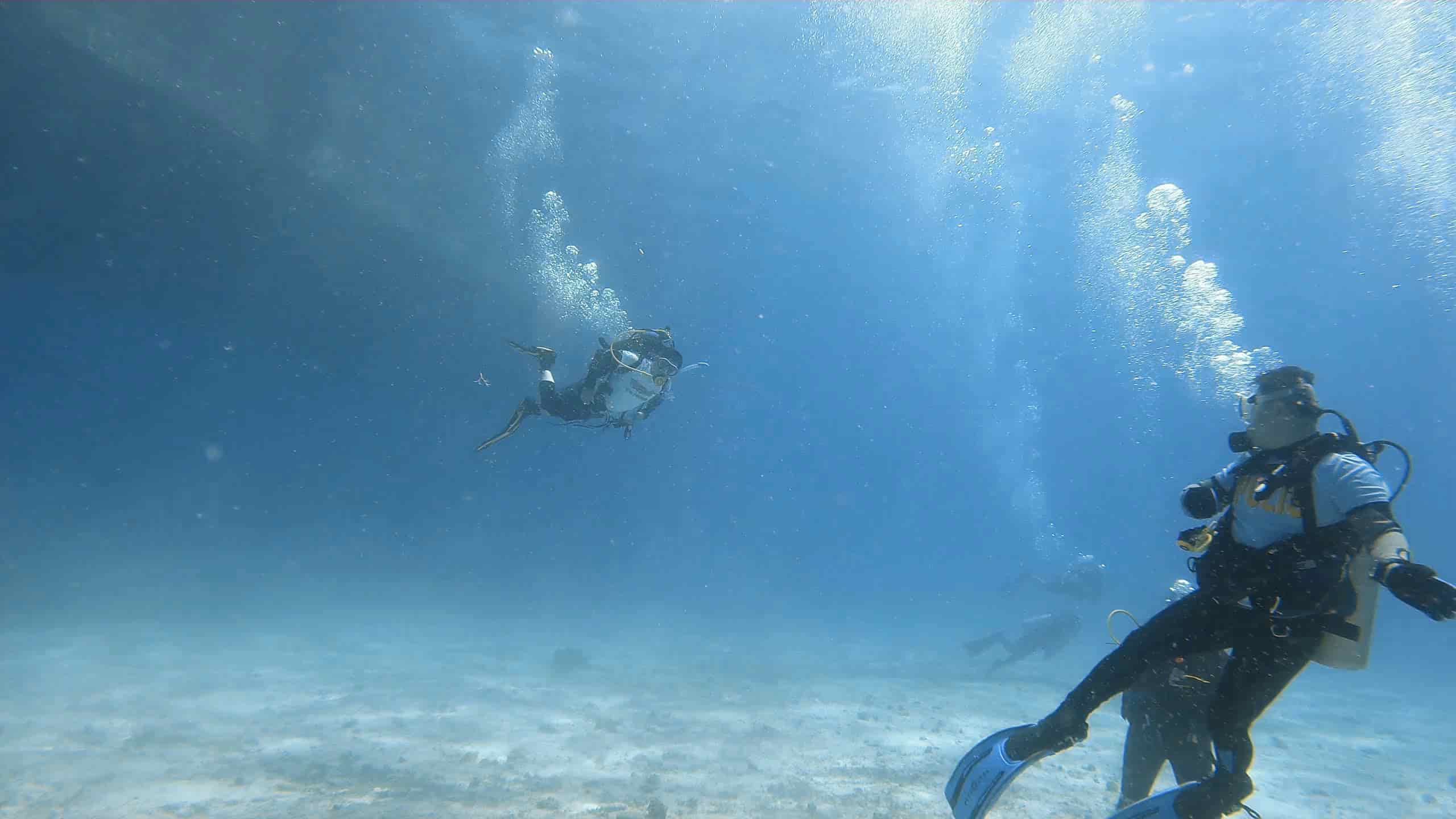}
\end{minipage}
\begin{minipage}[b]{0.1208\linewidth}
    \includegraphics[width=\linewidth]{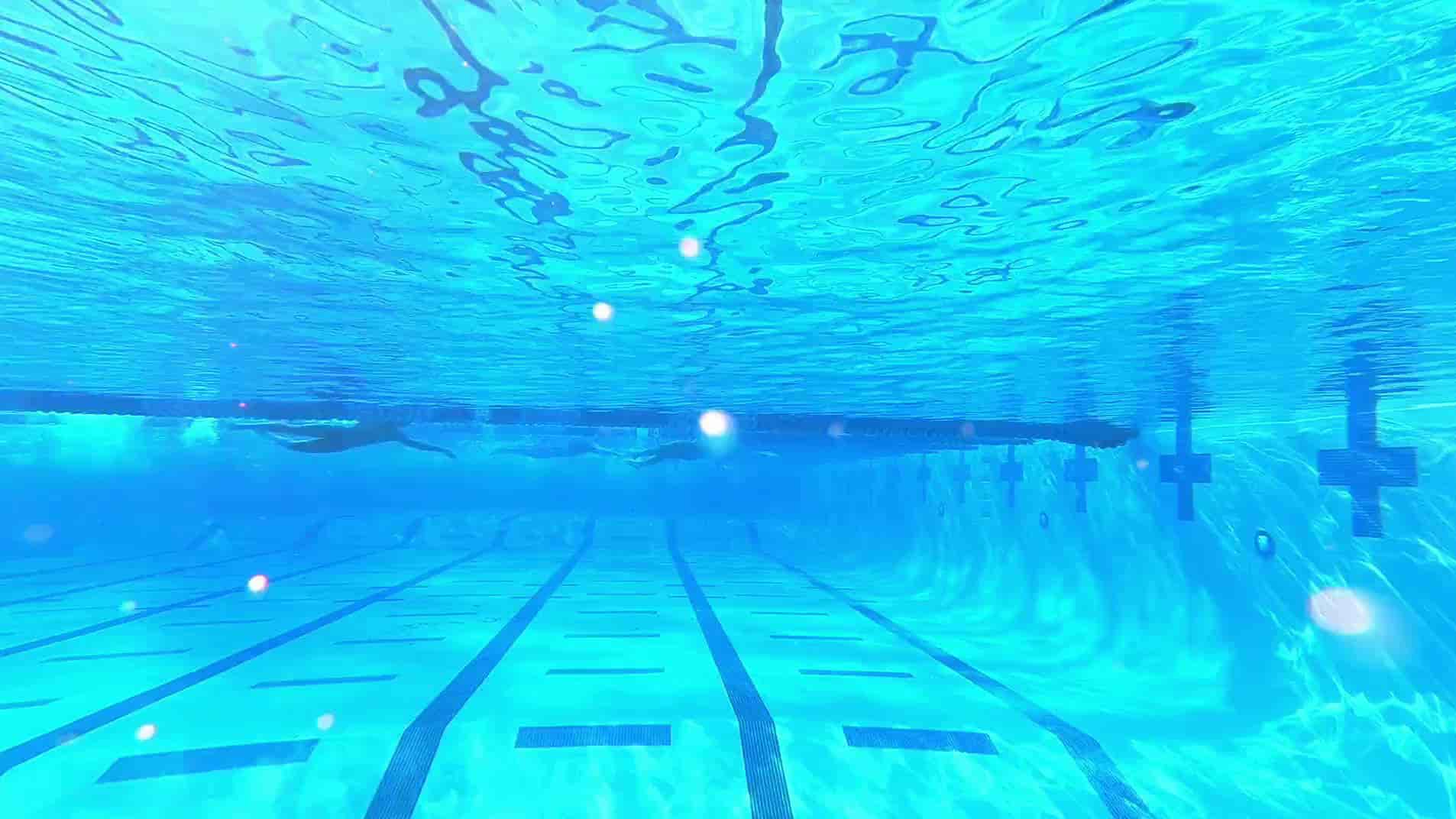}
\end{minipage}
\begin{minipage}[b]{0.1208\linewidth}
    \includegraphics[width=\linewidth]{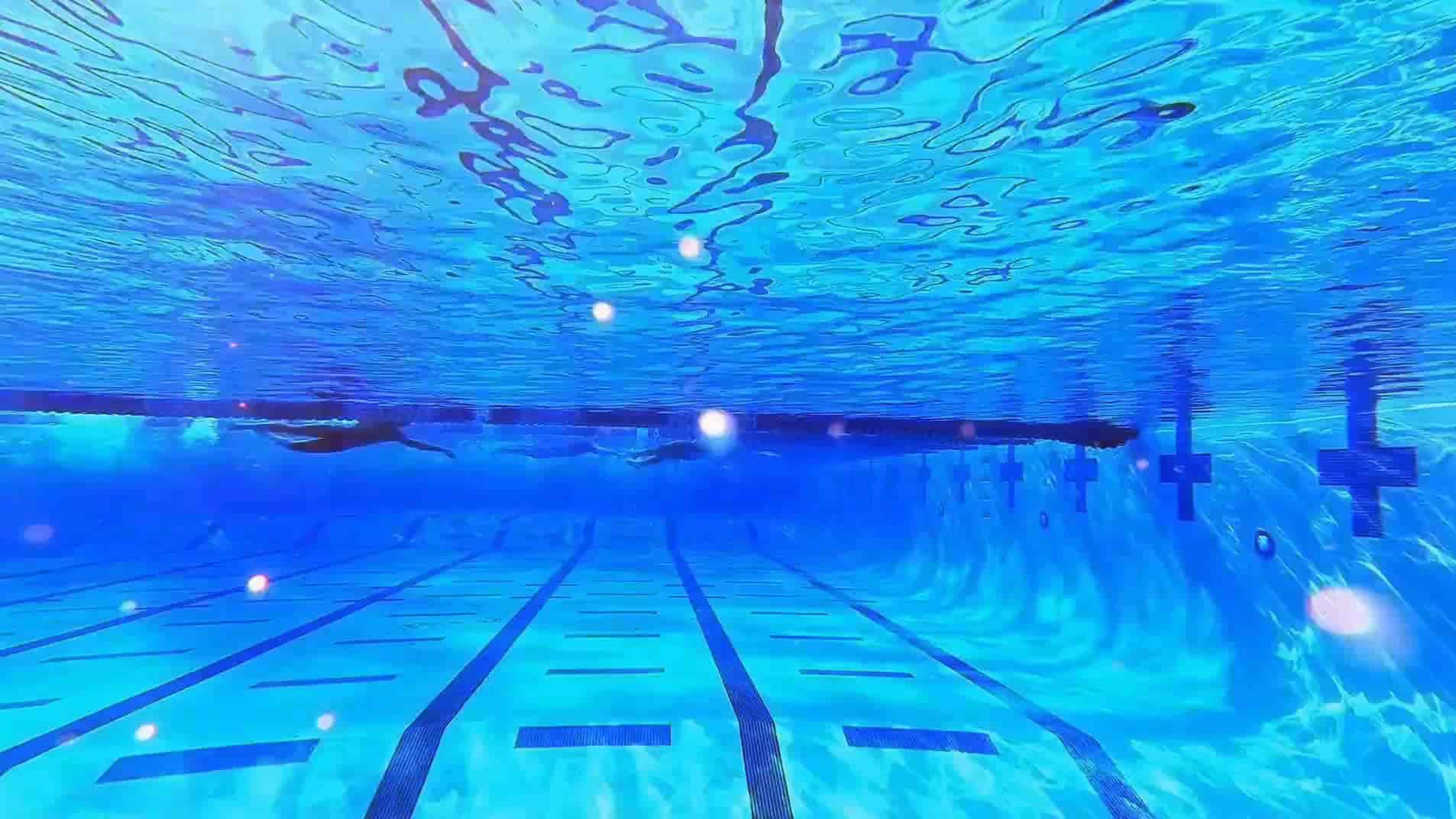}
\end{minipage}
\begin{minipage}[b]{0.1208\linewidth}
    \includegraphics[width=\linewidth]{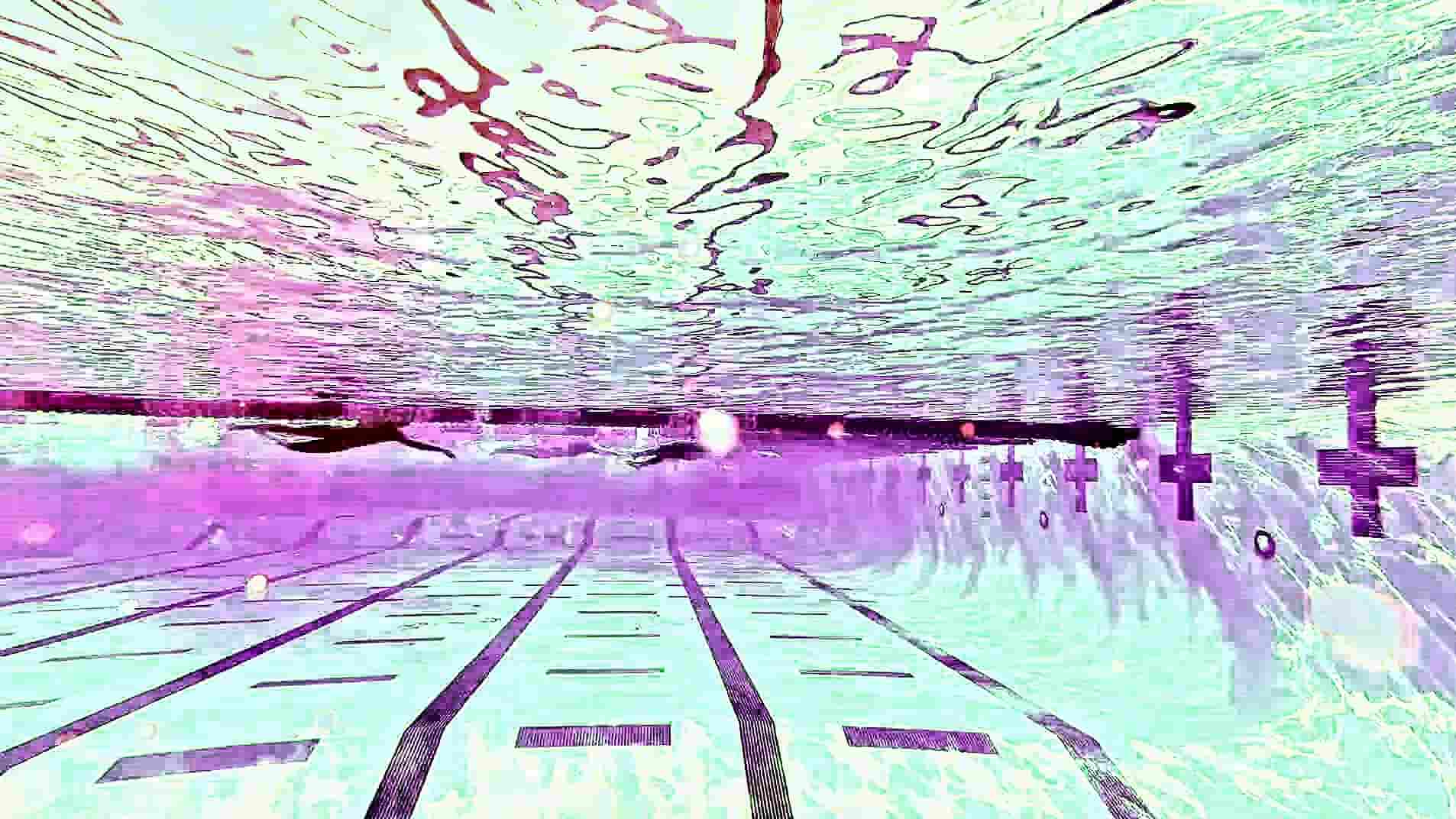}
\end{minipage}
\begin{minipage}[b]{0.1208\linewidth}
    \includegraphics[width=\linewidth]{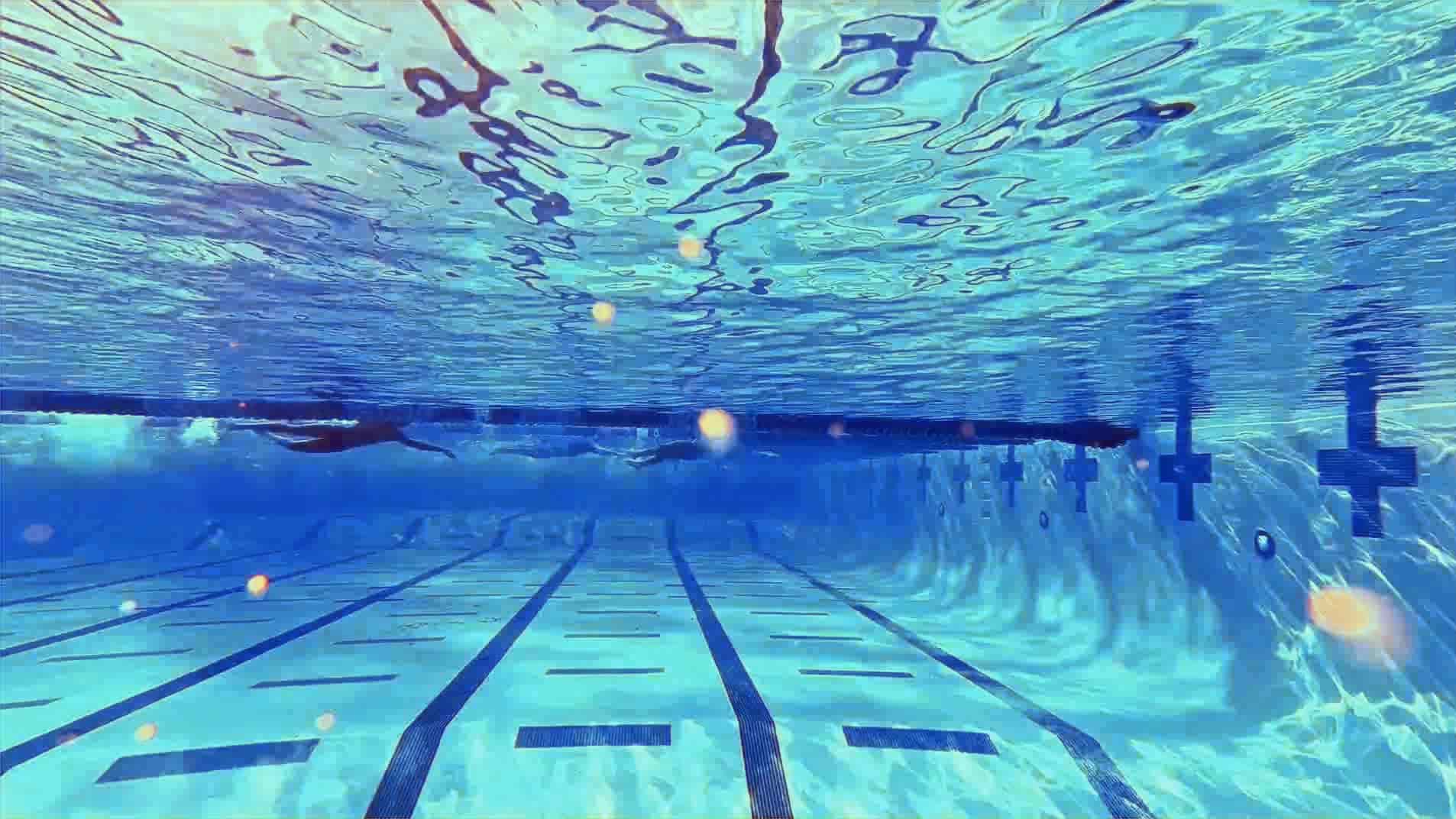}
\end{minipage} 
\begin{minipage}[b]{0.1208\linewidth}
    \includegraphics[width=\linewidth]{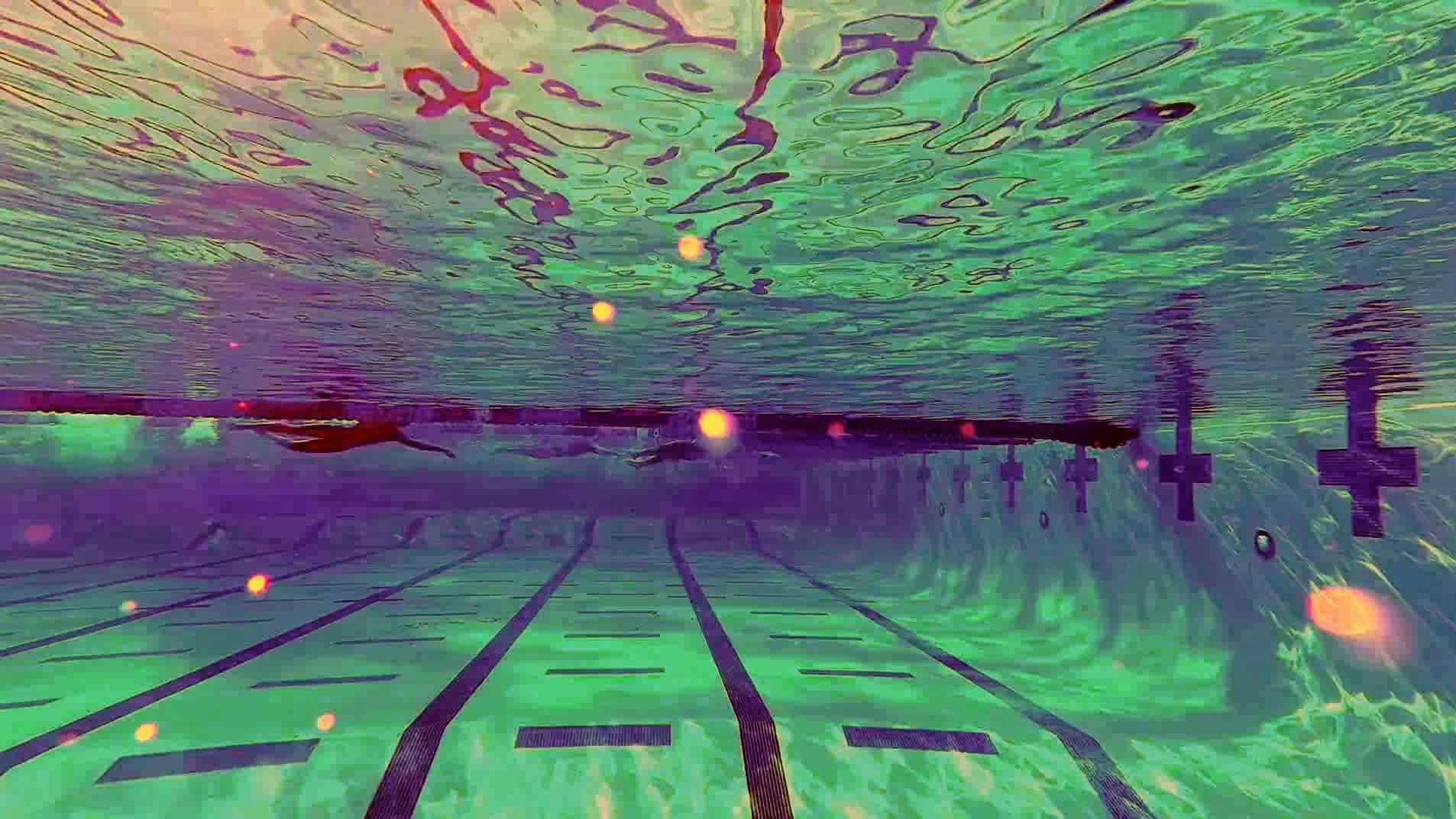}
\end{minipage}
\begin{minipage}[b]{0.1208\linewidth}
    \includegraphics[width=\linewidth]{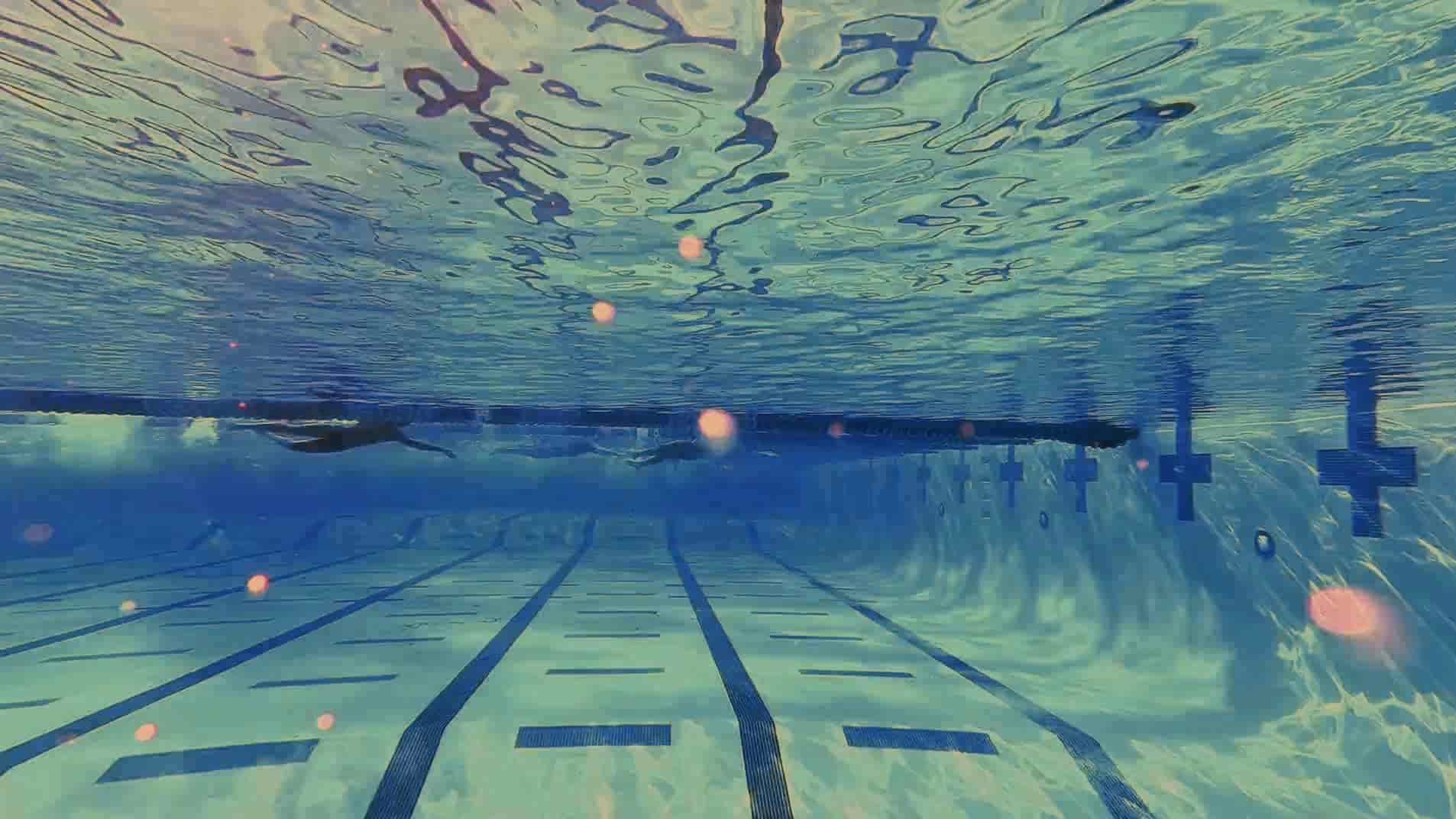}
\end{minipage}
\begin{minipage}[b]{0.1208\linewidth}
    \includegraphics[width=\linewidth]{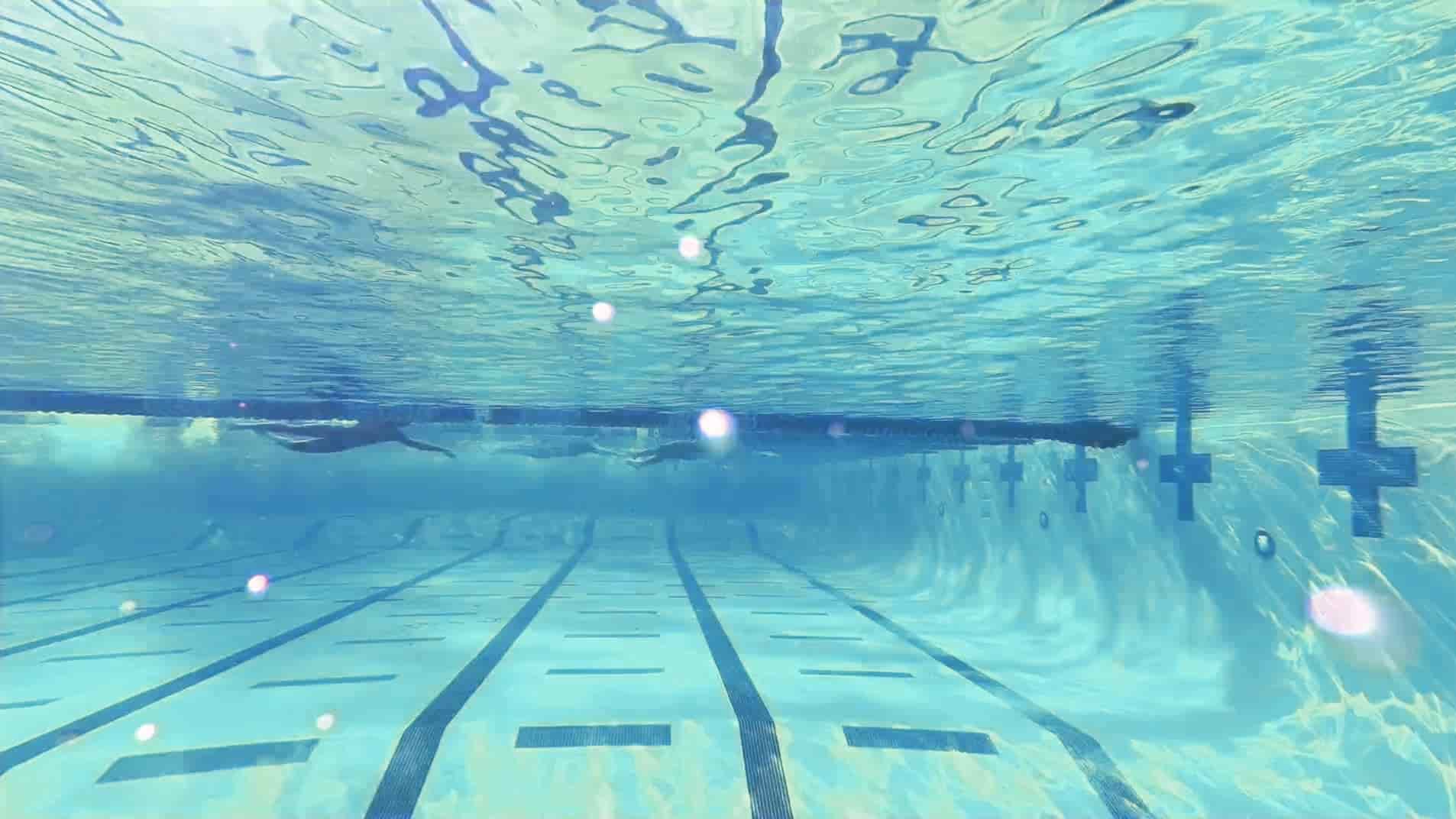}
\end{minipage}
\begin{minipage}[b]{0.1208\linewidth}
    \includegraphics[width=\linewidth]{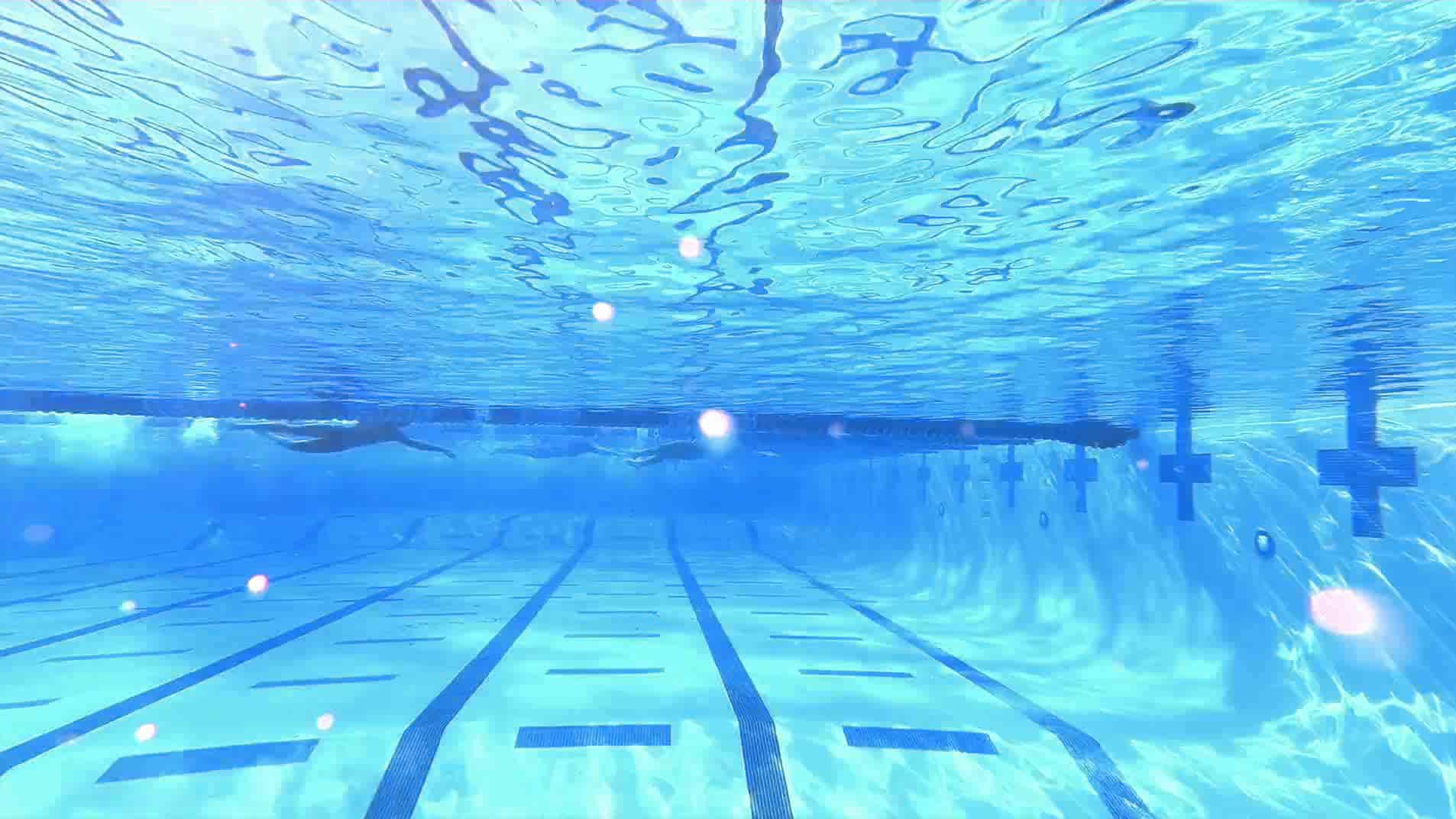}
\end{minipage}
\begin{minipage}[b]{0.1208\linewidth}
    \includegraphics[width=\linewidth]{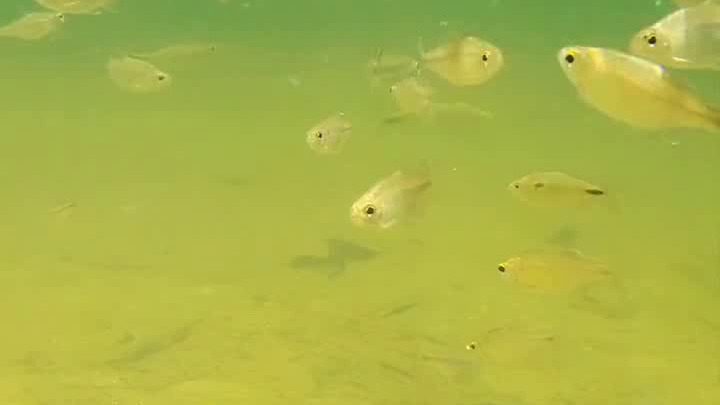}
    \subcaption*{Raw}
\end{minipage}
\begin{minipage}[b]{0.1208\linewidth}
    \includegraphics[width=\linewidth]{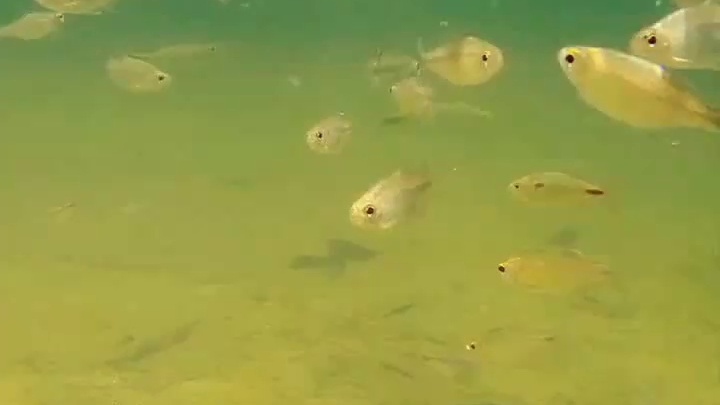}
    \subcaption*{Red Channle~\cite{galdran2015automatic}}
\end{minipage}
\begin{minipage}[b]{0.1208\linewidth}
    \includegraphics[width=\linewidth]{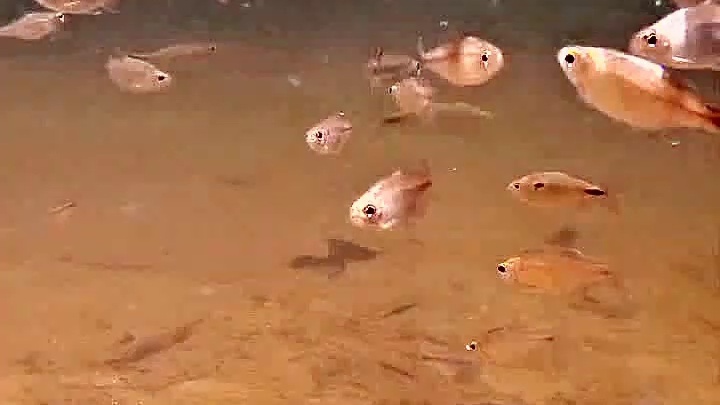}
    \subcaption*{MLLE~\cite{zhang2022underwater}}
\end{minipage}
\begin{minipage}[b]{0.1208\linewidth}
    \includegraphics[width=\linewidth]{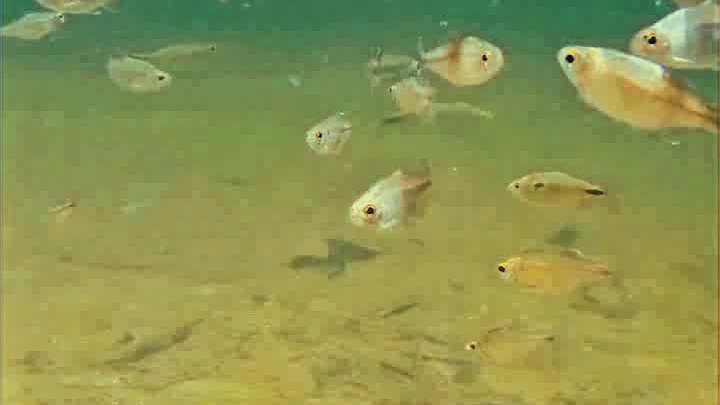}
    \subcaption*{URanker~\cite{guo2023underwater}}
\end{minipage} 
\begin{minipage}[b]{0.1208\linewidth}
    \includegraphics[width=\linewidth]{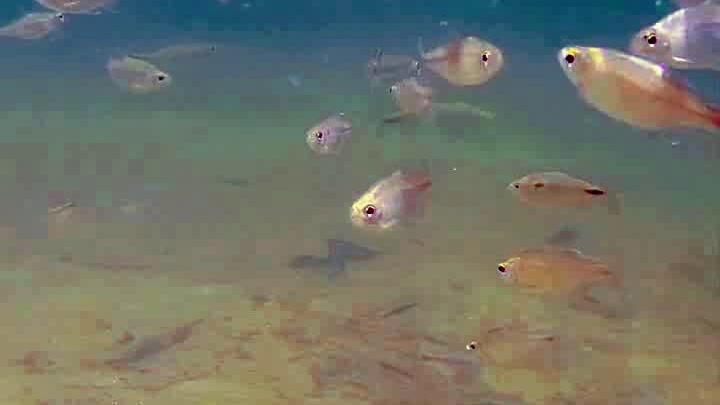}
    \subcaption*{USUIR~\cite{fu2022unsupervised}}
\end{minipage}
\begin{minipage}[b]{0.1208\linewidth}
    \includegraphics[width=\linewidth]{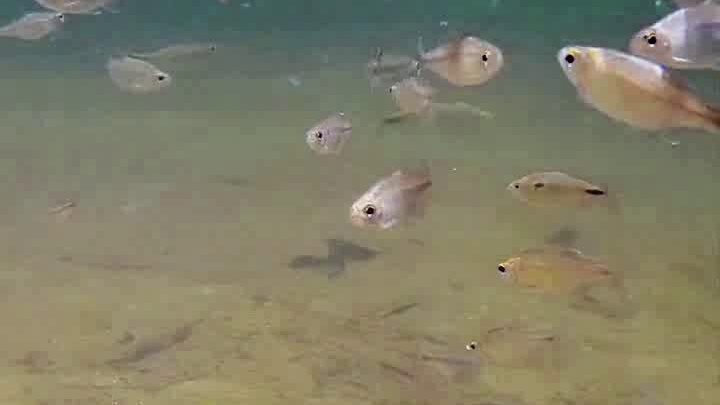}
    \subcaption*{PUIE~\cite{fu2022uncertainty}}
\end{minipage}
\begin{minipage}[b]{0.1208\linewidth}
    \includegraphics[width=\linewidth]{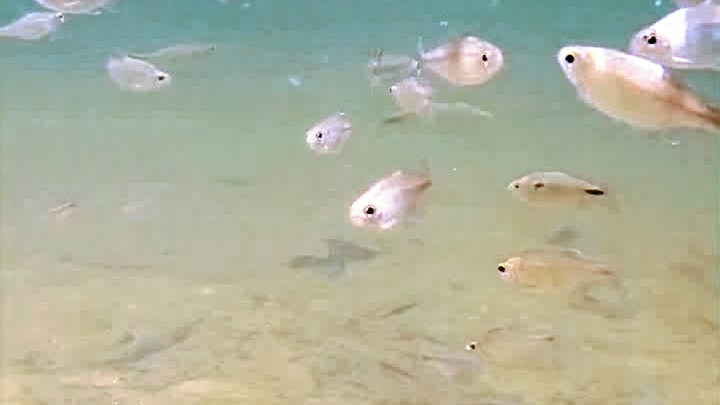}
    \subcaption*{Ours}
\end{minipage}
\begin{minipage}[b]{0.1208\linewidth}
    \includegraphics[width=\linewidth]{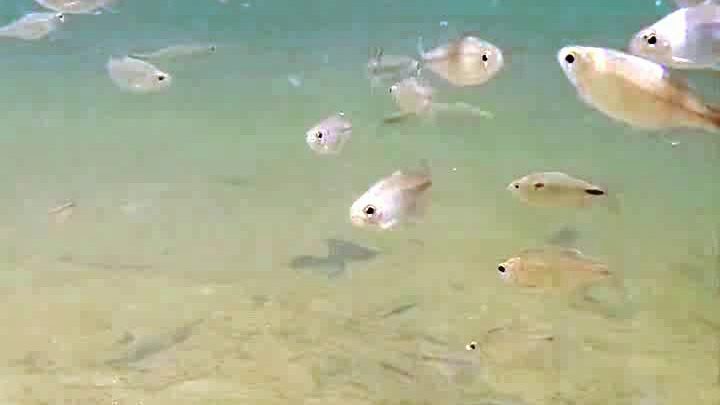}
    \subcaption*{GT}
\end{minipage}
    \caption{Visual comparisons with state-of-the-art methods on real underwater scenes.}
    \label{fig:results}
\end{figure*}
\label{sec:experiments}
\subsection{Settings}
{\bf Datasets.} 
UVEB contains 1208 paired training videos and 100 paired testing videos under 6 different scenes.

\noindent{\bf Comparison methods.} We only compare our method against 20 underwater image enhancement methods due to the lack of underwater video enhancement methods. Our UVE-Net is the first supervised underwater video enhancement method. The R, R$/$2, and R$/$6 are set to 30, 15, and 5 residual blocks in our model. We also provide a simplified model (''Ours-s'') to meet limited computational requirements. For the simplified model, the R, R$/$2, and R$/$6 are changed to 10, 3, and 1 residual blocks. The $t$ is set to 1 in the two models.
 
\noindent{\bf Evaluation metrics.} We utilize PSNR and MSE to evaluate the enhancement performance quantitatively. We also record memory usage and time cost for different methods during inferencing the UHD 4K videos.

\noindent{\bf Implementation details.} We implement our method with PyTorch and train it on 4 NVIDIA Tesla A40 GPUs. We use an ADAM optimizer for network optimization. The initial learning rate is set as 2$\times$${10}^{-4}$. The total number of iterations is 150K. The batch size is 4, and the patch size of input video frames is 512$\times$512.
\begin{table}[h]
\caption{Quantitative comparisons of enhanced video quality on UVEB dataset. In the Memory column, $^*$ represents the CPU, while without $^*$ represents the GPU. Top $1_{st}$, $2_{nd}$ results are marked in \textcolor{red}{red} and \textcolor{blue}{blue} respectively.}
\centering
\resizebox{0.47\textwidth}{!}{%
    \begin{tabular}{ccccc}
      \toprule  
      Methods& PSNR(dB)$\uparrow$& MSE($\times10^3$)$\downarrow$& Infrence time(s)& Memory(G)  \\
      \midrule  
      DCP~\cite{5206515}& 13.03& 3.7708& 1.8394& 0.05$^*$\\
      UDCP~\cite{drews2016underwater}& 10.75& 6.2848& 70.9177& 0.38$^*$\\
      GDCP~\cite{peng2018generalization}& 13.33& 3.7112& 7.6557& 0.74$^*$\\
      fusion-based~\cite{ancuti2012enhancing}& 17.73& 1.3916& 5.9321& 0.91$^*$\\
      MSCNN~\cite{ren2016single}& 13.17& 3.6562& 49.2594& 2.53$^*$\\
      Red Channle~\cite{galdran2015automatic}& 19.61& 1.0549& 5.6375& 0.59$^*$\\
      retinex-based~\cite{fu2014retinex}& 18.75& 1.1917& 9.0674& 0.74$^*$\\
      CLAHE~\cite{10.5555/180895.180940}& 19.71& 0.9139& 0.0503& 0.05$^*$\\
      GC~\cite{schlick1995quantization}& 16.61& 1.9759& 0.8557& 0.05$^*$\\
      HE~\cite{hummel1975image}& 15.78& 2.0156& \textcolor{red}{0.0403}& 0.05$^*$\\
      MLLE~\cite{zhang2022underwater}& 18.79& 1.2805& 7.2611& 1.22$^*$\\
      WWPF~\cite{zhang2023underwater}& 17.67& 1.4640& 17.6036& 1.15$^*$\\
      FspiralGAN~\cite{guan2023fast}& 18.67& 1.2353& \textcolor{blue}{0.0474}& 12.58\\
      CLUIE~\cite{li2022beyond}& 19.44& 1.0226& 0.4098& 24.68\\
      FA$^{+}$Net~\cite{jiang2023five}& 15.34& 2.3076& 0.1663& \textcolor{blue}{9.47}\\
      LANet~\cite{liu2022adaptive}& 21.49& 0.8369& 8.540& 33.99\\
      MetaUE~\cite{zhang2023metaue}& 15.91& 1.8831& 0.3784& 14.67\\
      PUIE~\cite{fu2022uncertainty}& 24.21& 0.4335& 0.5339& 33.64\\
      URanker~\cite{guo2023underwater}& 23.93& \textcolor{blue}{0.4286}& 0.2103& 14.74\\
      USUIR~\cite{fu2022unsupervised}& 21.64& 0.6516& 0.4208& 10.33\\
      UVE-Net-s (Ours-s)& \textcolor{blue}{24.43}& 0.5787& 0.0910& \textcolor{red}{5.6}\\
      UVE-Net (Ours)& \textcolor{red}{26.27}& \textcolor{red}{0.4059}& 0.675& 11.04\\
      \bottomrule 
    \end{tabular}
    }
    \label{tab:quanti}
\end{table}
\subsection{Comparisons with State-of-the-Art Methods}
{\bf Quantitative comparison.} \cref{tab:quanti} summarizes the quantitative results of our network and compares methods on UVEB. Our method outperforms other methods by a significant margin in PSNR and MSE metrics from these quantitative results. 
Specifically, our method further improves the PSNR from 24.21 dB to 26.27 dB and the MSE from 0.4286 to 0.4059.
For 4K videos, our simplified model (UVE-Net-s) has the smallest memory cost during inferencing the enhanced result of per frame. The UVE-Net-s can achieve an inference speed of 11 Frames per rate (FPS) on 4K videos as shown in the \cref{tab:quanti} and 25 FPS on 2K videos.

\noindent{\bf Qualitative comparison.} \cref{fig:results} visually compares enhanced results produced by our network and other methods on UVEB. Compared methods often lead to color distortion and noise in the enhancement results, while UVE-Net can remove color distortion better.

\subsection{Ablation Studies}
\hspace{0.16667in}We conduct a series of ablation studies to analyze the effectiveness of major components of our network. As shown in \cref{tab:Ablation}, the VE-Net here means the upper branch of UVE-Net with FEGM and FRGM replaced by two sets of traditional convolutions. VE-Net and FEGM mean a set of traditional convolutions replaced by the FEGM model. 

\noindent{\bf Effectiveness of network design.}
From columns (a), (b), and (d) of \cref{tab:Ablation}, compared to using only the upper branch, using the entire network has better quality improvement and less computational costs with the help of AE-Net. These improvements prove the effectiveness of network design. 

The lower branch converts its enhancement process into action information (convolutional kernels) and transmits it to the upper branch, allowing the upper branch to perform feature extraction and enhancement more efficiently. This strategy brings a significant improvement in the overall network performance. The entire network has fewer computational costs than the upper branch due to using group convolutions in FEGM and FRGM and processing low-resolution images in AE-Net, which have few computational costs.
\begin{table}
\caption{Ablation studies of major components in UVE-Net.}
\centering
\resizebox{0.47\textwidth}{!}{%
    \begin{tabular}{ccccc}
    \toprule
     & (a)& (b)& (c)& (d)\\
    \midrule  
    VE-Net& & \checkmark & \checkmark& \checkmark\\
    AE-Net& \checkmark & & \checkmark& \checkmark\\
    FEGM& & &\checkmark &\checkmark\\
    FRGM& & & &\checkmark\\
    \toprule  
    PSNR&24.41 &25.15 &26.20 &26.27\\
    MSE($\times10^3$)&0.5066 &0.4452 &0.4226 &0.4059\\
    Inference time(s)&2.749 &0.8827 &0.6943 &0.675 \\
    Memory(G)&5.48 &10.18 &11.04 &11.04 \\
    TFLOPs&10.34 &13.95 &12.79 &11.42  \\
    \bottomrule 
    \end{tabular}
    }
    \label{tab:Ablation}
\end{table}

\noindent{\bf Effectiveness of FEGM and FRGM.} The results in columns (b), (c) and (d) of \cref{tab:Ablation} verify the effectiveness of FEGM and FRGM. From columns (a), (b), (c) and (d), FEGM brings significant performance improvements to the network, and FRGM further improves network performance in the MSE metric from 0.4226 to 0.4059. The introduction of each module not only reduces computational complexity but also accelerates inference time. Both the FEGM and FRGM modules help the VE-Net complete enhancement faster and better.

%% file: sec/7_Conclusion.tex
\section{Conclusion}
\hspace{0.16667in}We propose the first large-scale and high-resolution paired underwater video enhancement benchmark. Our proposed UVEB dataset includes multiple types of underwater video degradation with assessment scores. Extensive experiments verify the superiority of the proposed UVE-Net on underwater video enhancement tasks. We also proposed a simplified model, UVE-Net-s, which enables real-time inference of 2K videos with good performance.

\noindent{\bf Acknowledgement.} 
This work was supported by  the National Natural Science Foundation of China (Grant No. 62171419), the finance science and technology Q19 project of 630 Hainan province of China under Grant Number ZDKJ202017 and  the Hainan Province Science and Technology Special Fund of China (Grant No. ZDYF2022SHFZ318).

%% file: main.bbl
\begin{thebibliography}{10}

\bibitem{huang2023contrastive}
Shirui Huang, Keyan Wang, Huan Liu, Jun Chen, and Yunsong Li.
\newblock Contrastive semi-supervised learning for underwater image restoration via reliable bank.
\newblock In {\em Proceedings of the IEEE/CVF Conference on Computer Vision and Pattern Recognition}, pages 18145--18155, 2023.

\bibitem{dai2017deformable}
Jifeng Dai, Haozhi Qi, Yuwen Xiong, Yi~Li, Guodong Zhang, Han Hu, and Yichen Wei.
\newblock Deformable convolutional networks.
\newblock In {\em Proceedings of the IEEE international conference on computer vision}, pages 764--773, 2017.

\bibitem{chan2022basicvsr++}
Kelvin~CK Chan, Shangchen Zhou, Xiangyu Xu, and Chen~Change Loy.
\newblock Basicvsr++: Improving video super-resolution with enhanced propagation and alignment.
\newblock In {\em Proceedings of the IEEE/CVF conference on computer vision and pattern recognition}, pages 5972--5981, 2022.

\bibitem{zhang2021learning}
Xinyi Zhang, Hang Dong, Jinshan Pan, Chao Zhu, Ying Tai, Chengjie Wang, Jilin Li, Feiyue Huang, and Fei Wang.
\newblock Learning to restore hazy video: A new real-world dataset and a new method.
\newblock In {\em Proceedings of the IEEE/CVF Conference on Computer Vision and Pattern Recognition}, pages 9239--9248, 2021.

\bibitem{patil2022video}
Prashant~W Patil, Sunil Gupta, Santu Rana, and Svetha Venkatesh.
\newblock Video restoration framework and its meta-adaptations to data-poor conditions.
\newblock In {\em European Conference on Computer Vision}, pages 143--160. Springer, 2022.

\bibitem{li2023progressive}
Runde Li and Lei Chen.
\newblock Progressive deep video dehazing without explicit alignment estimation.
\newblock {\em Applied Intelligence}, 53(10):12437--12447, 2023.

\bibitem{zhang2018adversarial}
Kaihao Zhang, Wenhan Luo, Yiran Zhong, Lin Ma, Wei Liu, and Hongdong Li.
\newblock Adversarial spatio-temporal learning for video deblurring.
\newblock {\em IEEE Transactions on Image Processing}, 28(1):291--301, 2018.

\bibitem{jin2023multi}
Shuo Jin, Meiqin Liu, Yu~Guo, Chao Yao, and Mohammad~S Obaidat.
\newblock Multi-frame correlated representation network for video super-resolution.
\newblock In {\em 2023 International Conference on Computer, Information and Telecommunication Systems (CITS)}, pages 01--07. IEEE, 2023.

\bibitem{li2023simple}
Dasong Li, Xiaoyu Shi, Yi~Zhang, Ka~Chun Cheung, Simon See, Xiaogang Wang, Hongwei Qin, and Hongsheng Li.
\newblock A simple baseline for video restoration with grouped spatial-temporal shift.
\newblock In {\em Proceedings of the IEEE/CVF Conference on Computer Vision and Pattern Recognition}, pages 9822--9832, 2023.

\bibitem{gao2023vdpve}
Yixuan Gao, Yuqin Cao, Tengchuan Kou, Wei Sun, Yunlong Dong, Xiaohong Liu, Xiongkuo Min, and Guangtao Zhai.
\newblock Vdpve: Vqa dataset for perceptual video enhancement.
\newblock In {\em Proceedings of the IEEE/CVF Conference on Computer Vision and Pattern Recognition}, pages 1474--1483, 2023.

\bibitem{series2012methodology}
BT~Series.
\newblock Methodology for the subjective assessment of the quality of television pictures.
\newblock {\em Recommendation ITU-R BT}, 500(13), 2012.

\bibitem{akkaynak2019sea}
Derya Akkaynak and Tali Treibitz.
\newblock Sea-thru: A method for removing water from underwater images.
\newblock In {\em Proceedings of the IEEE/CVF conference on computer vision and pattern recognition}, pages 1682--1691, 2019.

\bibitem{zhang2023metaue}
Zhenwei Zhang, Haorui Yan, Ke~Tang, and Yuping Duan.
\newblock Metaue: Model-based meta-learning for underwater image enhancement.
\newblock {\em arXiv preprint arXiv:2303.06543}, 2023.

\bibitem{li2020underwater}
Chongyi Li, Saeed Anwar, and Fatih Porikli.
\newblock Underwater scene prior inspired deep underwater image and video enhancement.
\newblock {\em Pattern Recognition}, 98:107038, 2020.

\bibitem{li2019underwater}
Chongyi Li, Chunle Guo, Wenqi Ren, Runmin Cong, Junhui Hou, Sam Kwong, and Dacheng Tao.
\newblock An underwater image enhancement benchmark dataset and beyond.
\newblock {\em IEEE Transactions on Image Processing}, 29:4376--4389, 2019.

\bibitem{peng2023u}
U-shape transformer for underwater image enhancement.
\newblock {\em IEEE Transactions on Image Processing}, 2023.

\bibitem{fu2022uncertainty}
Zhenqi Fu, Wu~Wang, Yue Huang, Xinghao Ding, and Kai-Kuang Ma.
\newblock Uncertainty inspired underwater image enhancement.
\newblock In {\em European Conference on Computer Vision}, pages 465--482. Springer, 2022.

\bibitem{liu2022adaptive}
Shiben Liu, Huijie Fan, Sen Lin, Qiang Wang, Naida Ding, and Yandong Tang.
\newblock Adaptive learning attention network for underwater image enhancement.
\newblock {\em IEEE Robotics and Automation Letters}, 7(2):5326--5333, 2022.

\bibitem{guan2023fast}
Yang Guan, Xiaoyan Liu, Zhibin Yu, Yubo Wang, Xingyu Zheng, Shaoda Zhang, and Bing Zheng.
\newblock Fast underwater image enhancement based on a generative adversarial framework.
\newblock {\em Frontiers in Marine Science}, 9:964600, 2023.

\bibitem{varghese2023self}
Nisha Varghese, Ashish Kumar, and AN~Rajagopalan.
\newblock Self-supervised monocular underwater depth recovery, image restoration, and a real-sea video dataset.
\newblock In {\em Proceedings of the IEEE/CVF International Conference on Computer Vision}, pages 12248--12258, 2023.

\bibitem{guo2023sky}
Yun Guo, Xueyao Xiao, Yi~Chang, Shumin Deng, and Luxin Yan.
\newblock From sky to the ground: A large-scale benchmark and simple baseline towards real rain removal.
\newblock In {\em Proceedings of the IEEE/CVF International Conference on Computer Vision}, pages 12097--12107, 2023.

\bibitem{chen2023snow}
Haoyu Chen, Jingjing Ren, Jinjin Gu, Hongtao Wu, Xuequan Lu, Haoming Cai, and Lei Zhu.
\newblock Snow removal in video: A new dataset and a novel method.
\newblock In {\em Proceedings of the IEEE/CVF International Conference on Computer Vision}, pages 13211--13222, 2023.

\bibitem{xu2023video}
Jiaqi Xu, Xiaowei Hu, Lei Zhu, Qi~Dou, Jifeng Dai, Yu~Qiao, and Pheng-Ann Heng.
\newblock Video dehazing via a multi-range temporal alignment network with physical prior.
\newblock In {\em Proceedings of the IEEE/CVF Conference on Computer Vision and Pattern Recognition}, pages 18053--18062, 2023.

\bibitem{alawode2022utb180}
Basit Alawode, Yuhang Guo, Mehnaz Ummar, Naoufel Werghi, Jorge Dias, Ajmal Mian, and Sajid Javed.
\newblock Utb180: A high-quality benchmark for underwater tracking.
\newblock In {\em Proceedings of the Asian Conference on Computer Vision}, pages 3326--3342, 2022.

\bibitem{5206515}
Kaiming He, Jian Sun, and Xiaoou Tang.
\newblock Single image haze removal using dark channel prior.
\newblock In {\em 2009 IEEE Conference on Computer Vision and Pattern Recognition}, pages 1956--1963, 2009.

\bibitem{drews2016underwater}
Paulo~LJ Drews, Erickson~R Nascimento, Silvia~SC Botelho, and Mario Fernando~Montenegro Campos.
\newblock Underwater depth estimation and image restoration based on single images.
\newblock {\em IEEE computer graphics and applications}, 36(2):24--35, 2016.

\bibitem{peng2018generalization}
Yan-Tsung Peng, Keming Cao, and Pamela~C Cosman.
\newblock Generalization of the dark channel prior for single image restoration.
\newblock {\em IEEE Transactions on Image Processing}, 27(6):2856--2868, 2018.

\bibitem{ancuti2012enhancing}
Cosmin Ancuti, Codruta~Orniana Ancuti, Tom Haber, and Philippe Bekaert.
\newblock Enhancing underwater images and videos by fusion.
\newblock In {\em 2012 IEEE conference on computer vision and pattern recognition}, pages 81--88. IEEE, 2012.

\bibitem{fu2014retinex}
Xueyang Fu, Peixian Zhuang, Yue Huang, Yinghao Liao, Xiao-Ping Zhang, and Xinghao Ding.
\newblock A retinex-based enhancing approach for single underwater image.
\newblock In {\em 2014 IEEE international conference on image processing (ICIP)}, pages 4572--4576. IEEE, 2014.

\bibitem{carlevaris2010initial}
Nicholas Carlevaris-Bianco, Anush Mohan, and Ryan~M Eustice.
\newblock Initial results in underwater single image dehazing.
\newblock In {\em Oceans 2010 Mts/IEEE Seattle}, pages 1--8. IEEE, 2010.

\bibitem{galdran2015automatic}
Adrian Galdran, David Pardo, Artzai Pic{\'o}n, and Aitor Alvarez-Gila.
\newblock Automatic red-channel underwater image restoration.
\newblock {\em Journal of Visual Communication and Image Representation}, 26:132--145, 2015.

\bibitem{lu2015contrast}
Huimin Lu, Yujie Li, Lifeng Zhang, and Seiichi Serikawa.
\newblock Contrast enhancement for images in turbid water.
\newblock {\em JOSA A}, 32(5):886--893, 2015.

\bibitem{2014Generative}
Ian Goodfellow, Jean Pouget-Abadie, Mehdi Mirza, Bing Xu, David Warde-Farley, Sherjil Ozair, Aaron Courville, and Y.~Bengio.
\newblock Generative adversarial nets.
\newblock In {\em Neural Information Processing Systems}, 2014.

\bibitem{li2017watergan}
Jie Li, Katherine~A Skinner, Ryan~M Eustice, and Matthew Johnson-Roberson.
\newblock Watergan: Unsupervised generative network to enable real-time color correction of monocular underwater images.
\newblock {\em IEEE Robotics and Automation letters}, 3(1):387--394, 2017.

\bibitem{fabbri2018enhancing}
Cameron Fabbri, Md~Jahidul Islam, and Junaed Sattar.
\newblock Enhancing underwater imagery using generative adversarial networks.
\newblock In {\em 2018 IEEE international conference on robotics and automation (ICRA)}, pages 7159--7165. IEEE, 2018.

\bibitem{ren2016single}
Wenqi Ren, Si~Liu, Hua Zhang, Jinshan Pan, Xiaochun Cao, and Ming-Hsuan Yang.
\newblock Single image dehazing via multi-scale convolutional neural networks.
\newblock In {\em Computer Vision--ECCV 2016: 14th European Conference, Amsterdam, The Netherlands, October 11-14, 2016, Proceedings, Part II 14}, pages 154--169. Springer, 2016.

\bibitem{hummel1975image}
Robert Hummel.
\newblock Image enhancement by histogram transformation.
\newblock {\em Unknown}, 1975.

\bibitem{schlick1995quantization}
Christophe Schlick.
\newblock Quantization techniques for visualization of high dynamic range pictures.
\newblock In {\em Photorealistic rendering techniques}, pages 7--20. Springer, 1995.

\bibitem{zhang2022underwater}
Weidong Zhang, Peixian Zhuang, Hai-Han Sun, Guohou Li, Sam Kwong, and Chongyi Li.
\newblock Underwater image enhancement via minimal color loss and locally adaptive contrast enhancement.
\newblock {\em IEEE Transactions on Image Processing}, 31:3997--4010, 2022.

\bibitem{zhang2023underwater}
Weidong Zhang, Ling Zhou, Peixian Zhuang, Guohou Li, Xipeng Pan, Wenyi Zhao, and Chongyi Li.
\newblock Underwater image enhancement via weighted wavelet visual perception fusion.
\newblock {\em IEEE Transactions on Circuits and Systems for Video Technology}, 2023.

\bibitem{li2022beyond}
Kunqian Li, Li~Wu, Qi~Qi, Wenjie Liu, Xiang Gao, Liqin Zhou, and Dalei Song.
\newblock Beyond single reference for training: underwater image enhancement via comparative learning.
\newblock {\em IEEE Transactions on Circuits and Systems for Video Technology}, 2022.

\bibitem{10.5555/180895.180940}
Karel Zuiderveld.
\newblock {\em Contrast Limited Adaptive Histogram Equalization}, page 474–485.
\newblock Academic Press Professional, Inc., USA, 1994.

\bibitem{fu2022unsupervised}
Zhenqi Fu, Huangxing Lin, Yan Yang, Shu Chai, Liyan Sun, Yue Huang, and Xinghao Ding.
\newblock Unsupervised underwater image restoration: From a homology perspective.
\newblock In {\em Proceedings of the AAAI Conference on Artificial Intelligence}, volume~36, pages 643--651, 2022.

\bibitem{guo2023underwater}
Chunle Guo, Ruiqi Wu, Xin Jin, Linghao Han, Weidong Zhang, Zhi Chai, and Chongyi Li.
\newblock Underwater ranker: Learn which is better and how to be better.
\newblock In {\em AAAI Conference on Artificial Intelligence}, volume~37, pages 702--709, 2023.

\bibitem{han2020underwater}
Ruyue Han, Yang Guan, Zhibin Yu, Peng Liu, and Haiyong Zheng.
\newblock Underwater image enhancement based on a spiral generative adversarial framework.
\newblock {\em IEEE Access}, 8:218838--218852, 2020.

\bibitem{islam2020fast}
Md~Jahidul Islam, Youya Xia, and Junaed Sattar.
\newblock Fast underwater image enhancement for improved visual perception.
\newblock {\em IEEE Robotics and Automation Letters}, 5(2):3227--3234, 2020.

\bibitem{naik2021shallow}
Ankita Naik, Apurva Swarnakar, and Kartik Mittal.
\newblock Shallow-uwnet: Compressed model for underwater image enhancement (student abstract).
\newblock In {\em AAAI Conference on Artificial Intelligence}, volume~35, pages 15853--15854, 2021.

\bibitem{jiang2023five}
Jingxia Jiang, Tian Ye, Jinbin Bai, Sixiang Chen, Wenhao Chai, Shi Jun, Yun Liu, and Erkang Chen.
\newblock Five a $^+$ network: You only need 9k parameters for underwater image enhancement.
\newblock {\em arXiv preprint arXiv:2305.08824}, 2023.

\bibitem{wang2022generation}
Zheyin Wang, Liquan Shen, Zhengyong Wang, Yufei Lin, and Yanliang Jin.
\newblock Generation-based joint luminance-chrominance learning for underwater image quality assessment.
\newblock {\em IEEE Transactions on Circuits and Systems for Video Technology}, 33(3):1123--1139, 2022.

\bibitem{lai2018fast}
Wei-Sheng Lai, Jia-Bin Huang, Narendra Ahuja, and Ming-Hsuan Yang.
\newblock Fast and accurate image super-resolution with deep laplacian pyramid networks.
\newblock {\em IEEE transactions on pattern analysis and machine intelligence}, 41(11):2599--2613, 2018.

\bibitem{pan2023deep}
Jinshan Pan, Boming Xu, Jiangxin Dong, Jianjun Ge, and Jinhui Tang.
\newblock Deep discriminative spatial and temporal network for efficient video deblurring.
\newblock In {\em Proceedings of the IEEE/CVF Conference on Computer Vision and Pattern Recognition}, pages 22191--22200, 2023.


\bibitem{RUIEORE}
Risheng Liu, Xin Fan, Ming Zhu, Minjun Hou, and Zhongxuan Luo.
\newblock Real-world underwater enhancement: Challenges, benchmarks, and solutions under natural light.
\newblock {\em IEEE Transactions on Circuits and Systems for Video Technology}, 30(12):4861--4875, 2020.

\bibitem{SAUD}
Qiuping Jiang, Yuese Gu, Chongyi Li, Runmin Cong, and Feng Shao.
\newblock Underwater image enhancement quality evaluation: Benchmark dataset and objective metric.
\newblock {\em IEEE Transactions on Circuits and Systems for Video Technology}, 32(9):5959--5974, 2022.

\end{thebibliography}
